\documentclass{article}
\usepackage{amssymb}
\usepackage{amsmath}
\usepackage{amsthm}
\usepackage[a4paper]{geometry}
\usepackage[english]{babel}
\usepackage{enumitem}
\usepackage{soul}
\usepackage[square,sort,comma,numbers]{natbib}
\usepackage{accents}
\usepackage[dvipsnames]{xcolor}
\usepackage[colorlinks,linktoc=true,
             citecolor= Cerulean!85!green!90!black,%Cerulean!75!green,
             linkcolor=YellowOrange,
             urlcolor=RubineRed!85!black,
             ]{hyperref}
             
\usepackage{graphicx}
\usepackage{wrapfig}
\usepackage{multirow}
\usepackage{extarrows}

\newcommand{\X}{\mathbf{X}}
\newcommand{\Y}{\mathbf{Y}}

\newcommand{\y}{\mathbf{y}}
\newcommand{\Z}{\mathbf{Z}}

\DeclareMathOperator*{\argmax}{arg\,max}
\DeclareMathOperator*{\argmin}{arg\,min}

\newtheorem{example}{Example}

\definecolor{airforceblue}{rgb}{0.36, 0.54, 0.66}
\definecolor{aliceblue}{rgb}{0.94, 0.97, 1.0}
\definecolor{darkgray}{rgb}{0.66, 0.66, 0.66}
\definecolor{darkgreen}{rgb}{0.0, 0.5, 0.0}

\definecolor{causalred}{HTML}{c00000}
\definecolor{spuriousblue}{HTML}{0070c0}
\definecolor{mixpurple}{HTML}{7030a0}
\definecolor{confoundergreen}{HTML}{71ad47}

\definecolor{pinkish}{rgb}{1,0.4,1}

\newcommand{\mch}[2]{{\color{orange}{\it }}}
\newcommand{\mc}[2]{{\color{orange}{[\textbf{Maheep:} ]}}}
\newcommand{\com}[3]{{\color{teal}{}}\xspace}

\usepackage{bbm}
\usepackage{longtable}
\newcommand{\indep}[3]{\perp\!\!\!\!\perp} 
\newcommand{\nindep}[3]{\not\!\perp\!\!\!\perp}
\usepackage{csquotes}

\usepackage[most]{tcolorbox}

\newtcbtheorem[auto counter,number within=section]{theo}%
  {Theorem}{fonttitle=\bfseries\upshape, fontupper=\slshape,
     arc=0mm, colback=blue!5!white,colframe=blue!75!black}{theorem}

\newtcbtheorem[auto counter,number within=section]{definition}%
  {Definition}{fonttitle=\bfseries\upshape, fontupper=\slshape,
     arc=0mm, colback=aliceblue,colframe=gray}{theorem}

\usepackage{subfig}

\usepackage{hyphenat}
\usepackage{blindtext}
\usepackage{breqn}

\begin{document}

%--------------------End Maheep's inserted packages------------

\newcommand{\hla}[1]{{\color{green}{#1}}}
\newcommand{\hlc}[1]{{\color{darkgreen}{\it [haoyang: #1]}}}

%%% Put macros here

\title{ Towards Trustworthy and Aligned Machine Learning:\\
{\Large A Data-centric Survey with Causality Perspectives}
}
\author{Haoyang Liu$^\dagger$, \, Maheep Chaudhary$^{\dagger}$\thanks{work done as remote intern, originally with Bundelkhand Institute of Engineering and Technology, joining Nanyang Technological University.}, and Haohan Wang \\[0.2cm]
School of Information Sciences,\\
University of Illinois Urbana-Champaign\\
\textsf{\{hl57, haohanw\}@illinois.edu, maheep001@e.ntu.edu.sg}\\
{\footnotesize $^\dagger$ equal contribution}\\
}
\date{}

\maketitle
\begin{abstract}
The trustworthiness of machine learning has emerged as a critical topic in the field, encompassing various applications and research areas such as robustness, security, interpretability, and fairness. Over the past decade, dedicated efforts have been made to address these issues, resulting in a proliferation of methods tailored to each specific challenge. In this survey paper, we provide a systematic overview of the technical advancements in trustworthy machine learning, focusing on robustness, adversarial robustness, interpretability, and fairness from a data-centric perspective, as we believe that achieving trustworthiness in machine learning often involves overcoming challenges posed by the data structures that traditional empirical risk minimization (ERM) training cannot resolve.

Interestingly, we observe a convergence of methods introduced from this perspective, despite their development as independent solutions across various subfields of trustworthy machine learning. Furthermore, we find that Pearl's hierarchy of causality serves as a unifying framework for categorizing these techniques. Consequently, this survey first presents the background of trustworthy machine learning development using a unified set of concepts, connects this unified language to Pearl's hierarchy of causality, and finally discusses methods explicitly inspired by causality literature. By doing so, we established a unified language with mathematical vocabulary 
as a principled connection between these methods across robustness, adversarial robustness, interpretability, and fairness under a data-centric perspective, fostering a more cohesive understanding of the field. 

Further, we extend our study to the trustworthiness of large pretrained models. We first present a brief summary of the dominant techniques in these models, such as fine-tuning, parameter-efficient fine-tuning, prompting, and reinforcement learning with human feedback. We then connect these techniques with standard ERM, upon which previous trustworthy machine learning solutions were built. This connection allows us to immediately build upon the principled understanding of the trustworthy method established in previous sections, applying it to these new techniques in large pretrained models, openning up possibilities for many new methods.  
We also survey the current existing methods under this perspective. 

Finally, we offer a brief summary of the applications of these methods and also discuss about some future aspects relating to our survey\footnote{For more information, please visit \href{http://trustai.one}{trustAI.one}}.

\end{abstract}

\section{Introduction}
The rapid advancements and widespread adoption of machine learning (ML) have led to its integration into various applications, ranging from healthcare and finance to autonomous vehicles and social media. As these applications become increasingly complex and consequential, the trustworthiness of ML systems has emerged as a crucial factor in ensuring their reliability, safety, and societal impact. Trustworthy ML is a multifaceted concept, encompassing a broad range of research areas. Over the past decade, substantial efforts have been made to address these challenges, resulting in a multitude of specialized methods designed to tackle specific aspects of trustworthiness.

While the scope of trustworthy ML encompasses multiple areas, 
in this survey, we chose to concentrate on the topics of robustness, security, interpretability, and fairness due to their significant impact as well as their internal connections. 
By concentrating on the technical aspects of these topics, we strive to provide a comprehensive understanding of the state-of-the-art methods and techniques that have been developed to address these challenges. 

While concentrating on the technical aspects, 
it is important to note the diversity of the methods 
from different perspectives. 
Among a rich set of methods introduced in recent years, 
our survey primarily focuses on the machine learning technical aspects 
of the methods from a data-centric perspective.

\paragraph{Data-centric Perspective}
By data-centric perspective, we refer to the understanding that 
believes the challenges toward trustworthy machine learning 
lies in the structure of the available data 
provided to train the models. 
The existence of such structures, 
such as spurious features \cite{zhou2021examining, singla2022salient, joshi2022all}, 
confounding factors \cite{greenland1999confounding, vanderweele2013definition, babyak2009understanding}, 
dataset bias \cite{tommasi2015deeper, unbiasedlook, khosla2012undoing},
often leads to the consequence that
vanilla training with empirical risk minimization (ERM) 
will often generate models that capture such undesired signals 
from the data. 

Following this data-centric perspective, we introduce a concrete example 
for the following-up discussions throughout this survey. 

\begin{example}
\label{exp1}
In a classification task, considering the problem of classifying images of sea turtles and tortoise \citep{wang2019learning}. 
To address this task, images are labelled for both types of animals by human annotators reading the images.
However, sea turtles typically live in the sea
while tortoises live in various environments. 
As a result, 
most images of sea turtles will have a blue sea background 
while images of the other class can have various backgrounds, 
creating strong correlations between the color of the backgrounds
and the label. 
Thus, 
an ERM trained model will likely pick up the backgrounds of the images
due to the strongly correlated signals, 
while, on the other hand, 
a marine biologist will suggest we classify the images 
through their feet or shell. 
\end{example}

Example~\ref{exp1} is a straightforward demonstration for the widely-accepted remark \emph{``Correlation is not Causation''}. 
While this remark is not new to the 
statistics and machine learning communities, 
numerous efforts have nonetheless been devoted to 
improve the model's predictive performances over benchmark datasets, 
resulting in techniques selected and favored regardless of 
whether the models are learning causal features or spurious features.
Then the consequence should be easy to anticipate:
when these trained models are deployed in the real-world where the spurious features 
can be different from those in lab, 
the models will underperform.

\begin{figure}
    \centering
    \includegraphics[width=0.8\textwidth]{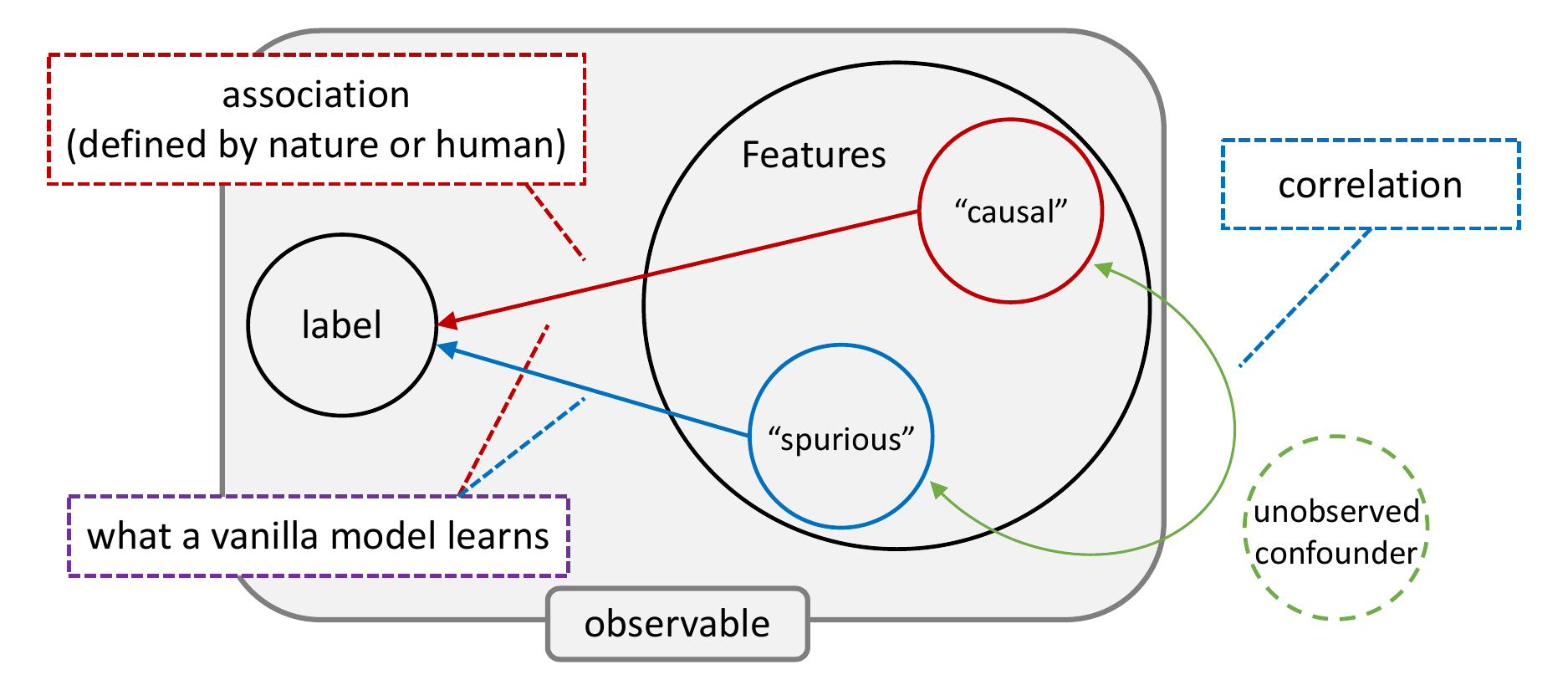}
    \caption{The core problem this survey paper scopes. We use the terms ``causal'' and ``spurious'' in quotes because the nature of these preferred features might depend on the context. For example, in the robustness generalization setting, the preferred features are usually the ``causal'' features, but in fairness discussion, the preferred features do not necessarily have to be ``causal'' features in the typical causality sense, although the mathematical tools used can be the same. 
    }
    \label{fig:main}
\end{figure}

There exist several explanations that potentially account for 
the performance disparities between 
the benchmark settings and the real-world settings \citep{storkey2009training, zhang2013domain,vahdat2017robustness}, 
but here we will devote the discussion to the data-centric understanding:
the performance gap is due to the fact that 
the models are learning through the spurious features 
in the datasets,                      
resulting in some over-estimated performances through the benchmarks \citep{wang2020high,geirhos2022imagenettrained, wang2023neural}.

Illustrated in Figure~\ref{fig:main} and as previously hypothesized by \citep{wang2020high}, 
the scope of this survey considers the root of the undesired performances 
in the real-world applications lies in the data: 
within the collected data used to train the supervised machine learning models, 
there exist some spurious features that are associated with the label, 
therefore, if a vanilla statistical model cannot differentiate the spurious features 
from the desired ones, the model will learn both the features and result in some undesired performances. 

It is also worth mentioning that the turtles vs. tortoise is not the only example used to highlight this issue in the literature. To the best of our knowledge, 
the first example in the deep learning era 
is the discussion on how the snow background 
is playing a role in Husky vs. wolf classification \cite{ribeiro2016should},
and then the community also often motivates their discussion 
with examples such as how the habitat plays a role in frog vs. animals without swamp scenes classification \cite{bahng2020learning} or camel vs. cow classification \cite{taghanaki2022masktune}, how the fisher is significantly correlated with fish classes in ImageNet classification \cite{brendel2019approximating}.

\paragraph{The Role of Causality}
Fortunately, for the problem in Example~\ref{exp1}, the solution is fairly apparent: 
\emph{causal analysis} is a field that studies the systematic way of 
understanding the causal relationship from data 
while staying least influenced by the statistical signals raised 
by spurious features or confounding factors. 
Thus, it seems to incorporate the established solutions of causal analysis 
into current deep learning solutions 
is a direction to solve the problem. 

Before we dive into the world of causal analysis, 
we find it beneficial to first clarify the role causality plays 
in machine learning with static datasets that are already collected beforehand. 
With the example of image classification, 
a question is often raised on 
``whether the pixels causes the label 
or the label causes the pixel.'' 
We consider neither of these directions is nature enough to serve the discussion, 
and we tend to formalize the problem in a way that
\emph{it is the pixels of the images that cause the human annotators of the dataset 
to label each image in a certain way}. 

With this setup of the problem, we can leverage the 
established concepts and solutions in causal analysis 
for machine learning problems. 
This survey aims to serve the role of summarizing the 
recent works incorporating the concepts and techniques from causal analysis 
for improving the robustness or interpretability of machine learning models, 
either when these works explicitly mention the inspiration 
or when these works potentially invent these techniques independently.
As we will demonstrate soon, while most of the methods 
are invented independently, 
they converge to the same statistical language, 
and the same statistical language is connected 
to Pearl's causal hierarchy.

\begin{figure}
    \centering
    \includegraphics[width=1.0\textwidth]{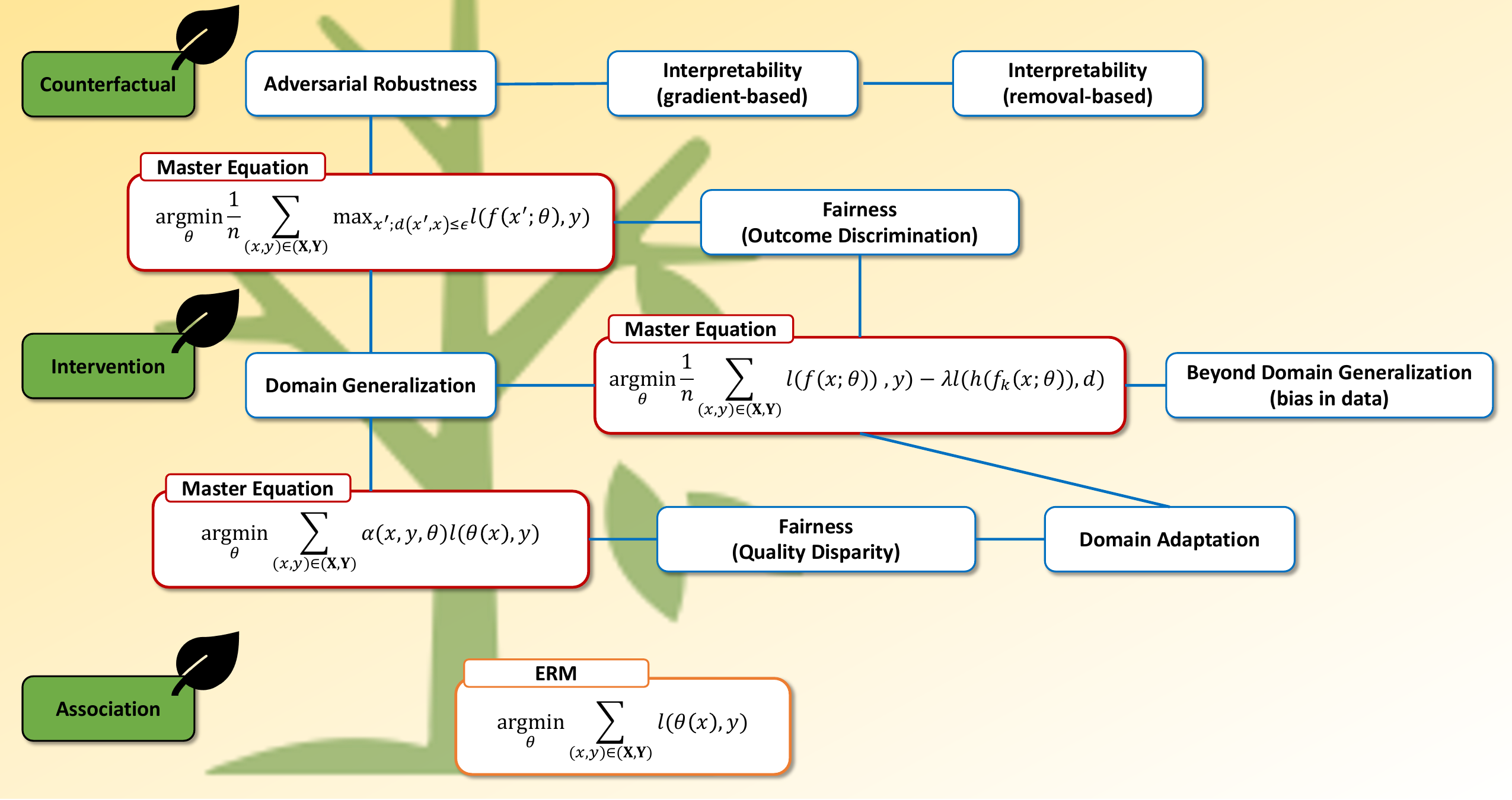}
    \caption{Summary of the major topics surveyed in this paper. Blue boxes: machine learning topics in trustworthy ML scoped by our survey; Red boxes: core techniques and master equations we summarized; Green boxs: the causality layers these techniques build upon.}
    \label{fig:summary}
\end{figure}

\paragraph{Survey Contents}
Therefore, our survey has the following main components (Figure~\ref{fig:summary}):
\begin{itemize}
    \item In Section~\ref{sec:robustness}, we will recapitulate the current state of the machine learning and deep learning techniques with an emphasis on the evidence suggesting the need of trustworthy machine learning in the real-world. 
    The section also serves the goal of delineating the background problems we are interested in surveying the techniques for, including 
    \begin{itemize}
        \item Robustness in generalization: In this survey, we further branch it into categories such as domain adaptation, domain generalization, and learning with the existence of dataset bias. We scope this topic as the study of how to maintain the predictive performances over additional datasets that human users consider similar. 
        \item Adversarial Robustness (Security): In this survey, we scope the topic regarding security mostly under the discussion of adversarial attack and defense, a study about how to carefully crafted certain noises imperceptible to human users but able to alter the model predictions, and a study about how to defend such noises. 
        \item Interpretability: In this survey, we use the term exchangeably with explainability, and consider it a study about how to translate the statistical decision process of models to users. We mostly focus on explaining the importance of features used by models. 
        \item Fairness: In this survey, we will discuss both fairness under the categories of outcome discrimination and quality disparity.
        We focus mostly on the technical aspects of the designing of the models given an established fairness criterion, instead of the design of such fairness criterion. 
    \end{itemize}
    While introducing these machine learning methods under each topic, we also condense the ideas behind each method to its mathematical backbone and demonstrate a conceptual principled understanding 
    of these methods through mathematical unification. 
    \item Section~\ref{sec:review:main} attempts to give the reader a thorough overview of various causality notions and a summary of the deep learning methods explicitly inspired by such notions. Overall, this section has been divided mostly into three key parts:
    \begin{itemize}
        \item The background of causality is summarised in Section~\ref{sec:background} using the structural causal model (SCM) and Pearl Hierarchy. Our vision is primarily focused on the debate surrounding the ``confounders,'' who are the principal villains in the quest for trustworthy machine learning. Additionally, we outline the fundamental issue with machine learning that makes confounding variables more likely to appear in the observed data and describe its effect on the estimated probability of output.
        \item Several causal notions that fall under the second level of causation, $\mathcal{L}_2$, are defined in Section~\ref{sec:intervention}; these concepts include the randomized controlled trial (RCT), instrument variable (IV), backdoor method, and front door method. Furthermore, we discuss the different works under the umbrella of the technique employed by these works to de-confound the machine learning model.
        \item Finally, we define the ideas that make up the third level of causation $\mathcal{L}_3$ in Section~\ref{sec:counterfactuals}. We also describe the ideas of treatment effects because they are partially based on $\mathcal{L}_3$. These notions are defined to a greater extent by exploring their employment in different machine learning works. 
    \end{itemize}
    Along with the introduction of these causality concepts, we will also delineate the machine learning methods that are explicitly supported by these concepts, and link these ideas back to the principled understanding in the previous section. 
    \item In section~\ref{sec:pretrained}, we put our discussions from the above two sections into the context of large pretrained models. 
    \begin{itemize}
        \item We first shift the views from the standalone models into a new paradigm of large pretrained models by offering summaries of the techniques dominating in this paradigm, such as fine-tuning, parameter-efficient fine-tuning, and prompting. Following this introduction, we will condense the fundamental concepts behind these techniques into the mathematical language of a typical ERM loss. 
        \item The essence of this summary allows us to apply the techniques discussed in prior sections to the realm of large pretrained models. To a certain extent, our mathematical summary holds the potential to predict future trustworthy machine learning methods that will be invented in the context of large pretrained models. We also offer a survey of the existing methods before our summary predicts.
    \end{itemize}
    
    \item Section~\ref{sec:application} summarizes the application of these techniques, mainly categorized as vision, language, and vision-language applications. 
    \item In Section~\ref{sec:con}, we will conclude with more explicit discussions of the sections above, and briefly discuss the potential future aspects under each perspective, and the unification language as a whole. 
\end{itemize}

\begin{table}
\footnotesize 
\centering
\begin{tabular}{p{3.5cm}cp{5cm}c}
\hline
category and rule & example & notation explanation & relationships \\ \hline
\multirow{9}{3.5cm}{data} & $\mathcal{X}$ & data space & $x\in\mathcal{X}$ \\
 & $\mathcal{Y}$ & label space & $y\in\mathcal{Y}$ \\
 & $(\X,\Y)$ & a dataset of features and label & $(x,y)\in(\X,\Y)$ \\
 & $(x,y)$ & a sample of features and label &  \\
 & $\mathcal{C}$ & set of indices of causal features &  \\
 & $\bar{\mathcal{C}}$ & the complement set of $\mathcal{C}$ &  \\
 & $\epsilon$ & exogenous variables acting as noise in the generative models &  \\
 & $(\Z,\emptyset)$ & a dataset of feature set $\Z$ with corresponding labels unavailable & $\Z\subset\mathcal{X}$ \\
 & $z$ & a sample of features & $z\in\Z$ \\ \hline
\multirow{5}{3.5cm}{model is denoted with a small-case letter followed by input placeholder $\cdot$ and a greek letter for its parameters} & $f(\cdot;\theta)$ & a model with parameters $\theta$ & $f(\cdot;\theta): \mathcal{X}\rightarrow \mathcal{Y}$ \\
 & $f_k(\cdot;\theta)$ & first $k$ layers of $f(\cdot;\theta)$ & $\textnormal{dom}(f_0(\cdot;\theta))$ means $\mathcal{X}$ \\
 & $h(\cdot;\phi)$ & a model with parameters $\phi$ & $h(\cdot;\phi): \textnormal{dom}(f_{k+1}(\cdot;\theta))\rightarrow \mathcal{Y}$ \\
 & $g(\cdot;\psi)$ & a model with parameters $\psi$ & see below \\
 & $F(\cdot)$ & a random function &  \\ \hline
\multirow{9}{3.5cm}{random variables are denoted by capitalized letters} & $X$ & \multicolumn{2}{l}{random variable for features} \\
 & $Y$ & \multicolumn{2}{p{5cm}}{random variable for label} \\
 & $C$ & \multicolumn{2}{p{5cm}}{random variable for causal features} \\
 & $\bar{C}$ & \multicolumn{2}{p{9cm}}{random variable for non-causal features} \\
 & $\tilde{C}$ & \multicolumn{2}{p{10cm}}{random variable for non-causal, but statistically related features (confounder)} \\
 & $\mathcal{G}$ & \multicolumn{2}{p{5cm}}{Directed Acyclic Graph} \\
 & $U$ & \multicolumn{2}{l}{exogenous variable in the graph} \\
 & $V$ & \multicolumn{2}{l}{endogenous variable in the graph} \\
 & $\textrm{PA}$ & \multicolumn{2}{l}{parents of a variable in the graph} \\ \hline
\multirow{5}{3.5cm}{values are denoted by small case letters} & $x$ & \multicolumn{2}{l}{value of $X$} \\
 & $y$ & \multicolumn{2}{l}{value of $Y$} \\
 & $c$ & \multicolumn{2}{l}{value of $C$} \\
 & $\bar{c}$ & \multicolumn{2}{l}{value of $\bar{C}$} \\
 & $\tilde{c}$ & \multicolumn{2}{l}{value of $\tilde{C}$} \\ \hline
\multirow{2}{3.5cm}{probability} & $P(X)$ & \multicolumn{2}{p{5cm}}{distribution of a random variable $X$} \\
 & $P(X=x)$ & \multicolumn{2}{p{10cm}}{probability of $X=x$ for discrete random variable $X$, also denoted as $P(x)$} \\ \hline
\multirow{2}{3.5cm}{symbols denoting special conditions} & $A \indep{} ..B$ & \multicolumn{2}{p{10cm}}{$A$ is independent of $B$} \\
 & $\mathcal{G}_{\bar{X}, \underaccent{\bar}{Y}}$ & \multicolumn{2}{p{10cm}}{all arrows out of $Y$ are removed and all the arrows coming to $X$ are removed in the original graph $\mathcal{G}$} \\ \hline
\end{tabular}
\caption{A summary of major notations we use throughout this survey. Due to the limitation of space, we define $g(\cdot;\psi)$ here as $g(\cdot;\psi): \textnormal{dom}(f_{k+1}(\cdot;\theta))\rightarrow \textnormal{dom}(f_{k'}(\cdot;\theta))$, where $k'\leq k$.}
\end{table}

\paragraph{Notations}
Throughout the survey, 
we aim to expand the narrative
with two intertwined main threads: 
one is an intuitive explanation of the 
high-level ideas that can help the readers to quickly 
understand the core innovation from each paper, 
and the other is a formalized discussion 
that aims to offer a rigorous 
delineation of the methods through 
our master equation summarising all the papers
under each section. 

\begin{wrapfigure}{R}{0.3\textwidth}
\footnotesize
\begin{tcolorbox}[colback=white!90!gray, colframe=white!70!black,left=1pt, right=1pt, top=0.5pt, bottom=0.5pt]
\textsf{representation}/\textsf{embedding} In our survey paper, we will use the terms representation and embedding interchangeably, and use both of them to refer to the intermediate results generated by the model, which encode raw data into a more abstract form. 
\end{tcolorbox}
\end{wrapfigure}

Thus, to serve our second goal, 
we first introduce the notations here. 
Due to the nature of our paper, 
we aim to expand the discussion 
from both machine learning perspective
and the causality perspective, 
we will introduce the notations 
used in each domain separately. 
From the machine learning perspectives, 
our notations
mostly serve the purposes to describe how to train
the models with regularizations over the datasets used. 
Thus, 
We will use $(\X,\Y)$ to denote a dataset of $n$ samples, 
with each samples denoted as $(x,y)$. 
We will use $P$ to denote distributions of random variables. 
We will use $\ell(\cdot, \cdot)$ to denote a generic loss function, 
and we will use $f(\cdot;\theta)$ to denote 
a function with the parameters $\theta$, 
with the entire hypothesis denoted as $\Theta$. 
Correspondingly, 
we will use $h(\cdot;\phi)$ to denote another model 
that usually plays the role to assist the training 
of $f(\cdot;\theta)$, 
and the input of $h(\cdot;\phi)$ is usually $f_k(\cdot;\theta)$ 
which means the $k$\textsuperscript{th} layer's output 
(i.e., the \textsf{representation}/\textsf{embedding})
from $f(\cdot;\theta)$;
in addition, we will use $g(\cdot;\psi)$ to denote 
a model that usually plays the role of generating data or internal representations, 
which is a function whose output space 
is the same as $f_k(\cdot;\theta)$, 
and we use $f_0(\cdot;\theta)$ to denote the input (data) of $f(\cdot;\theta)$. 

On the other hand, 
when we discuss in the context of causality topics, 
we will use capitalized letters to denote the variables, 
where some variables are reserved for special meanings. 
For example, 
$X$ is reserved for variables corresponding to input features, 
$Y$ for variables corresponding to labels,
$C$ for variables corresponding to causal features,
and therefore, 
$\bar{C}$ will be reserved for variables corresponding to non-causal features. 
Also, within non-causal features, 
there are also features that are statistically correlated with the label
(i.e., confounding factors), 
and we will use $\tilde{C}$ to denote them. 
We will use small letters 
such as $x, y, c, \bar{c}, \tilde{c}$ to denote the values of these random variables. 
As a simple example to show how some of these notations can be used, we denote the standard empirical risk minimization (ERM) as the following
\begin{align}
    \argmin_{\theta}\dfrac{1}{n}\sum_{(x,y)\in(\X,\Y)}l(f(x;\theta),y)
    \label{eq:erm}
\end{align}

\paragraph{Definition of Trustworthy and the Role of Stakeholders}
Before we dive deeper into the introduction of the technical contents of the paper, 
we need to first clarify the definition of trustworthy. 
Different communities might have different 
understanding of the concept. 
In this survey, we use the term trustworthy machine learning
to refer to the development, deployment, and use of machine learning models that are reliable, ethical, and transparent, 
and thus a trustworthy machine learning model is designed to be fair, accurate, and robust, and it is developed using transparent and explainable methods, as well as with security and privacy in mind.

Therefore, we consider trustworthy ML 
as an umbrella term to cover various aspects of machine learning such as fairness, security, privacy, explainability, and robustness. However, as we discussed previously, 
this survey will only discuss the topics of robustness, security (adversarial robustness),  fairness, and explainability. 

It is also worth noting the role of the additional elements, other than major components that are in ERM machine learning study (i.e., the statistical model and the data), plays in the definition of trustworthy ML. 

For example, machine learning robustness studies the topic of how to maintain the predictive performances of variations of the data distribution shifts, usually in the form of additional datasets that users consider similar to the training dataset, or perturbations that users consider should not alter the model's prediction. 
In other words, while the topic studies performance against variations of data, the variation needs to be specified, instead of being arbitrary. 
To put simply, \emph{robustness study must specify what the model is robust against}. 

\begin{figure}
    \centering
    \includegraphics[width=1.0\textwidth]{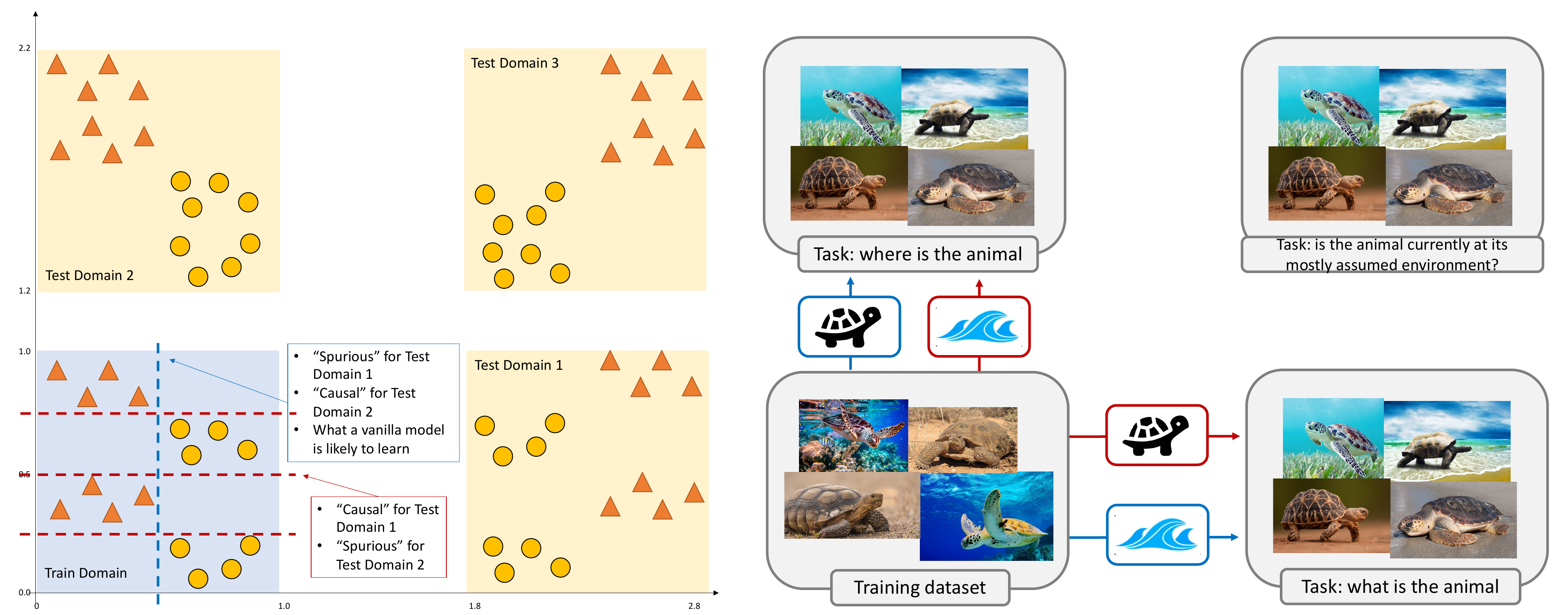}
    \caption{Illustrations to support the argument of in the main text that \emph{robustness study must specify what the model is robust against}. Left: examples created combining the examples used in \cite{wang2022toward2} and \cite{pmlr-v97-zhao19a} for a classification over triangles vs. circles; the labeling functions (decision boundaries) in the training domain are colored according to Test Domain 1 (the bottom right domain). 
    Right: intuitive examples created to illustrate the point of the left.}
    \label{fig:intro:example}
\end{figure}

To further illustrate this point, Figure~\ref{fig:intro:example} (left) is created from the inspiration of two previous works \cite{wang2022toward2,pmlr-v97-zhao19a} discussing machine learning robustness in the context of domain adaptation. 
In our example, the model can learn two possible 
decision boundaries from the data to classify triangles vs. circles, 
and which one is considered useful (or ``causal'') 
depends on where/what the model is used for.  
Figure~\ref{fig:intro:example} (right) 
is an intuitive illustration to explain the example 
on the left. 

Similarly, \emph{security study must specify against what, and sometimes to what degree}. 
Also, 
\emph{interpretability study must specify 
to what is considered interpretable to the stakeholders}, 
For example, as we will see later in detail, 
different assumptions in interpretations, 
such as ``the smaller set of features identified, the better''
or ``the more connected the features identified  (smoothness), the better'' will lead to distinct evaluations of interpretation methods, as well as distinct regularizations as part of the methods. 

In summary, with the above argument, 
we believe that trustworthy machine learning, 
as least within the scope of this survey that 
covers concrete directions such as 
robustness, security, fairness, 
and interpretability, 
is a topic that cannot be studied without specifications 
of the requests of stakeholders. 
As a result, the design of the methods will certainly 
require additional knowledge from such requests 
as part of the regularization or data augmentation procedure. 
We will see tons of evidence to support this argument 
later in this survey. Occasionally, there might be methods that do not use such prior knowledge 
explicitly, but we notice that these methods usually implicitly build upon different assumptions regarding the requests of stakeholders. 
Thus, 
\emph{Trustworthy machine learning cannot be studied without prior knowledge}. 
There might be communities who do not agree with our assertion, then it is most likely because the community defines trustworthy ML differently from ours.

\paragraph{Contributions}
Due to the extensive body of literature surrounding the topic of trustworthy machine learning, our survey diverges from the structure of traditional survey papers. While conventional survey papers tend to passively document various methods within each category, we take a more proactive approach by delving into the statistical underpinnings of each technique and summarizing their overarching design principles. This consolidation allows us to group multiple methods under a unified framework, facilitating a broader analysis and revealing that numerous methods within different trustworthy ML topics share a common rationale. In summary, our contributions can be outlined as follows:
\begin{itemize}
    \item We extensively surveyed papers under the umbrella of trustworthy ML, including robustness, security (adversarial robustness), fairness, and interpretability. We primarily focus on the technical aspects of the methods, and from a data-centric perspective. 
    \item We summarized each method down to its statistical backbone with its conceptual rationale, offering a principled understanding of trustworthy ML. The principled understanding allows us to build high-level connections between methods within and across topics under the umbrella of trustworthy ML, offering readers an efficient way to navigate through the vast sea of this field.  
    \item We connected the principled understanding to the well-established study of causality under Pearl's perspective, showing the connections between trustworthy ML and causality. Many trustworthy ML methods have been implicitly using the popular concepts in causality, suggesting new views for trustworthy ML by leveraging frontiers from causality. 
    \item We put our discussions in the context of large pretrained models. By summarizing the core ideas of the techniques of the large pretrained model's paradigm and connecting them to ERM. 
    With this summary and the principled understanding in previous sections, 
    we are able to potentially predict some of the future methods in the context of the large pretrained models over trustworthy machine learning topics. 
    \item We also offer intuitive languages to explain the ideas behind our mathematical works. 
    \item We also attempt to offer suggestions for the future development of trustworthy machine learning. 
\end{itemize}

\section{Trustworthy Machine Learning Topics from Data Perspective}
\label{sec:robustness}
With the significant results machine learning achieves on benchmark datasets 
on various application scenarios in the lab, 
the community is excited about extending its power into real-world applications. 
However, when it comes to the real world, 
the numerical performance (such as prediction accuracy) 
is not always the only important thing, 
especially when the real data are not as well prepared as the benchmark datasets. 
As a result, many other metrics of machine learning models are valued and studied.

This survey paper scopes its focus along three dimensions of these other valued metrics, namely the model's robustness, fairness, and interpretability, 
and the term ``trustworthy'' is used to refer to a model being robust, fair, and interpretable. 
Although in other literature, 
the term ``trustworthy'' might be used to refer to 
the models being resilient to label noises (label shift) \cite{storkey2009training, zhang2013domain, lipton2018detecting}, 
or private \cite{Abadi_2016,9044259}.  
we do not consider these or other merits 
part of the discussions in this survey paper. 

The remainder of this section is structured in a way that
we will first discuss each of these three major focuses in 
trustworthy machine learning, 
with detailed setup of the problems and high-level descriptions of the solutions
in each of the focuses. 
Then, with our summary of the problems and the solutions, 
we propose a hypothesis that 
an underlying common issue of all these challenges is that
these models are not learning what the models are expected to learn
\textit{i.e.}, the models are not learning the desired features.

\subsection{Robustness}
In modern machine learning communities, 
robustness is usually used to refer to the property 
that a model can maintain its performances over perturbed test data
when the perturbed data
has some ``tolerable'' shifts from the original data
and the causal features remain intact during the perturbation. 

\paragraph{Domain Adaptation}
The study of machine learning robustness in this regard has a long history. 
\emph{Domain adaptation} \citep{ben2007analysis,ben2010theory}, 
as one of the pioneers, 
studies the problem of how to maintain the model's predictive performances 
when the test are from a \textsf{domain} that is similar but different from the \textsf{domain} used to train the models. 

\begin{wrapfigure}{r}{0.5\textwidth}
\footnotesize
\begin{tcolorbox}[colback=white!90!gray, colframe=white!70!black,left=1pt, right=1pt, top=0.5pt, bottom=0.5pt]
\textsf{domain}: in our survey, we follow the convention of using the word ``domain'' to refer to a specific context or environment from which the data comes or in which the model is applied. In statistical studies, a domain is usually associated with a specific data distribution; in practical studies, a domain is usually considered a specific collection of data. \\
\textsf{source domain}: the domain with which the model is trained. \\
\textsf{target domain}: the domain with which the model is tested. 
\end{tcolorbox}
\end{wrapfigure}

The study of domain adaptation has inspired a long line of research
in both theoretical perspectives \citep{ben2007analysis,ben2010theory,MansourMR09,GermainHLM16,ZhangLLJ19,dhouib2020margin}
and empirical perspectives \citep{saenko2010da,kim2020learning,tzeng2017adversarial,peng2018zero}. 
The empirical evaluation is usually set upon a test scenario 
that the model is trained with one dataset from one distribution 
and evaluated from another dataset from another distribution
that is considered similar but different to the training one. 
This ``similar but different'' property is usually defined by human factors 
during the collection of the datasets used for domain adaptation study. 

For instance, under the setup of Example~\ref{exp1}, 
one domain adaptation study is to train 
the image classification model on photos of sea turtles vs. tortoise, 
and then test the model on sketches of the depicted animals 
without background. 
A ideal model is expected to perform well on sketches of animals 
even if it was trained on photos because 
an ideal model is supposed to be able to capture the causal features of depicted animals from photos, 
just as how a human will recognize the images in either photos or sketches because a human understands the true differences between a dog and a cat. 
\emph{Covariate shift} \citep{gretton2009covariate} is a formalization of this problem setup of domain adaptation. 

The theoretical discussion of domain adaptation 
has been expanded over decades \citep{ben2007analysis,ben2010theory,MansourMR09,GermainHLM16,ZhangLLJ19,dhouib2020margin}, 
probably pioneered by \citep{ben2007analysis,ben2010theory}. 
Recent works such as \cite{dhouib2020margin,wang2021toward} conceptually summarized the 
main idea of the generalization error bounds into 
two components 
(an estimatable term regarding the divergence between the source and target distributions, and an non-estimable term about the nature of the problem). 

Thus, 
most of the empirical methods devoted to this topic seeks to 
improve the models' performances by introducing regularizations 
to force the learned representations to be invariant across training and testing distributions, 
with the pioneering example of domain adversarial neural network (DANN) \citep{ajakan2014domain}, which introduces 
a ``domain classifier'' to differentiate two domains
at the representation/embedding space, 
and then a representation that 
offers minimum information to
this 
``domain classifier'' 
(\textit{i.e.} a representation invariant across domains)
is considered good for 
domain adaptation. 
Inspired by the theoretical discussions above, 
mostly from \cite{ben2007analysis,ben2010theory}, 
domain adversarial neural network has popularized the following 
formulation of training a model 
\begin{align}
    \argmin_{\theta} 
    \dfrac{1}{n}\sum_{(x,y)\in(\X,\Y)\cup
    (\Z,\emptyset)}l(f(x;\theta), y)
    - \lambda l(h(f_k(x;\theta);\phi), d), 
    \label{eq:domain_adaptation}
\end{align}
from which the parameters $\phi$ are obtained from 
\begin{align}
    \argmin_{\phi}\sum_{(x,y)\in(\X,\Y)\cup (\Z,\emptyset)}l(h(f_k(x;\theta);\phi), d),
    \label{eq:domain_adaptation:side}
\end{align}
where $(\Z,\emptyset)$ denotes the dataset from \textsf{target} distribution
with $\Z$ denoting features and $\emptyset$ denoting the unavailable labels, 
and $d$ denotes ``domain IDs'', an label-functioning variable that encodes 
the information of whether $x$ is from the \textsf{source} domain 
or \textsf{target} domain. 

In summary, we refer to the equation set of Equation~\ref{eq:domain_adaptation} and Equation~\ref{eq:domain_adaptation:side} as a DANN structure solution. This name is chosen following the fact that domain adversarial neural network (DANN) \citep{ajakan2014domain} is one of the most popular techniques using this set of equations.

\begin{wrapfigure}{r}{0.3\textwidth}
\footnotesize
\begin{tcolorbox}[colback=white!90!gray, colframe=white!70!black,left=1pt, right=1pt, top=0.5pt, bottom=0.5pt]
\textsf{activition}: the output of a node in a neural network, calculated by applying an activation function to its inputs.
\end{tcolorbox}
\end{wrapfigure}

The community later proliferative
progressed along this line 
to introduce many methods for the invariance across
domains/distributions, 
with extensions of the 
representation-learning-model (\textit{i.e.}, encoder)
to two copies \cite{tzeng2017adversarial}, 
extensions with additional alignment of 
\textsf{activation}
distributions 
between source and target domains
\cite{tzeng2015simultaneous}, 
extensions through additional domain-relevant
but task-irrelevant data \cite{peng2018zero}, 
and many others that target learning invariant 
representations
across domains \textit{e.g.}, \cite{bousmalis2016domain,zhuang2015supervised,kim2017learning}.

Another branch of domain adaptation techniques aim to introduce the invariance 
across training domain and test domain in a more explicit manner
by directly augmenting the training data to match the marginal distributions 
of the test data. 
For example, 
a popular approach is 
to generate the data with source domain semantics (i.e., $p(Y \vert X)$) but target domain style (i.e., $p(X)$), 
with a simple master equation
\begin{align}
    \argmin_{\theta} 
    \dfrac{1}{n}\sum_{(x,y)\in(\X,\Y)\cup (\Z,\emptyset)}\mathbb{E}_{z = g(h(x;\phi);\psi)}l(f(z;\theta), y),
    \label{eq:data_aug}
\end{align}
where we use $g(h(\cdot;\phi);\psi)$ to denote a data generation model, where $h(\cdot;\phi)$ maps the raw sample into embedding space, and $g(\cdot;\psi)$ maps the embedding space back to the data space. 

A mainstream choice is to use GAN or its variants as $g(h(\cdot;\phi);\psi)$ to generate the data to boost the performance for domain adaptation, such as \cite{hoffman2018cycada, mutze2022semi, bousmalis2017unsupervised, kim2017learning}. 

While the above summarizes the main approaches in domain adaptation, 
it is worth mentioning that 
there are also works arguing that 
Equation \eqref{eq:domain_adaptation} will not solve a domain 
adaptation problem sufficiently, 
but
most of these works assume the label shift setting such as \cite{ZCZG19,wu2019domain}, 
and thus they are not in the scope of our discussions.

In addition to the relative fixed train-test distributions split scenario, 
there are also works using intermediate 
domains (distributions) to help the adaptation process, 
usually terms as 
multi-step domain adaptation \citep{tan2015transitive,tan2017distant}
or 
gradual domain adaptation \citep{kumar2020understanding,chen2021gradual,wang2022understanding}. 
For a more dedicated summary and a general primer of domain adaptation, we refer readers to several focused literature reviews \citep{weiss2016survey,csurka2017domain,Wang_2018}.

\begin{figure}
    \centering
    \includegraphics[width=1.0\textwidth]{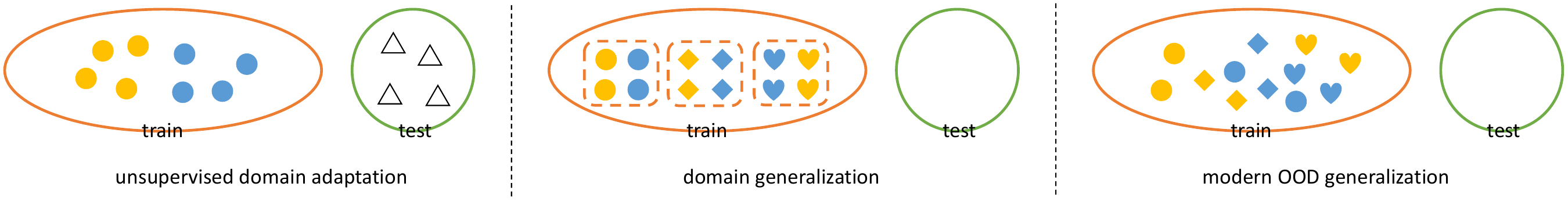}
    \caption{Conceptuall illustration about the differences between unsupervised domain adaptation, domain generalization, and more modern settings of OOD generalizations. The shape denotes the domain of the data, and the color denotes the label. The illustration is for the availability of data during the training time.}
    \label{fig:robustness}
\end{figure}

\paragraph{Domain Generalization}
An often asked question regarding the study of domain adaptation 
is what if we do not know the distribution the model to be tested with
during training.
In reality, this concern seems legitimate since after we build a model, 
we will expect it to perform consistently in future data distributions
that we are still unaware at this moment. 
Thus, as an answer to this question, the community starts to work on 
\emph{domain generalization} \citep{muandet2013domain},
for which a model is trained on a collection of distributions of training data
and then tested on new distributions unseen during training. 

Different from domain adaptation research, 
the development of 
domain generalization techniques in this deep learning era 
rarely build upon pioneering theoretical discussions. 
Instead, 
most of the development efforts can extend upon the 
empirical efforts in the domain adaptation field 
to extend the ``invariance between source vs. target distributions'' technique 
to the more suitable ``invariance among multiple training distributions'' techniques. 
Therefore, a large trunk of empirical 
works 
converge again to the common theme of being invariant 
across multiple distributions, 
despite being creative and innovative as each individual publication, 
such as 
\begin{enumerate}
    \item direct extension of DANN to multi-domain case \citep{li2018domain}  
    and further extensions like conditioning on the label \citep{li2018deep}
    or through divergence terms \citep{zhao2020domain} or others \citep{wang2016select,motiian2017unified,carlucci2018agnostic,akuzawa2019adversarial,ge2021supervised,nguyen2021domain,rahman2021discriminative,han2021learning}, with a master equation
    \begin{align}
    \argmin_{\theta} 
    \dfrac{1}{n}\sum_{(x,y)\in(\X,\Y)}l(f(x;\theta), y)
    - \lambda l(h(f_k(x;\theta);\phi), d), 
    \label{eq:domain_generalization:1}
    \end{align}
    where $d$ stands for the domain ID that is part of the dataset by the definition of domain generalization. Due to the similarity between Equations~\ref{eq:domain_generalization:1} and~\ref{eq:domain_adaptation}, $h(\cdot;\phi)$ can be estimated in the same way as in Equation~\ref{eq:domain_adaptation:side}. 
    \item enforcing invariance with generated corresponding 
samples in other domains \citep{shankar2018generalizing,yue2019domain,gong2019dlow,zhou2020deep,huang2021fsdr}. 
At a high level, the main idea is essentially data augmentation, with
\begin{align}
    \argmin_{\theta} 
    \dfrac{1}{n}\sum_{(x,y)\in (\X,\Y)}\mathbb{E}_{z = g(h(x;\phi);\psi)}l(f(z;\theta), y),
    \label{eq:domain_generalization:2}
\end{align}
which almost the same as Equation~\ref{eq:data_aug}, except for the definition of the marginal distribution one is interested in generating: 
domain adaptation aims to generate data following the marginal distribution of the target domain data, 
while domain generalization aims to generate data following the marginal distributions of other training (source) domain data.
As some concrete examples for domain generalization, \citep{shankar2018generalizing} perturbs data through the gradient with respective to the data 
to fool both the label and the domain classifiers (i.e., an adversarial attack process that we will discuss in the next part), 
and then use the generated data for training (i.e., an adversarial training process that we will discuss in the next part).
\cite{huang2021fsdr} introduced a bidirectional 
learning idea that involves 
the learning in both spatial domain and the frequency domain of an image, 
with domain randomization on the frequency domain as an augmentation. 
\cite{rahman2021discriminative} builds multiple extensions upon \cite{li2018domain} with different blocks for global domains and local sub-domains. 
    \item learning the same classifier across all the 
domains (e.g., invariant risk minimization) \cite{arjovsky2019invariant,ahuja2021invariance},
although with some counterpoints on the effectiveness of this thread \citep{rosenfeld2020risks,kamath2021does}. 
\end{enumerate}

It is also worth mentioning 
another thread of domain generalization works 
following the assumption domain-specific features 
can also help the empirical performances 
in domain generalization \cite{d2018domain,seo2020learning,bui2021exploiting}, 
although this thread is not in the scope of our discussion. 

In recent years, 
although it is fairly intuitive that using the extra domain
partition information 
will benefit empirical performances, 
the community continues to seek to 
free this last constraint of domain generalization 
to a more realistic scenario where the training datasets are not necessarily 
partitioned into multiple distributions/domains with clear boundaries
during training
\cite{wang2019learning2,wang2019learning,wang2020self,tian2022neuron,huang2022two}. 
It seems the community is using the terminology 
\textbf{Out-of-domain Domain (OOD) Generalization}
to largely refer to Domain Generalization. 
For more detailed discussions 
of topics in Domain Generalization 
and Out-of-domain Domain (OOD) Generalization, 
we refer the readers to more dedicated surveys
\cite{wang2022generalizing,shen2021towards}.

\paragraph{Countering Spurious Features (Dataset Bias)}
Another thread of research that usually falls into the 
scope of machine learning robustness 
is motivated by the concept of spurious features \cite{vigen2015spurious}, 
confounding factors \citep{mcdonald2014confounding}, 
or bias-in-data \citep{torralba2011unbiased}. 
Overall, 
in comparison to the topics discussed above, 
this thread of works centers more explicitly around the story in Example~\ref{exp1}
about the fact that the models might learn some undesired features (like backgrounds) other than
the ones that are semantically aligned with the human perception of the data. 

As there are multiple lines of works suggesting that
a fundamental challenge for the model to learn ``semantic'' (or causal) features instead 
of the spurious features lies in the construction of the dataset \cite{jo2017measuring,wang2020high,hermann2020origins}, 
or the existence of the confounding features (Example~\ref{exp1}). 
Therefore, 
most of these methods are designed following the same procedure: 
first identify the spurious features, 
and then analyze the statistical properties of the spurious features to 
build a reguarlization and/or a training process for the 
model to avoid learning these features.

For example, \cite{wang2019learning} investigates the problem that 
a vanilla computer vision model tends to learn the texture features from an image \cite{geirhos2018imagenet,shah2020pitfalls}, 
and build a side model that focuses particularly on learning textures 
and force the main model to learn information invariant to this side model. 
Following the similar main structure, \cite{wang2019learning2} 
studies the problem that sometimes model tend to learn a local patch of information, ignoring the idea from the whole images, 
and
constructs a side model 
that particularly focuses on learning patches of images to help the learning of the main model. 
Further, \cite{bahng2020learning} introduces a more systematic view to counter the bias of data with concrete examples such as 
CNN with smaller receptive fields for texture bias. 
More recently, 
aiming to break the requirement of a prior model design, 
there is a line of works investigating 
the statistical properties of certain datasets, and conclude that, for these datasets, 
the spurious features are usually the ones that are easier to learn. 
Following this observation, 
\cite{nam2020learning}
considers the features 
learnt at an early stage of the training spurious, 
and \cite{dagaev2023too} considers the features learnt by a shallow network spurious, 
then they can take advantage of these properties to counter the main model's learning of the spurious features. 

The above thread naturally converges to a master equation, 
\begin{align}
    \argmin_{\theta} 
    \dfrac{1}{n}\sum_{(x,y)\in (\X,\Y)}l(f(x;\theta), \y)
    - \lambda l(h(f_k(x;\theta);\phi), d), 
    \label{eq:bias}
\end{align}
with $\phi$ to be estimated with
\begin{align}
    \argmin_{\phi \in \Phi}\sum_{(x,y)\in (\X,\Y)}l(h(f_k(x;\theta);\phi), y).
    \label{eq:bias:side}
\end{align}
As one might notice, Equations~\ref{eq:bias} and~\ref{eq:domain_generalization:1} are exactly the same, 
however, the differences lie where Equation~\ref{eq:bias:side} is compared to Equation~\ref{eq:domain_adaptation:side}. 
The main difference lies that, 
where \ref{eq:bias:side} does not require explicit domain labels $d$, 
however, it usually requires dedicated chosen hypothesis space $\Phi$ such as
the models that only learn the texture of images etc. 

There is also some exceptions, for example, 
\cite{kim2019learning} assumes prior knowledge of bias information is available in the form of labels, and then directly reuses the DANN structure from Equation~\ref{eq:domain_adaptation} and~\ref{eq:domain_adaptation:side}. 

\paragraph{Adversarial Robustness}
Another widely studied topic under the robustness category 
is \emph{adversarial robustness}, 
which studies the model's reaction to samples that are transformed under certain criteria. 
The research field is popularized by the ``intriguing properties of neural networks'' \citep{szegedy2013intriguing,goodfellow2015explaining} 
through showcasing that we can generate samples that are 
perceptually indistinguishable to the original samples 
but completely alter the models' decisions. 

\begin{wrapfigure}{r}{0.4\textwidth} 
\vspace{-20pt}
  \begin{center}
    \includegraphics[width=0.4\textwidth]{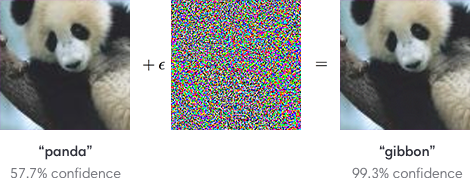}
    \caption{One of the most widely known illustrations about adversarial robustness, from \cite{goodfellow2015explaining}. It tells the story that one can inject (carefully crafted) noises to the image, creating a resultant image that appears identical to the original image, but deceive the model to predict it to be something else.}
    \label{fig:adversarial}
  \end{center}
  \vspace{-20pt}
  \vspace{1pt}
\end{wrapfigure} 

The community names these samples \emph{adversarial samples}, 
the process of generating them \emph{attack}, 
and correspondingly, the process of maintaining 
the model's prediction over these generated samples same as over the original ones \emph{defense}. 
Also, it is critical to note that there is usually a constraint 
regularizing the generation of adversarial sample 
in terms of the distance between the resultant adversarial sample 
and the original sample, 
otherwise, the research will become meaningless 
if the adversarial sample can be arbitrarily different 
from the original sample. 
Usually, we denote such a constraint as 
$d(x',x)\leq\epsilon$, where 
$x'$ is the generated sample and 
$d(\cdot,\cdot)$ is the distance metric of choice, 
with the most popular choices usually being $\ell_p$ norms.

The discovery of the intriguing property of adversarial sample inspires long lines of studies 
to innovation along the attack methods \citep{kurakin2016adversarial,papernot2016limitations,moosavi2016deepfool,carlini2017towards,papernot2016transferability,narodytska2016simple,Rozsa_2016,chen2017zoo,zhao2017generating,baluja2017adversarial,moosavi2017universal,cisse2017houdini,sarkar2017upset,liu2017enhanced,Oh_2017,athalye2018obfuscated,papernot2017practical,hayes2018learning,khrulkov2018art,su2019one,wong2019wasserstein,croce2020reliable, wang2020spanning, andriushchenko2020square, chen2020rays, guo2018low, cheng2019sign, das2022advcodemix, zhang2023transferable, zhuang2023pilot, zhu2023ligaa, li2023sibling, williams2023black}
as well as the defense methods \cite{gu2014deep,papernot2016distillation,cisse2017houdini,Luo2015Foveation,jin2015robust,lee2015manifold,lyu2015unified,sun2017hypernetworks,wang2016learning,wang2016using,wang2017adversary,nguyen2017learning,zantedeschi2017efficient,na2017cascade,kardan2017mitigating,strauss2017ensemble,nayebi2017biologically,krotov2017dense,ross2017improving,liao2018defense,ding2018mma, wang2019improving, carmon2019unlabeled, gowal2020uncovering, wu2020adversarial, pang2020bag, pang2020boosting, yang2020ml, detecting_anomalous_inputs, wang2022convergence, ma2021increasingmargin, ho2022attack, jin2023randomized, fu2023semi, yang2023adversarial, hsiung2023towards, wei2023cfa}. 

\begin{wrapfigure}{r}{0.5\textwidth}
\footnotesize
\begin{tcolorbox}[colback=white!90!gray, colframe=white!70!black,left=1pt, right=1pt, top=0.5pt, bottom=0.5pt]
\textsf{adversarial example}/\textsf{adversarial sample} refers to the data that has been modified slightly in a way that is intended to cause a model to make a mistake.\\ 
\textsf{attack} refers to the process or method used to generate adversarial examples.\\
\textsf{defense} refers to methods to make a model more robust against attacks.
\end{tcolorbox}
\end{wrapfigure}

While the community has progressed significantly along both the attack methods and the defense methods directions, 
the research efforts recently seemingly converge to the most powerful attack methods and its associated defense methods: 
for a while, 
PGD \citep{madry2018towards} is considered as the most powerful attack methods. 
Intuitively speaking, PGD can be considered as an opposite process 
of training a model with the gradient descent:
when we train a model with the gradient descent, 
we usually iteratively update the model parameters following the gradient to decrease the model's loss over the fixed data, 
when we use PGD, we iteratively update the data following the gradient to increase the fixed model's loss over the resultant data. 

With a most powerful attack method, the most powerful defense method is to simply train with the 
adversarial samples generated with the attack at each iteration along training \citep{madry2018towards}, 
which leads to the following equation
\begin{align}
    \argmin_{\theta} 
    \dfrac{1}{n}\sum_{(x,y)\in(\X,\Y)}\max_{x';d(x',x)\leq \epsilon} l(f(x';\theta), y).
    \label{eq:adversarial_defense}
\end{align}

As Equation~\ref{eq:adversarial_defense} once again aligns well with the 
one of the central themes above (e.g., Equation~\ref{eq:domain_generalization:2}), 
one might wonder that whether we can use the other major theme (e.g., Equations~\ref{eq:domain_adaptation} and~\ref{eq:domain_generalization:1})
to improve the method for adversarial robustness. 
In fact, there are indeed some works using, again, the DANN structure 
to improve adversarial robustness by considering the setting 
as a domain adaptation problem \cite{hou2020class,rasheed2022multiple}, 
but these methods do not seem to have shown its significance. 

Interestingly, 
a more significant thread of methods have demonstrated its empirical strength is closely tied 
to the Equations~\ref{eq:domain_adaptation} and~\ref{eq:domain_generalization:1} but in a much more simplified 
manner because of the nature of adversarial examples. 
Since the generation of adversarial examples is essentially a data augmentation process, 
thus there is a natural one-to-one correspondence between the original sample $x$, and the augmented sample $x'$ 
we do not really need a discriminator (i.e., the $h(\cdot;\phi)$) to push for the invariance between the embeddings 
learnt from these two samples, 
we can simply regularize the distance between these two embeddings, 
as the following:
\begin{align}
    \argmin_{\theta} 
    \dfrac{1}{n}\sum_{(x,y)\in(\X,\Y)}\max_{x';d(x',x)\leq \epsilon}\Big(l(f(x';\theta), y)\Big)
    + \lambda D(f(x';\theta), f(x;\theta)), 
    \label{eq:adversarial_defense:2}
\end{align}
where $D(\cdot,\cdot)$ stands for a distance metric of choice over the embeddings of samples fed into the model. 
Equation \ref{eq:adversarial_defense:2} corresponds to 
adversarial training \citep{madry2018towards} when $\lambda=0$, 
TRADES loss \citep{zhang2019theoretically} when $D(\cdot,\cdot)$ is KL divergence,
and ALP loss \citep{kannan2018adversarial} when $D(\cdot,\cdot)$ is squared $\ell_2$ norm. 
Other notable methods under this category are 
MART \cite{Wang2020Improving} and Consistency \cite{tack2021consistency}.

In addition to the development of defense methods for the statistical perspectives,   
the community has also been seeking to offer an intuitive understanding of the underlying causes 
of such ``intriguing properties''. 
One answer is points to the nature of data
by showing the 
the existence of such features that are imperceptible to human
but also predictive \citep{ilyas2019adversarial,wang2020high}. 
The existence of such features have also been validated by multiple other works 
showing that deep learning model has a tendency in learning the texture of images \citep{jo2017measuring,geirhos2018imagenet,wang2020high,hermann2020origins,shah2020pitfalls}, 
not necessarily in the context of adversarial robustness. 
These evidence credits the challenges 
of adversarial robustness to the perspective of data. 

Further, it is worth mentioning that, despite the popularity gained through adversarial robustness in the deep learning community recently, 
the statistical techniques of how to maintain a model's prediction 
toward certain distribution shifts 
equivalent to perturbing features within a $\ell_p$ ball
has been studied over decades in the statistics community 
under the name 
\textbf{Distributional Robustness Optimization} \cite{Rahimian_2022}. 
Some of these studies in recent years interestingly connects 
the regularization in loss terms \citep{shafieezadehabadeh2015distributionally,xu2008robust}
to the adversarial robustness behaviors in $\ell_p$ norms in linear models. 

There are many other related topics in adversarial robustness. 
For example, targeted adversarial attack 
is an extension of adversarial attack, in the sense 
that it does not only use the resultant image 
to deceive the model to predict into something, 
but direct it to predict into a specific class \cite{8294186}. 
Many of the vanilla adversarial attack methods above can be extended to its targeted version, as surveyed by \cite{8294186}. 
Another popular extension is to extend the adversarial training into embedding space \cite{dai2019adversarial,wang2020self}, 
where the above adversarial training idea (e.g., Equation~\ref{eq:adversarial_defense}) is 
used, but, instead of at the raw data level, it is used 
at the embedding/representation level. 
In other words, 
instead of augmenting $x$ into $x'$, 
it augments $f_k(x;\theta)$ into its perturbed counterpart. 

\paragraph{Connections of OOD Robustness and Adversarial Robustness}
While we have shown that the Equation~\ref{eq:adversarial_defense} has the same format with one of the major threads of methods in OOD robustness (such as domain adaptation and domain generalization), 
it is worth mentioning that 
the format of Equation~\ref{eq:adversarial_defense:2} is also not unique. 
It has been studied in the robustness literature 
in the name of consistency loss or alignment regularizations 
\cite{10.1145/3534678.3539438, liang2018learning, asai-hajishirzi-2020-logic, guo2019visual, sajjadi2016regularization, shah2019cycleconsistency, 8982954, 7298880, zhang2018regularizing, zheng2016improving}

These connections inspire us to think in a more high-level of what robustness means: 
robustness refers to the study of whether the model can maintain its performance under the shifts when the stakeholders do not consider these shifts should lead to degradation of the model's performance,
either the shift is more salient such as from color image to sketch (thus OOD generalization) or more subtle such that the stakeholders cannot observe the shift (thus adversarial robustness).

\subsection{Fairness}
When a machine learning model is robust against various shifts, 
it might be perform into the real-world without a noticeable performance drop on various situations. 
However, this does not necessarily mean that the 
machine learning models 
are ready to be deployed to serve all different tasks, 
especially on certain tasks where 
there are some sensitive attributes from the data that 
the models better neglect while building the statistical relationship. 
For example, 
a hiring\cite{employement_test} prediction software is not supposed to use ethnicity or gender as an attribute to avoid 
discrimination against certain populations.  
As another example, 
a face recognition algorithm is supposed to perform stably over 
all genders and skin colors \cite{adeli2019bias, wang2020mitigating, amini2019uncovering, kandge2022biasing, harrisonmitigating}. 

While there are evidence that the machine learning models 
are suffering various challenges above \cite{caton2020fairness,du2020fairness}, 
fortunately, the community is 
actively proposing powerful methods to mitigate these issues, 
and the research community usually refers to this thread of topics 
the study of machine learning fairness. 

While there are multiple topics the ML fairness study focuses, 
the problem are mainly categorized into two problems according to \cite{du2020fairness}:
the \emph{outcome discrimination}
and the \emph{quality disparity}.
\begin{itemize}
    \item Outcome Discrimination: it refers to the scenario 
that the ML model uses certain attributes 
to predict, such as 
learning the association between the ethnicity and the salary outcome
    \item Quality Disparity: it refers to the scenario 
that the ML model fails to generalize to samples 
with certain properties because of their lack of representation in the data, 
e.g., a model trained on Caucasian faces might not perform well on Asian faces. 
\end{itemize}

One might already notice that, 
the \emph{outcome discrimination} problem corresponds 
to the \emph{spurious feature} setup that we discussed in the 
robustness section, although the techniques largely use the \emph{domain adaptation} or \emph{domain generalization} ideas with the availability of the sensitive variables; 
while the \emph{quality disparity} is more conceptually related to the 
\emph{domain adaptation} or \emph{domain generalization} topics, but because the distributions are more explicitly defined here with sample disparity, 
there are usually more direct methods to align the distributions than a mathematical alignment that is often seen in the domain adaptation/generalization works. 

\paragraph{Outcome Discrimination}
Due to the similarity of the mathematical construction of the problem, the many solutions can also be categorized into 
what has been used in the above, such as
\begin{align}
    \argmin_{\theta} 
    \dfrac{1}{n}\sum_{(x,y)\in (\X,\Y)}l(f(x;\theta), y)
    - \lambda l(h(f_k(x;\theta);\phi), d), 
    \label{eq:fairness:1}
\end{align}
where $d$ now denotes the label of sensitive variable, 
with $\phi$ to be estimated with in the same way as in Equation~\ref{eq:domain_adaptation:side}.

\begin{wrapfigure}{r}{0.5\textwidth}
\footnotesize
\vspace{-3mm}
\begin{tcolorbox}[colback=white!90!gray, colframe=white!70!black,left=1pt, right=1pt, top=0.5pt, bottom=0.5pt]
\textsf{GAN-style} We use this term to broadly refer to modules that can generate samples like a GAN. This survey does not intend to analyze the detailed differences among data-generation models used. 
\end{tcolorbox}
\end{wrapfigure}
For example, as early-stage works, 
\citep{adel2019one,celis2019improved,edwards2015censoring,wadsworth2018achieving} mostly reuses DANN model with domain id replaced with sensitive attribute, 
\citep{beutel2017data} adopts a similar idea, but only uses a small amount of data. 
Further, 
\citep{madras2018learning,feng2019learning} uses \textsf{GAN-style} model, which is then extended by 
\cite{xu2019achieving} with the idea to use multiple GANs. 
\cite{zhang2018mitigating} builds the adversarial component from prediction of the main model to the sensitive variable.
Further, 
\citep{dwork2018decoupled} trains a classifier for each group, and then use domain adaptation to connects each classifier of the group. 
\citep{oneto2019taking} builds a model to predict sensitive attribute, and then augment the data to remove the part of the features that can predict the sensitive attribute.  
\cite{zemel2013learning} maintains the insensitivity to the sensitive variable through reconstruction. 

Another branch of efforts is to train with explicit fairness constraints, 
along which, 
\cite{celis2019classification} introduces a framework that connects to multiple existing fairness definitions and handle multiple existing fairness constraints. 
\cite{cotter2019optimization} introduces a 
proxy-Lagrangian formulation for optimizing non-convex objectives with
non-differentiable constraints that also connects to multiple fairness and other policy definitions. 
Along this thread, there is also a proliferation 
of methods focusing on explicitly building constraints 
into the process of learning such as 
\cite{goh2016satisfying,narasimhan2018learning,aghaei2019learning,berk2017convex,di2020counterfactual,huang2019stable}.

\paragraph{Quality Disparity}
On the other hand, 
quality disparity, due to its explicit construction 
of problem in terms of the model's lack of attention to samples that are not sufficiently represented in the data, 
the corresponding techniques usually directly weigh the samples to push the models to emphasize the minority samples.

For example, there is a major line of solutions focusing on weighting the samples differently, 
for example, 
\citep{goel2018non} increase the weight of the minority samples, 
\citep{jiang2020identifying} sample-weighting method that corresponds to multiple different fairness measures, 
\cite{krasanakis2018adaptive} introduces a process that iteratively adapts training sample weights. 
There is also a line of research working on the similar issue 
under a ``group-DRO'' term, 
such as \citep{hu2018does, nam2020learning, Sagawa2020Distributionally}

The sampling-weighting theme can also be potentially aligned to the central theme of this paper, as discussed in \cite{wang2022toward2}, 
although the discussion involves many more assumptions. 

There is also a branch of papers that is leveraging the 
domain invariance techniques to solve the problems of Quality Disparity, 
essentially, to align the distributional differences 
between the training distributions (where there is low density for the minority samples) and the testing distributions.
For example, 
\citep{wang2019repairing} generates data distribution that minimizes the disparity (generate target domain data, of data distribution that minimizes the source/target domain differences) 
\cite{jiang2020wasserstein} aligns the two distributions with Wasserstein distance. Therefore, once again, these methods 
can lead to the same central equation as used in Domain Adaptation \eqref{eq:domain_adaptation}. 

\paragraph{Evaluation of Fairness}
While we are presenting a summary of the techniques 
that have been invented and proved useful in various topics 
under the ``machine learning fairness'' category, 
it is worth noting that 
we believe the research about machine learning fairness 
involves many directions other than 
the development of machine learning methods. 

For example, 
one crucial topic is probably the actual meaning of being ``fair'', 
which has inspired multiple lines of discussions 
on either the societal aspects of ``fairness'' 
or the evaluation metrics of ``fairness''. 
As a technical survey, 
we do not intend to offer discussions 
on these aspects, 
readers of interest can refer to more dedicated 
surveys of relevant discussions \citep{mehrabi2021survey,caton2020fairness}

\subsection{XAI: Interpretability and Explainability}

\paragraph{Definitions and Evaluations} 
Another big branch of the study of trustworthy machine learning is the investigation of the techniques
to unveil the blackbox nature of the deep learning models, 
aiming to explain the working mechanisms of the stacked layers of matrices to the users 
with human comprehensible terms. 
This field is often described as to study the \emph{interpretability}, \emph{explanability}, or even \emph{understandability} of the neural networks, 
with subtle differences in the definitions of each 
\cite{lipton2018mythos,DoshiVelez2017TowardsAR,barredo2020xai}.
Here in this paper, we will not dive deep to analyze the exact definitions 
of each, but to follow some other customs \cite{zhang2021survey} 
to use these terms exchangeably: 
we use these terms to describe the study of techniques that report 
a set of features the models use to make the predictions, 
which corresponds to the data perspective theme of our survey. 
On the other hand, the branch of works aiming to explain 
how the building blocks of matrices are wired together for a model 
to make predictions is not in the scope of our discussion. 

The diverse set of the definitions leads to a diverse set of evaluation metrics. 
As one might expect, 
one of the ideal evaluations in terms of performance is to test 
whether the interpreted results (i.e., the features identified by the interpretability methods) can directly speak to the users (domain experts)
in a comprehensible manner 
\cite{DoshiVelez2017TowardsAR}.
However, this evaluation is probably also the least favorable choice in terms of efficiency
as it involves human evaluation (i.e., surveying users to vote out a rank of methods).

As alternatives, there is a list of other evaluation methods introduced to evaluate the interpretability methods
by quantitatively measuring certain properties of the identified features. 
For example, 
one branch is to masking out the identified features 
and then test for the model's performance degradation 
of the same model 
\cite{samek2017evaluating}
or retrained models 
\cite{julius2018sanity,hooker2019benchmark}.
There are also other evaluation methods that emphasize on other properties, 
for example, 
\cite{DoshiVelez2017TowardsAR}
emphasizes the sparsity of identified features, 
while  
\cite{Montavon_2018}
emphasizes the 
the smoothness of the identified features. 

\paragraph{Methods in Interpretability and Explanability}
In a nutshell, we notice that different evaluation methods can directly inspire 
the design of methods. 
For example, many methods directly support the 
formula of a ``main equation'' for interpretability (identification of features) 
regularized with a constraint, 
where the constraint 
can regularize the identified features 
to be sparse, smooth, etc, 
dependent on the evaluation metrics. 
As one might expect, 
the ``main equation'' once again converges to a central theme 
that we will present after we discuss the techniques in details
in the following paragraphs. 

We will start by iterating the argument in \cite{dhurandhar2018explanations} 
about the features being ``minimally and sufficiently present'' 
and ``minimally and necessarily absent''. 
In short, they search for the features that
if not perturbed, the prediction will not change, 
and if perturbed, the prediction will change.
While there are multiple other efforts to define the importance of features 
in a similar manner \cite{covert2020understanding, aas2021explaining, lundberg2017unified, goyal2019counterfactual},
as one may expect, 
such definitions will directly guide a golden 
strategy of locating such features for model explanation: 
perturbing the features of interest
and then compare the model's output under certain metrics of interest
(such as whether the prediction shifts). 

This main idea of perturbing features and then comparing the output 
for explanation has been considered as 
a central theme of model explainability or interpretability by \cite{covert2021explaining}, 
which summarizes multiple relevant techniques 
such as IME \cite{vstrumbelj2009explaining,strumbelj2010efficient},
SHAP \cite{lundberg2017unified}, 
SAGE \cite{covert2020understanding}, 
LIME \cite{ribeiro2016should}, and many others. 
One can refer to detailed discussions \cite{covert2021explaining} for a summary of more methods on this theme. 
However, one shall notice that summaries like \cite{covert2021explaining}
considers the first step of the explanation routine as ``removing'' of features, 
whereas here we refer to ``perturbation'', which we consider is more general, and
fits the central theme of our entire survey better. 

One difference between ``removal'' and ``perturbation'' is that 
``removal'' fixes the feature values one can use, usually to be zeros \cite{zeiler2014visualizing,schwab2019cxplain,petsiuk2018rise}, 
default values \cite{ribeiro2016should,dabkowski2017real}, 
or certain values according to the 
(conditional) marginal distribution \cite{lundberg2017unified,covert2020understanding,zintgraf2017visualizing};
meanwhile, 
``perturbation'' does not have clear values to set, thus allowing more methods to be categorized into this theme. 

For example, 
activation maximization \cite{erhan2009visualizing,simonyan2013deep}
perturbs the features to maximize the output of a model 
(e.g., the activation of a certain class of the prediction layer)
to search for patterns of input that are most responsible for the class. 
In practice, it can also be implemented in a way that
the users start from an existing image of a certain class
and apply the perturbation to convert the image to maximize the activation of another class \cite{engstrom2019learning}, 
so that the patterns that are responsible for the prediction will be more visually recognized. 
This usage in practice will probably remind the readers of 
the \textsf{adversarial attack} methods discussed in previous sections, 
which essentially is about perturbing the features of the data
to alter the prediction of the model. 
However, adversarial attack constraints the perturbation to be invariant to a human's perception (usually favors high-frequency perturbations), 
while interpretation methods usually constrain the perturbation to be meaningful to a human's perception (usually favors low-frequency perturbations) \cite{mahendran2015understanding,yosinski2015understanding,nguyen2016synthesizing}. 

Finally, the activation maximization usually uses the gradient information of the model 
to perform the perturbations, which seems a natural idea given the popular connections between the gradient 
and its input, as well as the important role the gradient has played in adversarial attack methods. 
Further, it is worth mentioning that the usage of gradient has played a significant role 
along in the thread of model interpretation, 
known as gradient-attribution methods.
Methods such as GradCam \cite{selvaraju2017grad,zhou2016learning} have been widely 
used by the community. 
Other works \cite{zhou2016learning,zhang2021survey} have also categorized other popular methods 
such as DeepLIFT \cite{shrikumar2017learning}, LRP \cite{bach2015pixel}, and integrated gradient \cite{sundararajan2017axiomatic}
as gradient-based methods. 
In addition, we believe the connection between perturbation (or removal) based methods and gradient-attribution methods is probably more mathematically fundamental: 
the definition of ``gradient'' is the evaluation of the function output of the infinitesimal shift (i.e., perturbation) of the input (i.e., features).  

In addition, it seems natural that the perturbation (or removal) based methods can be accelerated by the gradient-attribution methods. However, we do not see published papers that explicitly connnect these two threads. 

Overall, as a summary of the techniques discussed above, we aim to attempt a master equation that outlines the techniques 
into one equation: 
\begin{align}
    x^* = \argmax_{x': d(x',x)\leq \epsilon} e(f(x',\theta))
    \label{eq:xai}
\end{align}
where $x^*$ denotes the explanation of the input $x$, $e$ represents the evaluation function discussed above (such as change of the prediction or negative activation function), $d$ represents the constraint discussed above (usually those favoring low-frequency components), 
and the choice of features and the target values. 

Different from the above master equations, 
Equation \eqref{eq:xai} does not seem to offer an elegant enough mathematical guidance for the methods in this section, in comparison to the ones in previous section. 
For example, while both equations \eqref{eq:adversarial_defense} and \eqref{eq:xai} use $d(x',x)\leq \epsilon$, it will take more mathematical efforts to correspond the constrain to each method in this section, while it barely requires additional efforts to correspond it to most methods in the adversarial robustness section. 
Regardless, the master equation should still offer an adequate conceptual summary of most of the methods, 
which will be enough for us to continue to the next part of this survey. 
Along the preparation of this manuscript, we also notice concurrent works that connect interpretability and adversarial robustness 
from more technical perspectives \cite{liu2020adversarial}.

\paragraph{Connections Between Interpretability and Robustenss}
Despite the expansive set of promising techniques 
that aims to continue to improve the explanation techniques, 
it is worth mentioning that 
many of these explanation techniques are fairly easy to be fooled. 
For example, 
\cite{heo2019fooling} fine-tunes the model 
with additional regularizations to shift the attention maps, 
and \cite{dombrowski2019explanations} leverages the gradient of model with respect to the image
to perturb the image features to manipulate the explanation, 
which is a process highly relevant to (targeted) adversarial attack. 
One can refer to a more systematic discussion on this regard \cite{vadillo2023fool}. 

However, the above discussion of the possibility of fooling an 
interpretation method leads to another question:
whether it is the issue of the interpretability method
or the issue of the model itself. 
In fact, there is a long line of methods 
that has been discussing whether 
a robust model that has been trained on the right features
naturally has multiple desired properties. 
For example, 
\cite{engstrom2019adversarial} showed that an adversarially robust vision model has a chance to perform well on a variety of different vision tasks by learning a representation that is better 
aligned with the human visual system. 
The merit of adversarial robust models on learning a representation that is more aligned with the preferences of the stakeholders has been supported directly or indirectly by many works of different nature, such as \cite{zhang2019interpreting,wang2020high, chen2022adversarial, yi2021improved, sadria2023adversarial, terzi2020adversarial}

Despite a long line of work suggesting that 
a more robust model tends to behave better with the stakeholders, 
the conclusion of whether a robust model is enough is not clear at this moment. 
This line of debate nonetheless validates one point:
a robust model is more favorable than a vanilla model, 
although it might not be ideal enough, 
which we conjecture is because the 
robust models are not robust enough yet. 

\begin{figure}
    \centering
    \includegraphics[width=0.9\textwidth]{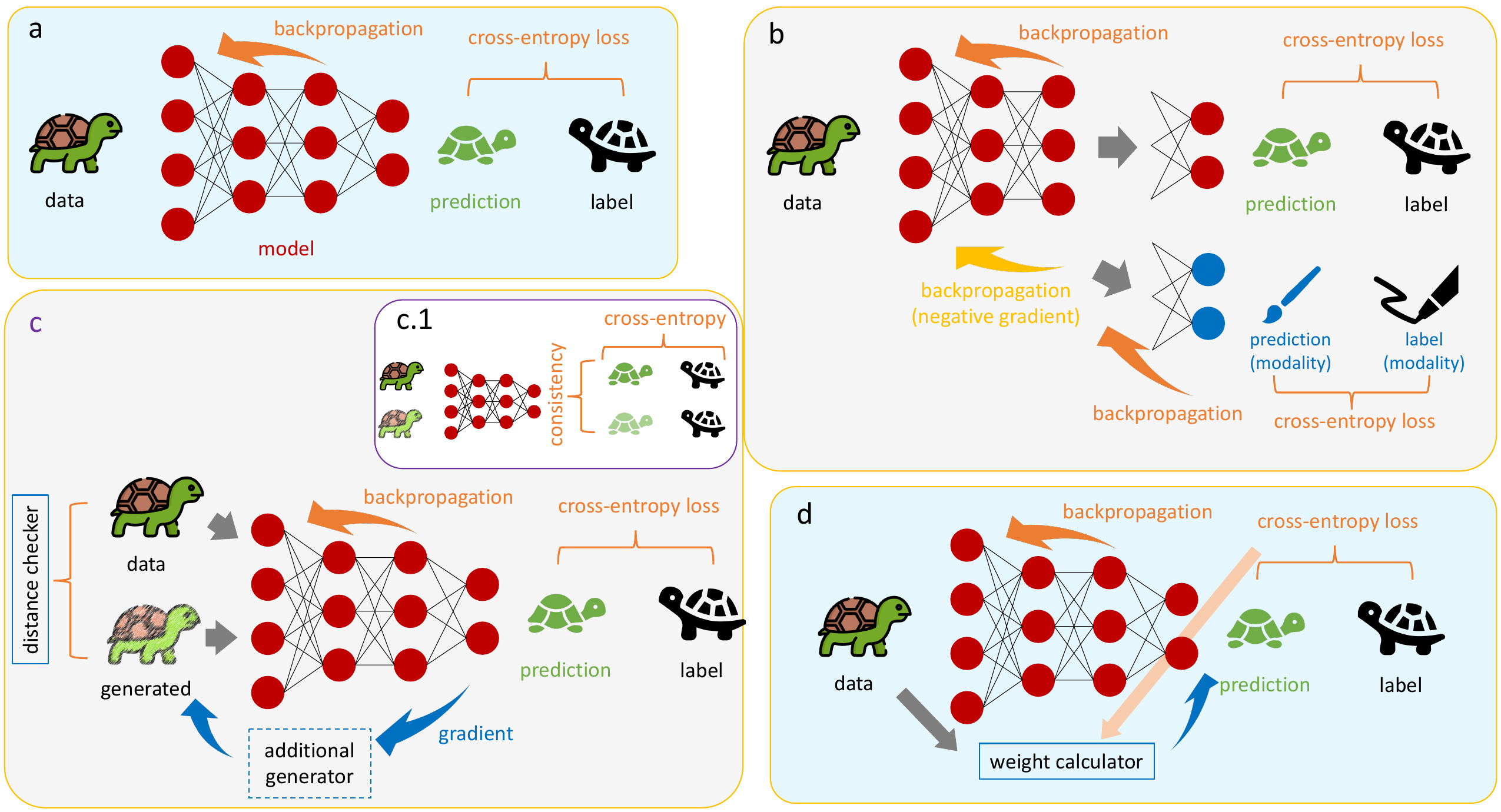}
    \caption{A summary of methods in our converged theme of trustworthy machine learning. (a) standard ERM loss. (b) DANN structure construction of model, corresponding to master equation~\ref{eq:master:1}. (c) worst-case data augmentation strategy, corresponding to master equation~\ref{eq:master:2}. (c.1) shows an upgraded version of the loss design of (c) that usually leads to an improved empirical performance. (d) is sample-reweighting method that corresponds to master equation~\ref{eq:master:3}, and it can be plugged onto all previous methods.}
    \label{fig:method}
\end{figure}

\subsection{A Theme of Trustworthy Machine Learning from Data Perspective} 
\label{subsec:master_eq}
With separate discussions of multiple threads of different 
topics in trustworthy machine learning
and the mathematical and conceptual summarization of the main ideas, 
we hope we have convinced our readers 
that
many of the methods discussed here, 
although innovative and powerful in other aspects, 
converge to an interesting theme of trustworthy machine learning. 

As we can see, two significant formulations repeatedly appear in the discussion 
of methods across different aspects of trustworthy machine learning topics: 
the first one, probably popularized by the domain adversarial neural network, 
is 
\begin{align} \label{eq:master:1}
    \argmin_{\theta} 
    \dfrac{1}{n}\sum_{(x,y)\in(\X,\Y)}l(f(x;\theta), y)
    - \lambda l(h(f_k(x;\theta);\phi), d), 
\end{align}
where choices of $h(\cdot;\phi)$ and $d$ depend 
on the exact applications, 
as we discussed above;
the second one, probably popularized by adversarial training in adversarial robustness literature, 
is 
\begin{align} \label{eq:master:2}
    \argmin_{\theta} 
    \dfrac{1}{n}\sum_{(x,y)\in(\X,\Y)}\max_{x';d(x',x)\leq \epsilon} l(f(x';\theta), y),
\end{align}
where choices of $d(\cdot,\cdot)$ and $\epsilon$ will depend 
on the exact applications. 
One might consider the generation of $x'$ will also vary and depend
on the applications, 
however, in our formulation, 
we consider that when $d(\cdot,\cdot)$ and $\epsilon$ are well defined, 
the generation of $x'$ under a maximization will also be determined. 

In addition, there is also a plug-in component that one can directly inject onto Equations~\ref{eq:master:1} and~\ref{eq:master:2} to weigh the samples differently. 
For example, we can denote the weighting factor as
$\alpha(x, y,\theta)$, and the standard ERM loss function Equation~\ref{eq:erm} can be directly upgraded to the following
\begin{align}\label{eq:master:3}
    \argmin_\theta \dfrac{1}{n}\sum_{(x,y)\in(\X,\Y)}\alpha(x,y,\theta)l(f(x;\theta),y),
\end{align}
the same technique can be directly plugged onto Equations~\ref{eq:master:1} and~\ref{eq:master:2}. 

Further, in one of our previous works, 
we have shown that, even for these 
views of trustworthy ML, 
there is a higher-layer converged understanding 
of trustworthy machine learning from the 
data perspective \citep{wang2022toward2}. 
In our formulation, 
we showed a unified generalization error bound 
that can lead to the above two formulations of methods. 
Our unified generalization error bound 
essentially suggests that 
a path to developing such methods 
is the identificaiton 
of those features that are statistically correlated in a dataset, 
but spurious in practical settings, 
and informing the models about these features
with either regularizations or augmentations.

\section{Trustworthy Machine Learning in Causality Perspectives}
\label{sec:review:main}
The previous section provided an overview of trustworthy machine learning across multiple topics, revealing a common theme of data-centric machine learning techniques that involve discarding or perturbing certain features to achieve trustworthiness as defined by human experts. 
This core technique has a conceptual connection to the topic of causality in Pearl's language, particularly in the connections between feature perturbation and the intervention and counterfactual concepts in Pearl's causal hierarchy. In the second half of this survey, we will offer another overview of recent trustworthy machine learning papers, but this time from the perspective of Pearl's causal hierarchy. We will organize the papers that explicitly mention the causality terms and associate them with the levels of Pearl's causal hierarchy. 

Before delving deeper into the core techniques of intervention and counterfactual causation, we will first provide a brief overview of the concepts and terminologies of causality.

\subsection{Background in Causality} \label{sec:background}

Causality is a fundamental concept in many fields.
In its most basic form, causality refers to the relationship between cause and effect, where a cause is an event or condition that produces an effect. However, establishing causal relationships is often challenging, as many factors can influence an outcome, and it can be difficult to distinguish between causal and non-causal relationships.

\subsubsection{Background: Confounding Variable and SCM}

In causal inference, we are often interested in finding the causal effect of a variable $X$ (``treatment variable'') to another variable $Y$ (``outcome variable''). Such causal effect cannot be estimated in general from the statistical association between $X$ and $Y$ in observational data, due to the spurious correlation brought by one or more variables. More formally, according to the associational criterion in~\citep{pearl09causality}, we say that $X$ and $Y$ are confounded if there exists a variable $Z$ which is not affected by $X$ but is associated with both $X$, and $Y$ conditional on $X$. 
We refer to $Z$ as a confounder, or confounding variable, for the relationship between $X$ and $Y$. Although in the context of causal diagrams or Bayesian networks, ``confounder'' is often used to refer only to variables that causally influence both $X$ and $Y$~\cite{pearl09causality}, we follow the terminology in more general discussions~\cite{pourhoseingholi2012control, zhao2020training} and use the word ``confounder'' interchangeably with ``confounding variable'' to incorporate a broader range of variables that create spurious correlations between $X$ and $Y$. For example, following the story in Example~\ref{exp1}, 
we can consider the sea environment as the confounder.

\begin{figure}[h]
    \centering
    \subfloat[\centering ]{{\includegraphics[width = 0.3\textwidth]{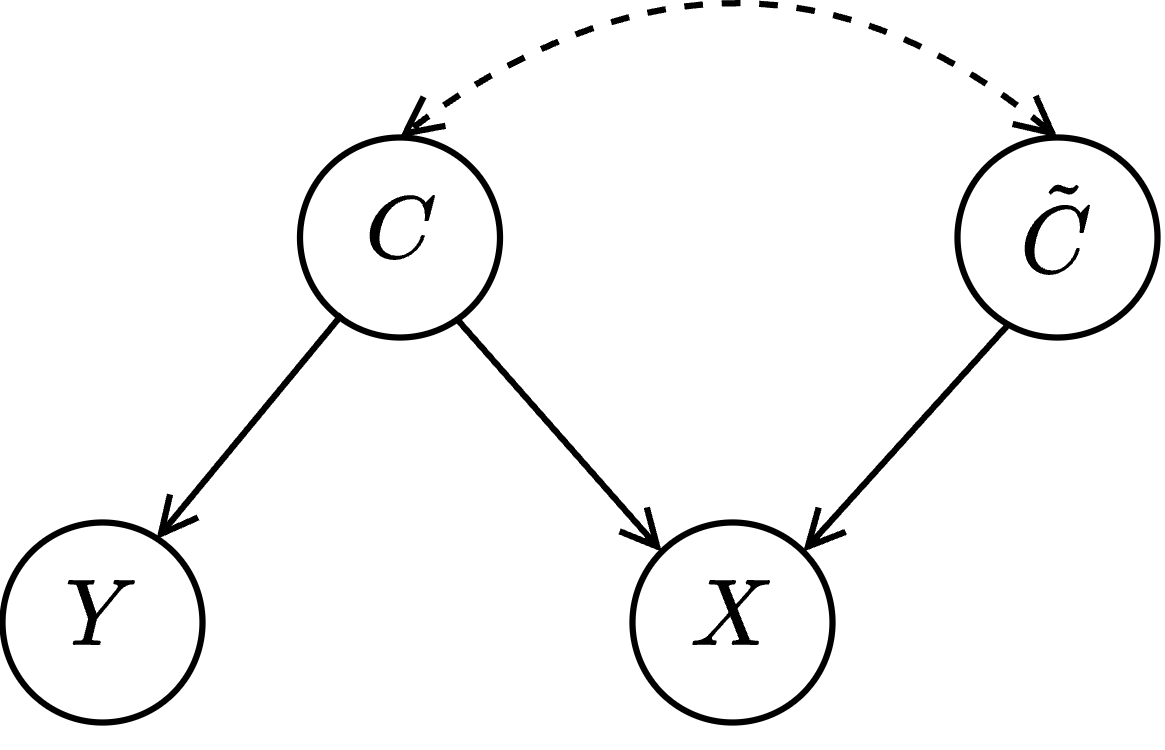} }}
    \hspace{0.1\textwidth}
    \subfloat[\centering \textcolor{gray}{} ]{{\includegraphics[width=0.3\textwidth]{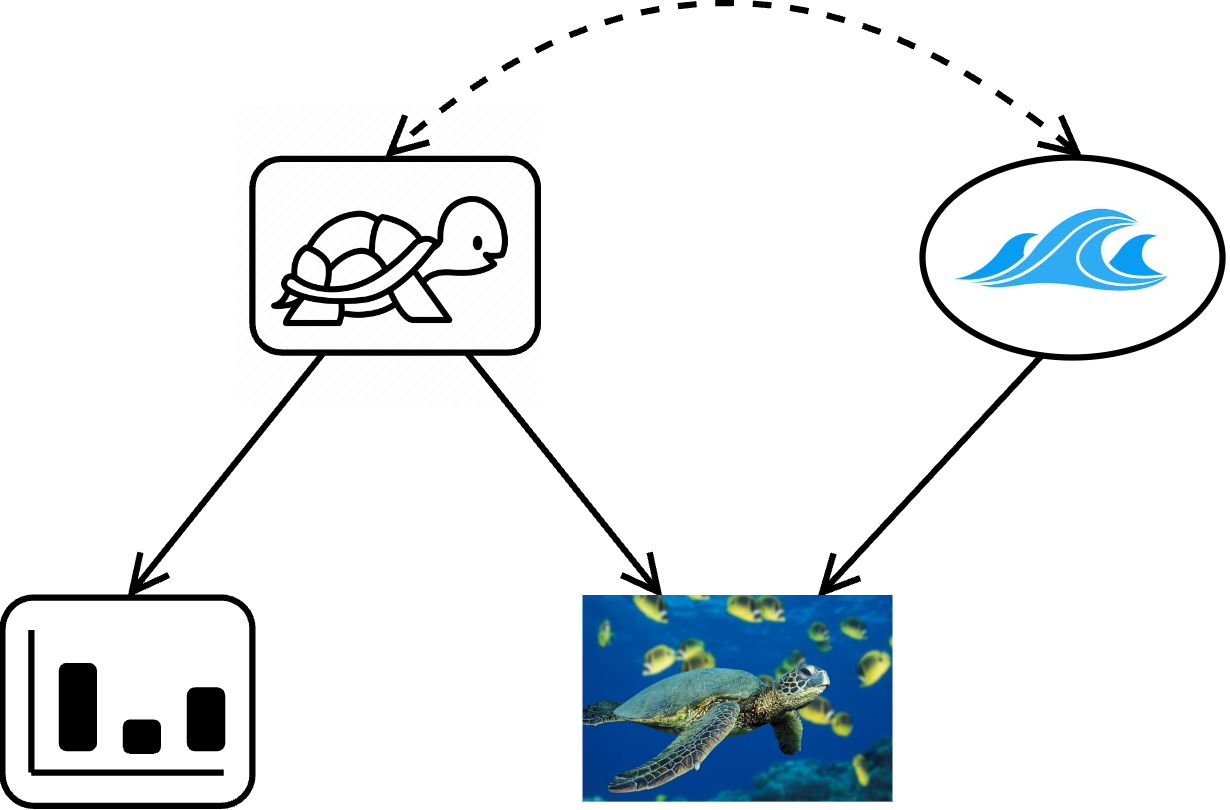} }}%
    \caption{(a): The graphical model representing the data generation process for the sea turtle vs. tortoise classification. An arrow in $C \rightarrow X$ means $C$ is a direct cause of $X$. A dashed double-arrow arc means there is unobserved confounder between $C$ and $\tilde{C}$. (b): An informal, illustrated version of (a)}
    \label{fig:confounder_main}%
\end{figure}

Intuitive as it might seem, 
it is not easy to formalize the notion of ``cause" and ``effect", or tell which correlations are ``spurious'',  
using the standard language of probability theory. 
Even with a fully specified population density function, 
we are still unable to make predictions about a hypothetical distribution where a new treatment is imposed 
unless we make some assumptions about the generating mechanisms of the variables~\citep{pearl09causality}.
To better study the causal relationship between variables,
the community has introduced a formal language 
called Structural Causal Model (SCM). 

\begin{definition}{\footnotesize Structural Causal Model}{scm} \label{box:scm}
\footnotesize
A structural causal model is a 4-tuple 
$(U, V, \mathcal{F}, P(U)$, where
$U = \{U_1, U_2, ..., U_n\}$ is a set of exogenous variables,
which account for factors or influence from outside the model, 
and are not caused by any other variables in the model.
$V = \{V_1, V_2, ..., V_n\}$ 
is a set of endogenous variables, whose values are determined by other variables in the model.
$\mathcal{F} = \{\mathcal{F}_1, \mathcal{F}_2, ..., \mathcal{F}_n\}$ 
represents a set of functions
such that 
$V_i = \mathcal{F}_i(\textrm{PA}_i,U_i)$ for $i=1,\ldots,n$,  
where $\textrm{PA}_i \subseteq V$
represents the parents of the variable $V_i$.
$P(U)$ is a probability distribution over the exogenous variables $U$. 
\end{definition}

Every SCM is associated with a graphical representation
that illustrates the causal relationship among variables in the SCM, 
referred to as the Graphical Causal Model, or ``graphical model''.
We assume that exogenous variables are mutually independent, 
and omit them from the graphical model for simplicity.
Otherwise, if $U_1$ and $U_2$ are dependent, 
we add a dashed double-arrow curve between the endogenous variables $V_1$ and $V_2$
in the graphical model,
meaning they are confounded.

Let Figure~\ref{fig:confounder_main}(a) be the graphical model of the training data for the sea turtle vs. tortoise image classification in Example~\ref{exp1}. Among the endogenous variables, $C$ represents the biological feature of the animal, such as the shape, color and texture of its shell and feet; $\tilde{C}$ represents the background of the image; $X$ and $Y$ are the image and label respectively. Exogenous variables represent external factors, e.g., $U_C$ may represent the individual characteristics of the animal, and $U_X$ may be the photographic conditions. The causal diagram encodes our assumptions about the data generation process, that $C$ and $\tilde{C}$ cause $X$, that $Y$ is only caused by $C$ except the error term $U_Y$, etc. Although the biological feature $C$ and the background $\tilde{C}$ do not cause each other, there is spurious correlation between them due to the data collection process. For example, the data may only include photos of animals in their natural habitats at certain places. We consider the correlation spurious because it may change in another set of photos collected from different places (e.g. a thermal transportation box), or in images of cartoon and art paintings where animals can appear in any background. A model that constantly performs well should be one that makes the prediction based only on the biological features.

Given a causal diagram, one can infer the conditional independence and dependence relations between variables from graphic patterns such as chains, forks, colliders and the d-separation criteria. For a detailed introduction of them, please refer to~\cite{glymour2016causal}. Suppose we are interested in the relation between $C$ and $X$, we can infer from Figure~\ref{fig:confounder_main} that $\tilde{C}$ is a confounder between them through the path $(C,\tilde{C},X)$.

In Section~\ref{sec:data_to_prediction}, we will go beyond this single example and discuss the connections between machine learning and causal inference in a broader setting. 

\subsubsection{Connections to Machine Learning Development} \label{sec:data_to_prediction}
We now shift back to the discussions of the machine learning topics. 
In Section~\ref{sec:robustness}, we have discussed a summary of trustworthy machine learning covering multiple different topics 
and converged the topic to a shared data-centric theme. 
Here, we continue to study this theme. 
We believe the challenge is mainly caused by a major non-robust assumption taken for many machine learning models:
$(x_1, y_1),...,(x_n, y_n)$ are realizations of random variables that
are i.i.d. (independent and identically distributed) with joint distribution $P(X,Y)$ \cite{peters2017elements}. In other words, previous machine learning models are usually evaluated based on the same dataset distribution used for the training, 
which often does not reflect the true testing scenario in practice.
Therefore, the community has investigated a long line of research focusing 
on topics that the testing scenario is different from the training scenario, 
as discussed in the ``robustness'' in Section~\ref{sec:robustness}. 

This disparity between training and testing has not been emphasized by the machine learning models for a long time, 
probably because the concentration on improving accuracy over i.i.d data
is one of the fastest ways to facilitate method development. 
As a result, machine learning development does not often recognize the underlying causal model of the data-generating process. 
Therefore, the models' so rich ability helps them capture all kinds of patterns in the data correlated with the output (termed as ``curse of universal approximation'' in certain prior work \cite{wang2022toward2}), including both causal features $C$ and non-causal features $\tilde{C}$. There is a pressing need to eliminate confounders' impact on the result (more on this in ``\ref{sec:intervention}'' and ``\ref{sec:counterfactuals}'')

\begin{figure}[h]
    \centering
    \subfloat[\centering Assumed in this survey]{{\includegraphics[width = 0.28\textwidth]{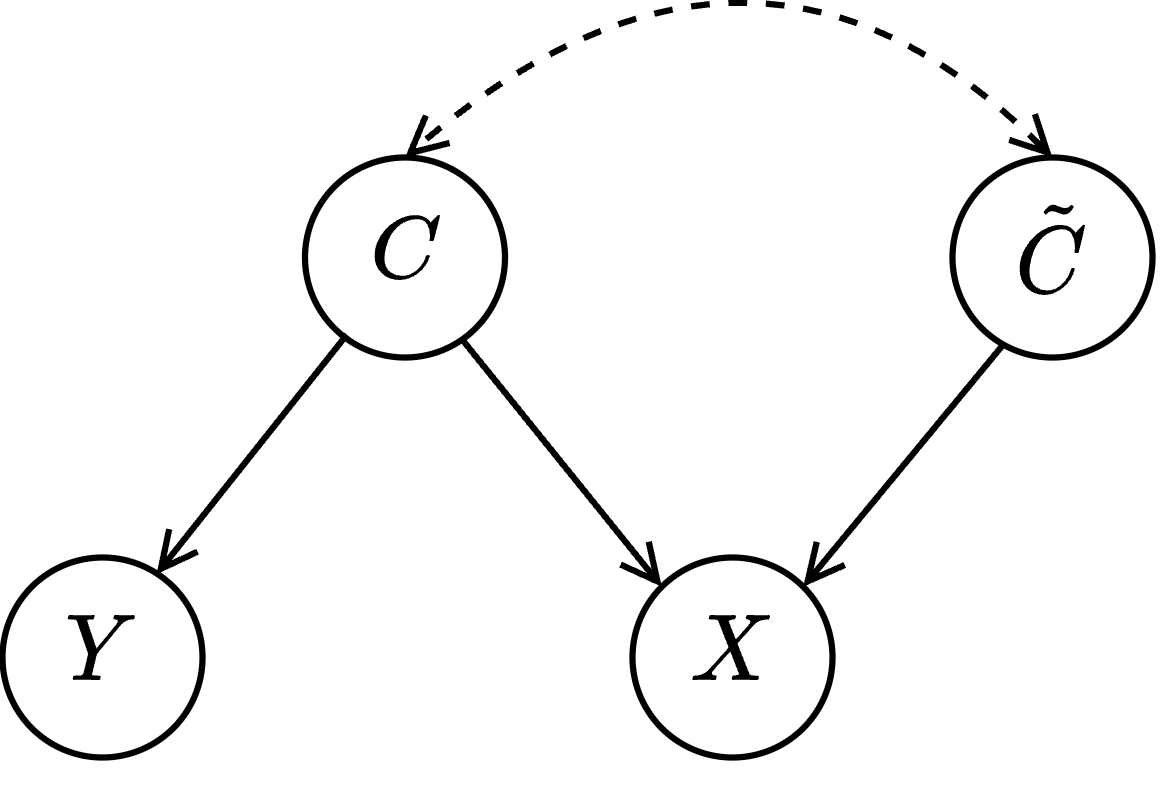} }}
    \hspace{0.05\textwidth}
    \subfloat[\centering Unbiased]{{\includegraphics[width=0.22\textwidth]{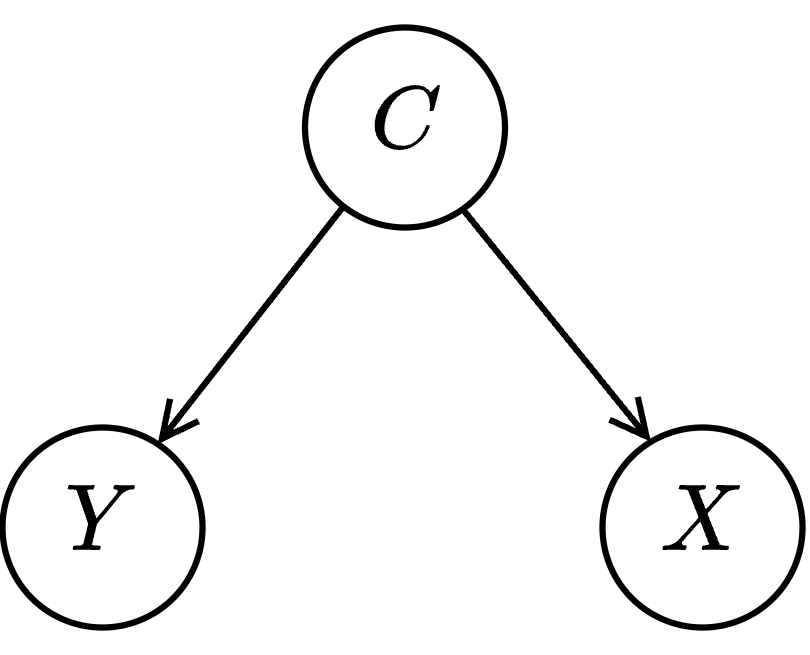}}}%
    \hspace{0.07\textwidth}
    \subfloat[\centering Alternative]{{\includegraphics[width=0.22\textwidth]{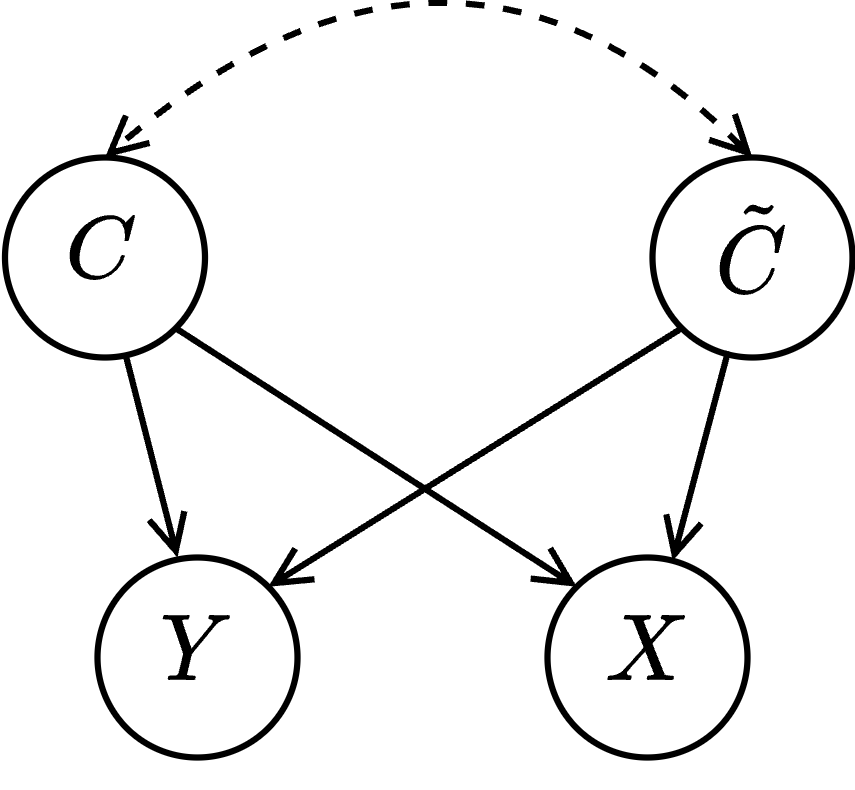}}}%
    \caption{Different graphical models for the data generation process. (a) is the main graphical model in our survey representative of many different settings; (b) is the graphical model of an ideally unbiased dataset that doesn't contain $\tilde{C}$; (c) is an alternative graphical model which adds a causal link $\tilde{C} \rightarrow Y$ to (a), but cannot be distinguished from (a) by the machine learning model.}
    \label{fig:comparison}%
\end{figure}

Recent works at the intersection of causal inference and machine learning \cite{Zhang2020CausalIF, Wang2021CausalAF, yue2020interventional, wang2020visual, kocaoglu2018causalgan, feder2022causal, feder-etal-2021-causalm} have introduced different graphical models encoding various assumptions about the data generation process. After extensive investigation, we use Figure~\ref{fig:comparison}(a) as the main graphical model for our survey because it is most representative of a wide range of settings. It is the same as Figure~\ref{fig:confounder_main}(a) except that now we assign much broader meanings to the variables. $C$ represents the causal features often related to the aim of the task, such as object appearance and location in an object detection task, or reviewers' attitude towards an item in a sentiment classification task, etc. 
$\tilde{C}$ represents the non-causal features that should not be leveraged by the model for predictions. In fairness considerations, $\tilde{C}$ often denotes demographic information, while in domain generalization and adaptation, it typically reflects domain-specific biases and is usually categorical or can be approximated as such. $\tilde{C}$ can also be challenging to model, when it is multi-dimensional and continuous, as in adversarial attacks, or involves high-level concepts difficult to separate from causal variables $C$ \cite{feder-etal-2021-causalm}.

As shown in Figure~\ref{fig:comparison}(a), we assume that a datapoint $X$ is generated by causal features $C$ and non-causal features $\tilde{C}$. Some unmeasured variables produce non-causal correlation between $C$ and $\tilde{C}$. While some works consider annotation artifacts~\citep{malaviya-etal-2022-cascading} or incomplete information in the causal features about the label~\cite{Wang2021CausalAF}, which implies a causal link from $\tilde{C}$ to $Y$, we assume in our survey that the dataset is carefully prepared such that $Y$ is the ground truth label that only depends on $C$.

Suppose there is an ideally unbiased dataset, whose graphical model is given in Figure~\ref{fig:comparison}(b). 
Based on the equation $P(y|x) = \sum_c P(c|x)P(y|c)$, and assuming a fixed $P(y|c)$, a model trained to estimate $P(y|x)$ on the data distribution would learn a good estimator of the causal feature $\hat{P}(c|x)$. 
In Figure~\ref{fig:3ladder_scm}(a), however, $\tilde{C}$ confounds the relationship between $C$ and $X$ through the path $({C},\tilde{C},X)$. 
More specifically, the path produces a non-causal association between $C$ and $X$, where the association between $C$ and $\tilde{C}$ is often a bias in the dataset. 

Consider the alternative graphical model of the data generation process in Figure~\ref{fig:comparison}(c), where $\tilde{C}$ is a direct cause of $Y$. While one may notice that it is not observationally equivalent to the graphical model in Figure~\ref{fig:comparison}(a), it is impossible for a statistical model observing only $X$ and $Y$ to differentiate between those two graphical models. 
Instead, in Figure~\ref{fig:comparison}(a) the path $(\tilde{C},C,Y)$ produces a non-causal association between $\tilde{C}$ and $Y$, which the model may capture regardless of the causality. In practice, $\tilde{C}$ is often shallow features that can be learned in the first few layers of a neural network \cite{wang2019learning}, or at the early stage of training \cite{nam2020learning}, which may exacerbate the model's tendency to use $\tilde{C}$ for prediction. 

SCM provides us with a convenient tool to identify the confounders and qualitatively analyze the undesired behaviors of machine learning models when deployed in non-IID settings. In the remainder of this section, we will see that causal inference provides us with a lot of powerful tools to more quantitatively estimate causal effect and remove the influence of confounders, which has been increasingly used in recent works across different topics of trustworthy machine learning. In fact, many other works discussed in Section~\ref{sec:robustness} which did not explicitly use causal tools can also be revisited and understood from a causal perspective. 

\subsubsection{Background: Levels in Pearl's causal hierarchy}

We now continue to offer the background in causality literature. 
These discussions might be perceived as overly detailed for some readers with a working knowledge of causality. However, we believe these discussions are essential as we later will map this causal hierarchy to current machine learning methods. 
We hope such mapping will immediately help set the expectations of what current trustworthy methods can achieve. 

Pearl's causal hierarchy (PCH) \cite{pearl1995causal, pearl2000models, pearl2018book} provides a unified framework for discussing different aspects of causality. 
The hierarchy consists of three levels of causation ($\mathcal{L}_1$, $\mathcal{L}_2$, and $\mathcal{L}_3$). The first level is associational causation, which works on conventional statistics and does not incorporate any causal techniques to identify the causal relationships between the variables. In other words, $\mathcal{L}_1$ does not distinguish between ``correlation'' and ``causation'', and seeks to identify correlations in the observed data. 

Unlike $\mathcal{L}_1$, the remaining two levels of causation ($\mathcal{L}_2$, and $\mathcal{L}_3$) adhere to the principle that ``correlation is not causation''. Although the causal mechanisms behind a system are often unobservable, they leave observable traces in the form of data that can be analyzed. The second level of causation is typically referred to as intervention, while the third level is known as counterfactuals. The primary difference between intervention and counterfactuals lies in their ability to consider scenarios that contradict the observed data. Intervention involves asking and answering questions about the effect of an action on the resulting distribution of the overall observed data. 
In other words, intervention often operates on a population level to estimate the effect of the intervention. 
In contrast, counterfactual reasoning involves considering hypothetical scenarios at an individual level, including those that did not actually occur, to quantify the effect of an intervention.

\paragraph{First Level ($\mathcal{L}_1$)}
The first level of causal hierarchy deals with the question: ``How likely is $Y$ given that one observes $X$?''. This level focuses on measuring statistical associations between variables, represented by the conditional probability $P(Y|X)$. However, such associations alone cannot establish causal relationships, as they can be influenced by confounding variables not accounted for in the analysis. For example, data might reveal that ice-cream and sunglasses sales are highly correlated in a certain region, but this might simply reflect the influence of a confounding variable - hot weather, which boosts both ice-cream and sunglasses sales \cite{huff2023lie}. Most of the conventional statistical and machine learning methods primarily seek to find correlational patterns in data. Although this might be useful under the i.i.d. assumptions, these patterns often fail to generalize to new scenarios. This is because they do not provide insight into the underlying causal mechanisms generating the data.

\begin{figure}[h]
    \centering
    \subfloat[\centering Before intervention]{{\includegraphics[width = 0.3\textwidth]{figs/new/causal_graph.png} }}
    \hspace{0.1\textwidth}
    \subfloat[\centering After intervention ]{{\includegraphics[width=0.3\textwidth]{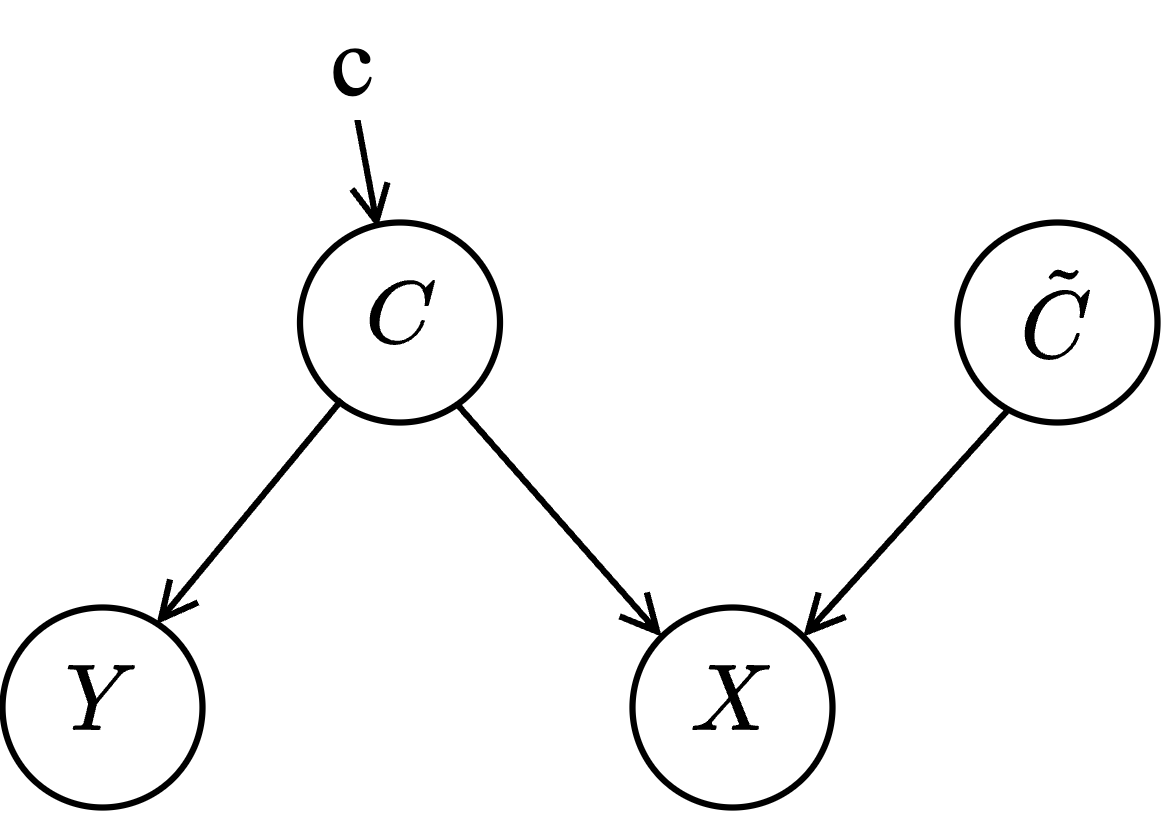}}}%
    \caption{The effect of intervention on the graphical models of the data generation process.}
    \label{fig:3ladder_scm}%
\end{figure}

\begin{definition}{\footnotesize do-Calculus}{do_calculus}
\footnotesize 
    The $do-$calculus is a system that replaces the conditional distribution with an intervened distribution forcing the value of a variable, such that it is randomly assigned without any influence of its parents. It consists of three schemes that provide graphical $\mathcal{G}$ criteria for when certain substitutions may be made.
    
    \begin{description}
        \item[\textbf{Rule 1.}] $P(Y|do(X), Z, W) = P(Y|do(X), W)$, if $(Y \indep{}.. Z | W, X)_{\mathcal{G}_{\bar{X}}}$ 
        \item[\textbf{Rule 2.}] $P(Y|do(X), do(Z), W) = P(Y|do(X), Z, W)$, if $(Y \indep{}.. Z | W, X)_{\mathcal{G}_{\bar{X}, \underaccent{\bar}{Z}}}$
        \item[\textbf{Rule 3.}] $P(Y|do(X), do(Z), W) = P(Y|do(X), W)$, if $(Y \indep{}.. Z | W, X)_{\mathcal{G}_{\bar{X}, \overline{Z(W)}}}$
    \end{description}
\label{box:do_calculus}
\end{definition} 

\paragraph{Second Level ($\mathcal{L}_2$)}
The second level of Pearl's hierarchy, known as interventionist causation, involves hypothetical or ``conditional'' questions such as ``How likely would Y be if one were to make X happen?''.
This level aims to understand the implications of intervening in a system.

In Pearl's framework, an intervention is an action that changes the value of a variable by some external mechanism. For example, in Figure~\ref{fig:3ladder_scm}, if we intervene on variable $C$, we enforce a value to it regardless of the original mechanism that generates $C$. This intervention effectively removes all incoming edges to this node. The causal effect of this intervention is then gauged by the consequent change in the outcome variable.
For example, to see whether there is a causal effect of local ice-cream sales on sunglasses sales, we may implement a policy to close all the ice-cream shops and see the change \cite{glymour2016causal}. In Example~\ref{exp1}, if we want to know whether a machine learning model erroneously uses color information as a heuristic (such as the blue color in an ocean background for sea turtles), we may convert all images to grayscale to see how the distribution of predictions change.

For interventional experiments, a gold standard is the randomized controlled trials. By randomizing the assignment of treatment, we make sure that any change in the outcome is only a result of the treatment, which helps us make sound scientific conclusions and informed decisions. However, this method might not always be practical, ethical, or economically feasible. In causal inference, intervention is formalized by the $do$-operation, and the $do$-calculus provides a set of rules for manipulating expressions involving $do$-operations. There exist various techniques to estimate the effect of interventions from observational data, such as adjustment methods, Inverse Probability Weighting, and Instrument Variables, to name a few. In recent years, there is a growing interest in the machine learning community to apply these techniques to find robust patterns in the data that capture the causal relationship between variables. This is fueled by the potential these techniques have in enhancing the trustworthy properties of machine learning models.

\paragraph{Third Level ($\mathcal{L}_3$)}    

The third level of causation, known as counterfactuals, also deals with hypothetical scenarios. 
It allows questions like ``Given that one observed $X$ and $Y$, how likely would $Y$ have been if $X'$ had been true?'', where $X'$ may contradict the observed event.
In practice, it is difficult to directly observe counterfactual outcomes, and so counterfactual causation relies on statistical methods to estimate these outcomes.
Counterfactual and intervention might appear similar. 
However, 
in interventions, we focus on what will happen on average if we perform an action on overall observed samples, whereas in counterfactuals we focus on what would have happened if we had taken a different course of action in a specific situation, given that we have information about what actually happened.

For example, imagine we have a machine learning model that decides whether to grant loans to someone based on income, employment status, credit score, and age. Suppose the model rejected the loan application from an individual with medium income, part-time job, good credit score, and older age. The person would be interested in finding a counterfactual explanation such as ``What if I had a full-time job instead of a part-time job?'' If the model would have approved this application in this hypothetical setting, it could motivate the applicant to find a full-time job. From a fairness perspective, if the model's prediction would have flipped by merely changing the demographic group of the applicant, we know that the model may contain social bias that needs to be addressed. As the top level of causality, counterfactual analysis enables us to answer causal questions that span across various hypothetical scenarios, and isolate the treatment effect by different mechanisms. Compared to intervention, it often requires stronger assumptions and more accurate specification of the causal model, because it often involves extrapolation outside the support of the observed data. 

From the above text, we can see the differences in information-richness among the three levels of causation:
higher layers $\mathcal{L}_i$ encode more information than the lower layers, forming a hierarchy $\mathcal{L}_3 > \mathcal{L}_2 > \mathcal{L}_1$ \cite{pearl2018book}.
Therefore, to answer questions related to Layer $i$, knowledge of Layer $i$ or higher is necessary. 
Further, 
it is highly unlikely for the layers of PCH to collapse, meaning that $\mathcal{L}_1$ contains answers to $\mathcal{L}_2$ and $\mathcal{L}_2$ contains answers to $\mathcal{L}_3$, as it requires capturing the exact representation of the population in the samples, which is difficult to achieve \cite{pearl2018book}. This forms the basis of the development of PCH and is stated in the Causal Hierarchy Theorem \cite{bareinboim2020pearl}.

Overall, Pearl's causal hierarchy provides a comprehensive framework for reasoning about causality in a variety of contexts. By understanding the different levels of the hierarchy, it is possible to develop more reliable and trustworthy machine-learning techniques that take into account the complexities of causal relationships, 
and it will also help us set up the expectations 
for 
what degree of trustworthiness 
a method can eventually achieve, 
despite it might perform well on certain benchmarks.
In the following sections, we will discuss how Pearl's hierarchy has been used in recent research on trustworthy machine learning.
Following the seminal books on causal inference \cite{pearl09causality, glymour2016causal}, we assume that all variables are discrete in the math derivations in this section, to ensure consistency in notation. These derivations can be easily extended to the continuous case through integration on probability density functions.

\subsection{Intervention: the second level} \label{sec:intervention}
Observational data often provides limited insight into the structural causal model that generated it, partly due to the difficulty of discerning whether observed associations between variables reflect causal relationships. 
This challenge arises because many variables' roles - causal or merely correlational - remain unclear. 
By external control and manipulation of these variables, we can investigate their potential causal influences more effectively. This active manipulation and observation of effects is a key component of the second level of causation ($\mathcal{L}_2$) and allows us to construct a more accurate representation of the true data-generating SCM.

In causal inference, quantitative measurement of causal effect is facilitated by the $do$-operator, which forces a variable $X$ to take the value $x$, denoted as $do(X=x)$ or $do(x)$. Formally, given a structural causal model $\mathcal{M}$, the intervention $do(x)$ is defined as the substitution of structural equation $X = \mathcal{F}_X(\textrm{PA}_X,U_X)$ with $X = x$. 

Intervening on a variable is different from conditioning on it, which can be explained via the example in Figure~\ref{fig:3ladder_scm}. In Figure~\ref{fig:3ladder_scm}(a), the path $(C,\tilde{C},X)$ produces spurious correlation between $C$ and $X$. By conditioning on $C$, we narrow our focus to part of the sample space where $C=c$ in the distribution. If we change the value of $C$ to condition on, $\tilde{C}$ is also likely to change due to the statistical association between them. In contrast, by intervening on $C$, we change the distribution by removing all edges pointing to $C$ and assigning a value $c$ to it. If we change the value of $c$ for intervention, the change will not be transmitted to $\tilde{C}$. 

Note that Figure~\ref{fig:3ladder_scm}(a) is the main causal diagram in our survey, and our above example has implications in the context of machine learning. If we conduct the stochastic intervention~\cite{pearl09causality} on the data distribution $P$ by assigning a distribution $P(C)$ to the causal feature $C$, we can generate a new distribution $P_m$ where the marginal distribution of $C$ remains the same but $\tilde{C}$ no longer confounds the association between $C$ and $X$. Further, we will show that such intervention will remove the confounding effect of $\tilde{C}$ on the association between $X$ and $Y$, and hence de-confound the prediction of a machine learning model. Consider the probability of $Y$ conditional on $X$ for distributions $P$ and $P_m$,

\begin{align} 
P(y|x) &= \frac{P(x, y)}{P(x)} 
= \frac{\sum_c P(x, y, c)}{{P(x)}} 
= \frac{\sum_c P(c)P(x|c)P(y|x, c)}{P(x)} \notag\\
&= \frac{\sum_c P(c)P(x|c)P(y|c)}{\sum_c P(c)P(x|c)} \label{eq:xyc:0}\\
P_m(y|x) &= \frac{\sum_c P_m(c)P_m(x|c)P_m(y|c)}{\sum_c P_m(c)P_m(x|c)}
= \frac{\sum_c P(c)P_m(x|c)P(y|c)}{\sum_c P(c)P_m(x|c)} \notag\\
&= \frac{\sum_c P(c)P(x|do(c))P(y|c)}{\sum_c P(c)P(x|do(c))} \label{eq:xyc:1}
\end{align}
\begin{wrapfigure}{r}{0.4\textwidth}
\footnotesize
\begin{tcolorbox}[colback=white!90!gray, colframe=white!70!black,left=1pt, right=1pt, top=0.5pt, bottom=0.5pt]
\textbf{Intuition:} Equations~\ref{eq:xyc:0}, \ref{eq:xyc:1} can be understood as following: Suppose the presence of objects in images (e.g., sea turtle features) is the causal feature we care about. If we intervene on the objects in the data generation process (e.g., the photographing process), and then train a model on the intervened data distribution, the model will learn to pick up the object information without being distracted by other factors (e.g., the background).
\end{tcolorbox}
\end{wrapfigure}
where we use the condition that the generating mechanism for $Y$ is not changed, i.e. $P_m(y|c) = P(y|c)$. Comparing Equations~\ref{eq:xyc:0}, \ref{eq:xyc:1}, it can be seen that the relationship between $X$ and $Y$ is confounded by $\tilde{C}$ only through the factor $P(x|c)$. Because $P(x|do(c))$ removes the confounding, a model trained to fit the statistical association between $X$ and $Y$ on the interventional distribution $P_m$ will not be affected by the confounder $\tilde{C}$. 

While the above derivation gives a conceptual direction, directly intervening on $C$ is often impractical. $C$ is the underlying causal feature of the data examples which we usually don't have access to and is hard to model. Luckily, a series of methods in causal inference and statistics related to the notion of intervention has enabled us to overcome the technical difficulties. In the remainder of this subsection, we will introduce these methods and review recent works that apply them to topics in trustworthy machine learning. Among them, recent works using adjustment methods are often based on a set of causal assumptions different from that in our main graphical model, which will be detailed in Section \ref{sec:backdoor_adjustment}, \ref{sec:frontdoor_adjustment}. 
We will also revisit recent works discussed in Section~\ref{sec:robustness} to understand them from a causal perspective.

\paragraph{Randomized Controlled Trial} \label{sec:rct}
Randomized Controlled Trial (RCT) is a scientific methodology that operates on the principle of random assignment or collection of samples in different classes, ensuring that any observed differences in outcomes are due to the intervention rather than confounding variables. 
For example, in Example~\ref{exp1}, rather than collecting images where sea turtles are mostly beside the sea, we carefully collect a balanced set of images where sea turtles occur in all kinds of backgrounds according to the marginal distribution of background, and similarly for the images of tortoises. However, it is often impossible to collect such dataset due to the prohibitive cost and sometimes unattainable conditions (e.g. a sea tortoise near a crater).

In the machine learning literature, data augmentation is often used to get an enlarged dataset where the confounding effect of $\tilde{C}$ is removed. This can be seen as a RCT method from a causal perspective. During this process, we first identify the confounder $\tilde{C}$ and its distribution. For each training datapoint $x$, we generate a minimally perturbed version of $x$, denoted as $x_{\tilde{c}}$, by setting the value of its confounder $\tilde{C}$ to a value $\tilde{c}$ while keeping everything else the same in the process where $x$ was generated. We repeat this process by sampling different values $\tilde{c}$ from the marginal distribution $P(\tilde{C})$ independently of the causal feature $C$ to get different $x_{\tilde{c}}$'s, resulting in a randomized dataset. Standard training on this dataset gives the loss function as in Equation~\ref{eq:cda:rct}.  
\begin{align}
    \argmin_{\theta} 
    \dfrac{1}{n}\sum_{(x,y)\in (\X,\Y)}\mathop{\mathbb{E}}_{\tilde{c} \sim D_{\tilde{C}}}l(f(x_{\tilde{c}};\theta), y),
    \label{eq:cda:rct}
\end{align}
Data augmentation that randomizes confounders is equivalent to intervention in terms of effects. To show this, we can have a more formal look at this process. Let $P$ and $P'$ denote the probability distributions of the original data and the randomized data respectively. Assume that the marginal distribution of $C$ and $C'$ remains the same but they become mutally independent after randomization, i.e. $P'(c) = P(c), P'(\tilde{c}) = P(\tilde{c}), P'(c, \tilde{c}) = P'(c)P'(\tilde{c})$. The structural equations for $X$ and $Y$ should remain the same, i.e., $P'(y|c)=P(y|c), P'(x|c, \tilde{c})=P(x|c, \tilde{c})$, 

\begin{align}
\begin{split}
P'(x|c) &= \frac{P'(x,c)}{P'(c)} = \frac{\sum_{\tilde{c}}P'(x,c,\tilde{c})}{P'(c)} = \sum_{\tilde{c}}P'(x,\tilde{c}|c) \\
&= \sum_{\tilde{c}}P'(\tilde{c}|c)P'(x|c,\tilde{c}) = \sum_{\tilde{c}}P'(\tilde{c})P'(x|c,\tilde{c}) \\
&= \sum_{\tilde{c}}P(\tilde{c})P(x|c,\tilde{c}) = P(x|do(c))
\end{split}
\label{eq:rct:explain}
\end{align}
\begin{wrapfigure}{r}{0.4\textwidth}
\footnotesize
\begin{tcolorbox}[colback=white!90!gray, colframe=white!70!black,left=1pt, right=1pt, top=0.5pt, bottom=0.5pt]
\textbf{Intuition}: Equation~\ref{eq:rct:explain} can be understood with the follows. If we randomize the existence of objects in images (e.g., mismatching the animals with backgrounds randomly before taking photographs), we break the natural tendency of certain objects to occur in certain backgrounds. Then model trained with such data will be able to pick up the object information for prediction, without being influenced by the background.
\end{tcolorbox}
\end{wrapfigure}
where the last equation is the backdoor adjustment formula, which will be introduced in Section~\ref{sec:backdoor_adjustment}. This shows that the statistical association between $X$ and $C$ in the randomized distribution $P'$ captures the causal effect between them in the original, biased distribution $P$, as if we had conducted the intervention $do(c)$ on the original dataset. Then, we can analogously derive $P'(y|x)$ following Equation~\ref{eq:xyc:1} and conclude that a model trained on $P'$ will not be affected by the confounder $\tilde{C}$.

A large body of work using data augmentation for trustworthy properties (Section~\ref{sec:robustness}) can be understood from the perspective of Randomized Controlled Trials, despite that the concrete design of the augmentation method varies with the problem settings. In domain generalization and domain adaptation, $\tilde{C}$ is often the domain category associated with domain-specific bias (e.g. style or texture features). $\tilde{C}$ is often implicitly assumed to conform to a uniform distribution, because different domains are considered equally important. Recent work has conducted data augmentation on source (training) domain images by matching the style of the target domain for domain adaptation \citep{hoffman2018cycada, mutze2022semi, bousmalis2017unsupervised, kim2017learning}, or the styles of other training domains for domain generalization \citep{shankar2018generalizing,yue2019domain,gong2019dlow,zhou2020deep,huang2021fsdr}. In the fairness literature, $\tilde{C}$ may be demographic categories, and recent works have used data augmentation to alleviate bias and discrimination in machine learning models \cite{sharma2020data, pastaltzidis2022data, zhao-etal-2018-gender, zmigrod-etal-2019-counterfactual}.

When $\tilde{C}$ is multi-dimensional or continuous, estimation of the inner expectation in Equation~\ref{eq:cda:rct} is expensive. An alternative formulation has been proposed, which finds the worst-case perturbation on the confounder $\tilde{C}$ and then minimize the loss.
\begin{align}
    \argmin_{\theta} 
    \dfrac{1}{n}\sum_{(x,y)\in (\X,\Y)}\mathop{\max}_{\tilde{c}}l(f(x_{\tilde{c}};\theta), y),
    \label{eq:cda:rct:2}
\end{align}

Because $\max_{\tilde{c}}l(f(x_{\tilde{c}};\theta), y)$ is an upper bound on $\mathbb{E}_{\tilde{c} \sim D_{\tilde{C}}}l(f(x_{\tilde{c}};\theta), y)$, solving Equation~\ref{eq:cda:rct:2} has similar effect to Equation~\ref{eq:cda:rct} on the model.
It emphasizes the worst-case scenario where the confounder $\tilde{C}$ breaks the prediction of a model when it is expected to be invariant to $\tilde{C}$. It has been extensively used in adversarial robustness studies. From a causal perspective, $\tilde{C}$ is a attack-related feature representing whether the data is attacked, the attack algorithm and configurations, the initialization of the noise, etc. The counterfactual $x_{\tilde{c}}$ is often realized as a perturbed version of $x$, where an $\ell_p$ norm constraint is placed on the perturbation to ensure that it does not change the causal feature of the image perceived by humans. Then Equation~\ref{eq:cda:rct:2} converges to Equation~\ref{eq:master:2} in our discussion of the common theme of trustworthy machine learning in Subsection~\ref{subsec:master_eq}. Unlike Equation~\ref{eq:cda:rct}, this formulation does not require prior knowledge on the distribution of $D_{\tilde{C}}$, which is difficult to get in this scenario and some others. This worst-case formulation goes beyond adversarial robustness studies and has been used in fairness \cite{wang-etal-2021-dynamically, wang2022fairness} and domain generalization \cite{shankar2018generalizing, long2022domain} research as well.

Apart from RCT, several other options exist for de-biasing a model, such as Instrument Variable, Backdoor adjustment, and Front-door adjustment method.
We will continue our discussion with Instrument Variable (IV) \cite{wright1928tariff, baiocchi2014instrumental}. 

\begin{figure}[h]
    \centering
    \subfloat[\centering Prior Instrument Variable(IV) Learning]{{\includegraphics[width=0.27\textwidth]{figs/new/causal_graph.png} }}
    \hspace{0.1\textwidth}
    \subfloat[\centering Post Instrument Variable(IV) Learning]{{\includegraphics[width=0.33\textwidth]{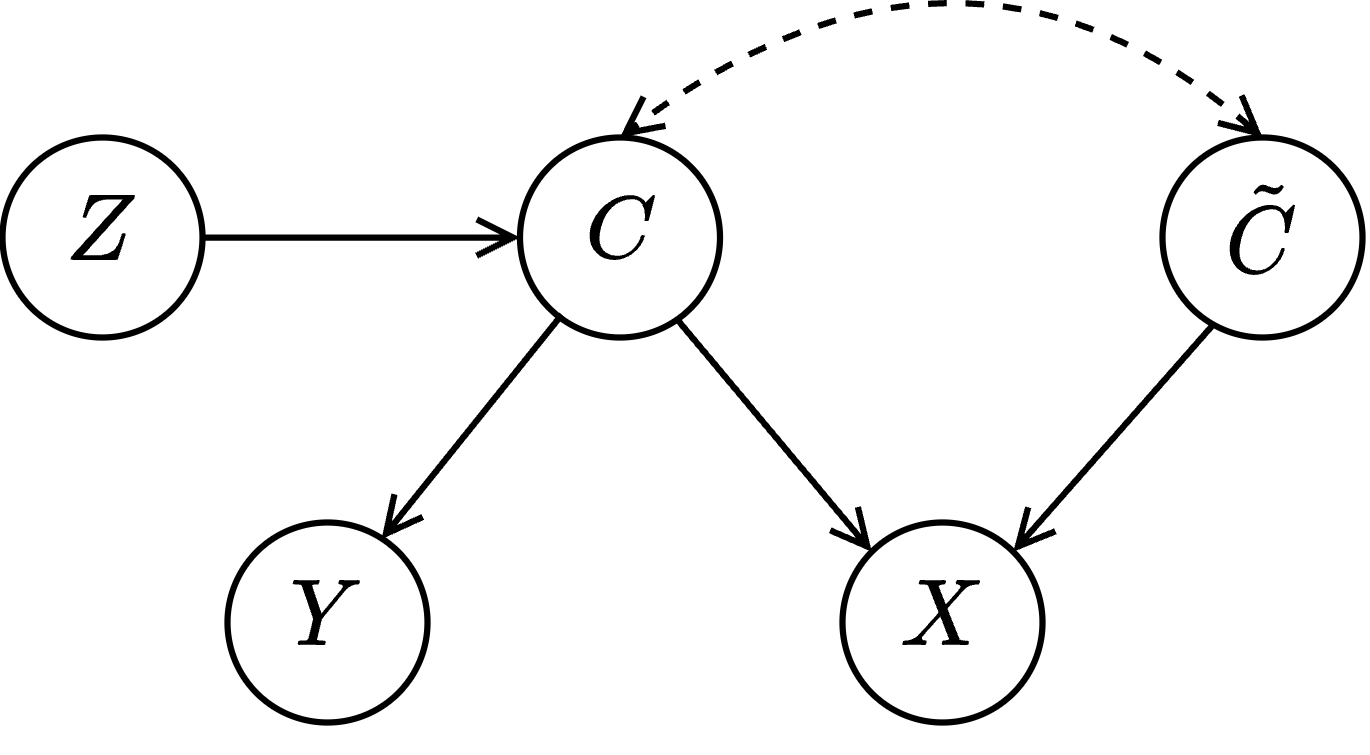} }}%
    \caption{An example of instrument variable. (a): The original graphical model, where $\tilde{C}$ is unobserved. (b): We may use $Z$ as the instrument variable, to get an unbiased estimate of the causal effect of $C$ on $X$.
    }
    \label{fig:instrument_variable}%
\end{figure}

\paragraph{Instrument Variable} It is a variable that is associated with the treatment variable, and influences the outcome variable solely through the treatment variable. The Instrument Variable method is often used to estimate the causal relationship between variables in the presence of unobserved confounders. For example, assume we cannot observe $\tilde{C}$ in the graphical model in Figure~\ref{fig:instrument_variable}. We may choose $Z$ as the instrument variable to estimate the causal effect of $C$ on $X$. 

To achieve this, we first use $Z$ to predict the value of $C$, and then use the estimated value of $C$ to predict the value of $X$, resulting in an unbiased estimator of the causal effect. In the case of linear models, this method is known as the two-stage least squares (2SLS) method \cite{theil1992estimation}. 

\begin{definition}{\footnotesize Instrument Variable}{iv}
\footnotesize
Instrument Variable $Z$ is an exogenous variable introduced such that it affects $X$ and has no independent effect on the outcome variable $Y$. $Z$ should satisfy the following properties:
\begin{enumerate}
    \item $Z$ is associated with $X$, i.e. $P(X|Z) \neq P(X)$;
    \item $Z$ is independent of $Y$ given $X$, i.e. $Z \indep{}.. Y|X$
\end{enumerate}
\end{definition}

Recent works studying IV in machine learning settings have mainly focused on low-dimensional structured data. For example, \cite{pfister2022identifiability} proposed a method to identify sparse causal effect in linear models in presence of unobserved counfounders. They developed graphical criteria for identifiability of the causal effect, and proposed an estimator based on the limited information maximum likelihood (LIML) estimation \cite{anderson1949estimation, amemiya1985advanced}. \cite{wu2022learning} studied the estimation of treatment effect when the data is collected from different sources without access to source labels. They modeled the latent source labels as Group Instrument Variables (GIV), and used a Meta-EM algorithm to iteratively optimize the data representations and the joint distribution for GIV reconstruction.
\cite{kawakami2023instrumental} developed parametric and non-parametric methods to estimate the average partial causal effect (APCE) of a continuous treatment using instrument variables.
\cite{saengkyongam2022exploiting} focused on the independence of the IV with outcome variable conditioned upon input variable. They used this independence to improve the identification and generalization of causal effects using the proposed HSIC-X estimator. 
Recent works \cite{thams2022identifying, mogensen2023instrumental} have studied IV methods for random processes. \cite{thams2022identifying} used the conditional instrument variable method by identifying sets of variables at different lags, while \cite{mogensen2023instrumental} integrated the covariance matrix over time to find moment equations. For high-dimensional data such as image or text, some works \cite{kim2023demystifying, tang2021adversarial, hua2022causal, wang2022causal} have considered adversarial or random perturbations on the input data or features as the instrument variable, to improve the robustness of deep learning models.

\paragraph{Intervention on the feature level} 
In real-world scenarios, it is often difficult to do experimental intervention. Counterfactual data augmentation is also challenging when $\tilde{C}$ is elusive or involves high-level concepts. One line of work instead intervene on the feature level~\citep{jiang2021tsmobn, feder-etal-2021-causalm}, to learn a representation that captures $C$ for the downstream task while providing minimal information about $\tilde{C}$. From a causal perspective, this shares the spirit of ``process control''~\cite{pearl09causality}, i.e., intervening the process influenced by the treatment variable. 
For example, \cite{jiang2021tsmobn} intervene on the feature representation variable by normalizing it for each confounded data point, modifying the representation of all data points with reference to one particular distribution, leaving no effect of the confounder. 

A large body of work on domain adaptation~\citep{tzeng2017adversarial, tzeng2015simultaneous, peng2018zero, bousmalis2016domain, zhuang2015supervised, kim2017learning}, domain generalization~\citep{li2018domain, li2018deep, zhao2020domain, wang2016select,motiian2017unified,carlucci2018agnostic,akuzawa2019adversarial,ge2021supervised,nguyen2021domain,rahman2021discriminative,han2021learning} and 
fairness~\citep{adel2019one,celis2019improved,edwards2015censoring,wadsworth2018achieving, beutel2017data} discussed in Section~\ref{sec:robustness} aligns the distribution of the representation of data at different strata of the confounder $\tilde{C}$, which falls into this category from our causal perspective. Many of these methods do the feature-level intervention in a min-max game, iteratively training a side model to capture the confounder and the main model to be invariant to it, based on Equation~\ref{eq:master:1repeat}
\begin{align} \label{eq:master:1repeat}
    \argmin_{\theta} 
    \dfrac{1}{n}\sum_{(x,y)\in(\X,\Y)}l(f(x;\theta), y)
    - \lambda l(h(f_k(x;\theta);\phi), d), 
\end{align}
When we have pairs of data with the same causal feature $C$, it is often more effective to do sample-wise alignment \cite{mahajan2021domain}. This is connected to data augmentation, if we consider model output as the last-layer feature. For example, assume that $\tilde{C}$ follows a Uniform distribution over the set of possible values $\{\tilde{c}_1, \dots, \tilde{c}_n\}$. Assume $l$ is a loss function that can be considered as a distance metric (e.g. $\ell_p$ loss for $p \geq 1$), so it should satisfy the triangle inequality. Then for $i \neq j$, the below relation holds
\begin{align}
l(f(x_{\tilde{c}_i};\theta), f(x_{\tilde{c}_j};\theta)) < l(f(x_{\tilde{c}_i};\theta), y) + l(f(x_{\tilde{c}_j};\theta), y)
\end{align}
where we use $f(\cdot; \theta)$ to denote the model's output after softmax. 

Summing over all $(i, j)$ pairs results in the following
\begin{align}
\sum_{1<i<j<n} l(f(x_{\tilde{c}_i};\theta), f(x_{\tilde{c}_j};\theta)) < (n-1) \sum_{i=1}^n l(f(x_{\tilde{c}_i};\theta), y)
\end{align}
This shows that minimizing the loss on the augmented data also minimizes an upper bound on the pairwise distance between outputs corresponding to different strata of $\tilde{C}$.

\subsubsection{Backdoor Adjustment} 
\label{sec:backdoor_adjustment}

The backdoor adjustment is a commonly used method to estimate the causal effect of a treatment variable $X$ on an outcome variable $Y$. By conditioning on a set of properly chosen variables $Z$ (``adjustment variables"), we can remove their confounding effect and get the causal effect of $X$ on $Y$ from observational data alone without actually conducting the intervention. To achieve this, $Z$ need to satisfy a set of conditions, often known as the \emph{backdoor criterion}. The backdoor criterion and the adjustment formula are defined in \ref{theorem:ba}.

\begin{definition}{\footnotesize Backdoor Adjustment}{ba} \label{box:backdoor_adjustment}
\footnotesize
A set of variables $Z$ satisfies the Backdoor criterion 
relative to $\{X, Y\}$ in a DAG, 
if no node in $Z$ is a descendant of $X$,
and 
$Z$ blocks every path between $X$ and $Y$ that contains an arrow into $X$.
Then the causal effect of $X$ on $Y$ is given by
\begin{equation} \label{eqn:backdoor}
    P(Y=y|do(X=x)) = \sum_{z}P(Y=y|X=x,Z = z)P(Z = z)
\end{equation}

\end{definition}

The backdoor criterion can be intuitively understood from its graphic implications on the SCM. $Z$ should be chosen such that:
\begin{itemize}
    \item It blocks all spurious paths between $X$ and $Y$;
    \item It leaves all directed paths from $X$ to $Y$ unperturbed;
    \item It doesn't create new spurious paths. 
\end{itemize}

These three conditions ensure that conditioning on $Z$ blocks and only blocks the spurious paths between $X$ and $Y$. 
For a formal proof of Equation~\ref{eqn:backdoor}, please refer to~\cite{pearl09causality}.

\begin{figure}[h]
    \centering
    \subfloat[\centering Backdoor]{{\includegraphics[width = 0.3\textwidth]{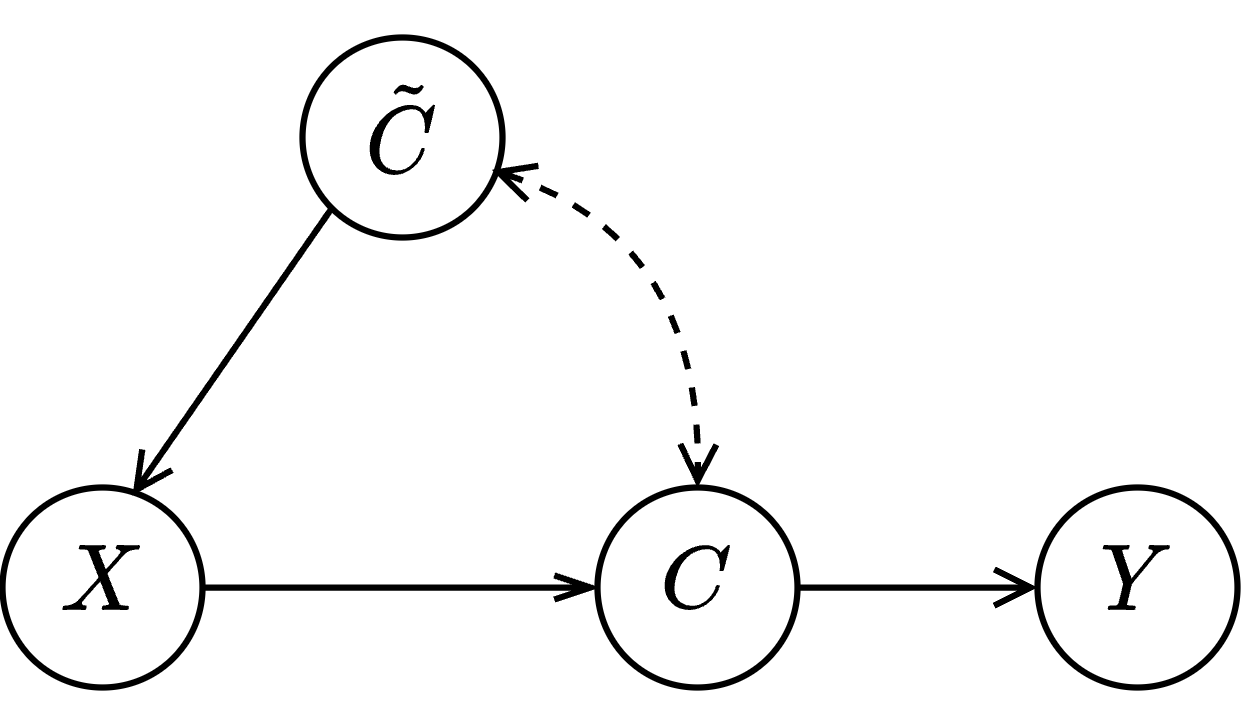} }}
    \hspace{0.1\textwidth}
    \subfloat[\centering Frontdoor]{{\includegraphics[width=0.32\textwidth]{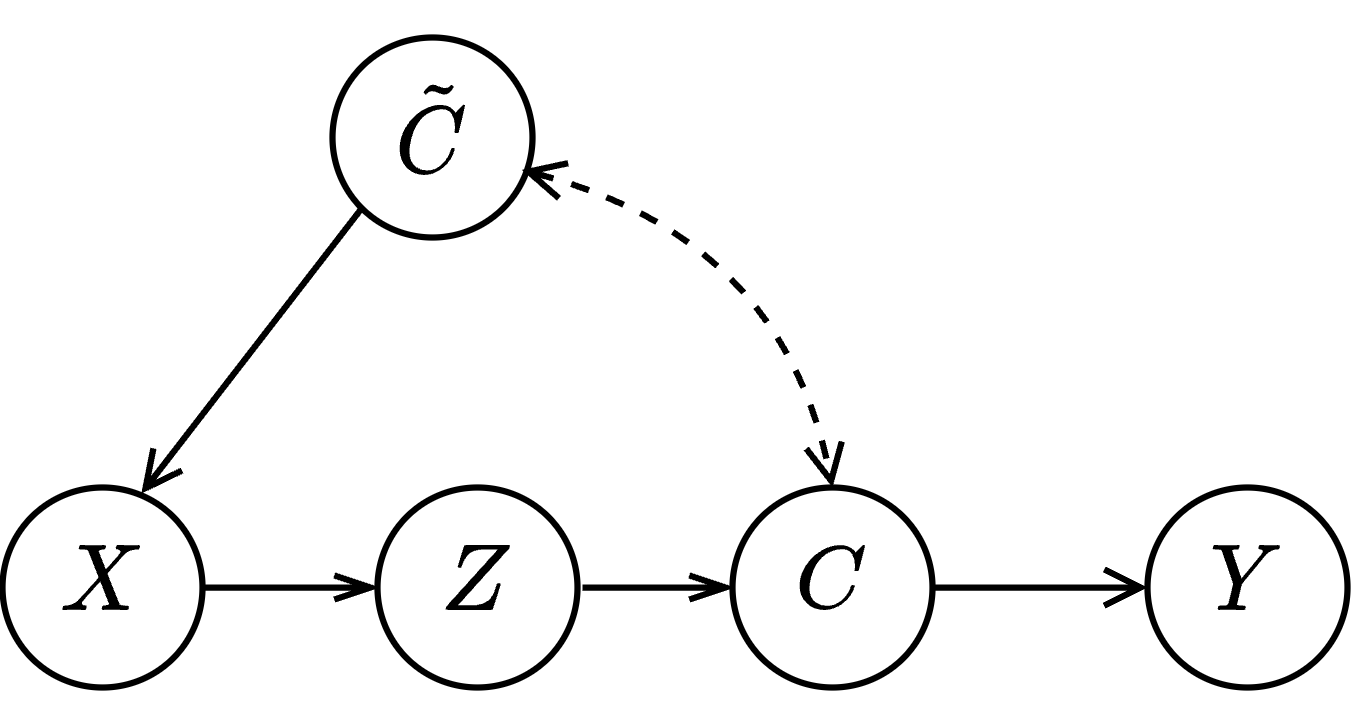}}}%
    \caption{Graphical models of the data generation processes underlying recent machine learning methods that use backdoor and front-door adjustment.}
    \label{fig:adjustment}%
\end{figure}

Based on the causal assumptions encoded in the graphical model in Figure~\ref{fig:comparison}(a) and discussed in Section~\ref{sec:data_to_prediction}, it is difficult to directly apply backdoor adjustment to machine learning tasks because $C$ is unobserved. 
Instead, recent works using adjustment-based methods have implicitly made causal assumptions with a different conception of the causal feature $C$: instead of the latent factors that generate $X$, they consider $C$ as the information conveyed in and decoded from $X$. This results in a reversal of the causal link between $C$ and $X$ and a graphical model in Figure~\ref{fig:adjustment}(a). Now their is a causal relation between $X$ and $Y$ through the path $(X,C,Y)$, and $\tilde{C}$ satisfies the backdoor criterion relative to $\{X, Y\}$. This eliminates the need to model $C$ and makes it more practical to apply adjustment methods to high-dimensional data such as image. 

Because backdoor adjustment provides a principled way to remove the confounding effect without interventional experiments, it has gained popularity in recent years in machine learning research \cite{9643182, CIActionUnit2022, 10.1145/3474085.3475540, shao2021improving, Zhang2020CausalIF, chen2021unbiased, Wang2021CausalAF, yue2020interventional, deng2021comprehensive, wang2021weaklysupervised, wang2020visual, Chen2021DependentML}. These works studied problems from diverse background but with a common aim to learn the causal effect of the input $X$ on the output $Y$. 
In these methods, the first step is to identify the confounder and its distribution in the dataset. For example, \citep{CIActionUnit2022} found that the individual habits of facial muscle movement is a confounder for facial action unit recognition, and \cite{Zhang2020CausalIF} found that the co-occurrence relationship among objects is a confounder for image classification models to produce correct activation maps. $\tilde{C}$ is often assumed to be a discrete variable with a set of possible values
$\{\tilde{c}_j\}_{j = 1}^{m}$, each corresponding to a class label or a sample group (such as facial images of the same individual). $\tilde{C}$ is often assumed to conform to the Uniform distribution, or a distribution estimated from the training set.

Next, the model architecture is modified to incorporate $\tilde{C}$ as a covariate, getting an estimator of $P(y|x,\tilde{c}_j)$ as below
\begin{align}
\hat{P}(y|x,\tilde{c}_j) = f_y((x, \alpha(x, \mathbf{\tilde{c}}_j) \mathbf{\tilde{c}}_j);\theta)
\end{align}
where $f_y(\cdot; \theta)$ is the output probability of the model corresponding to label $y$, $\mathbf{\tilde{c}}_j$ is the representation of ${\tilde{c}_j}$, and $\alpha(x, \mathbf{\tilde{c}}_j)$ is the weight assigned to $\tilde{c}_j$ for the sample $x$, representing the probability that $x$ belongs to the stratum $\tilde{c_j}$.

Then, backdoor adjustment is applied to get an estimator of $\hat{P}(y|do(x))$
\begin{align}
\hat{P}(y|do(x)) = \sum_{j=1}^m \hat{P}(y|x,\tilde{c}_j)P(\tilde{c}_j)
= \sum_{j=1}^m f_y((x, \alpha(x, \mathbf{\tilde{c}}_j) \mathbf{\tilde{c}}_j);\theta) P(\tilde{c}_j)
\end{align} \label{eqn:backdoor_ml}

To improve the efficiency, the Normalized Weighted Geometric Mean \cite{xu2015show} is often used to move the outer summation into the feature level.
\begin{align}
\hat{P}(y|do(x)) = f_y\left(x, \sum_{j=1}^m \alpha(x, \mathbf{\tilde{c}}_j) P(\tilde{c}_j) \mathbf{\tilde{c}}_j;\theta\right) 
\end{align}

Finally, the model is trained with maximum likelihood estimation, based on the following equation
\begin{align}
    \argmin_{\theta} 
    \dfrac{1}{n}\sum_{(x,y)\in (\X,\Y)} l\left(f\left(x, \sum_{j=1}^m \alpha(x, \mathbf{\tilde{c}}_j) P(\tilde{c}_j) \mathbf{\tilde{c}}_j;\theta\right), y\right) 
    \label{eq:master:backdoor:0}
\end{align}

Following the early work \cite{Zhang2020CausalIF, wang2020visual, yue2020interventional}, different implementation choices have been made to adapt to different problem settings. To get the representation $\{\mathbf{\tilde{c}}_j\}_{j = 1}^{m}$, a common method is to average the features of all samples corresponding to the same $\tilde{c}_j$
\cite{9643182, CIActionUnit2022, 10.1145/3474085.3475540, Zhang2020CausalIF, shao2021improving, chen2021unbiased, wang2021weaklysupervised, wang2020visual, nan2021interventional, yue2020interventional}. 
This approach helps to gain the average features containing the confounding information, even if the confounders are unobservable and there is little knowledge about them. However, it does not explicitly separate causal from non-causal features. To overcome this limitation, recent works such as \cite{shao2021improving, chen2021unbiased, deng2021comprehensive, qi2020causal, wang2020visual} modify this methodology by only including context (e.g. background) features
corresponding to each group. 
For instance, \cite{shao2021improving} used class activation maps (CAMs), \cite{qi2020causal} built the confounder representation based on query type in a multi-model scenario,
\cite{wang2020visual} took the arithmetic average of the region of interest features of associated objects, while \cite{deng2021comprehensive} took weighted average of the sample features based on output probability of the model.
In addition, \cite{9656623} used the observed confounders to training an unbiased classifier, which was then used to stratify the remaining confounders. 
\cite{9607759, chadha2021ireason} did not stratify the confounders, but identified them based on temporal dependencies between confounders and other features. 

Regarding the weight $\alpha(x, \mathbf{\tilde{c}}_j)$, \cite{nan2021interventional, wang2021weaklysupervised} took the simple approach to set $\alpha=1$, \cite{CIActionUnit2022, 10.1145/3474085.3475540, 9643182,Zhang2020CausalIF, wang2020visual} used an attention mechanism \cite{vaswani2017attention} to model the alignment of the sample to the confounder, while \cite{yue2020interventional, deng2021comprehensive} used the model's output probability corresponding to the label. In addition, there are different ways to fuse the confounder representation with the sample, such as concatenation \cite{Zhang2020CausalIF}, simple addition \cite{nan2021interventional, qi2020causal}, or they can be processed by different layers before added together \cite{wang2021weaklysupervised, CIActionUnit2022}. 

Before we conclude this section, it is worth mentioning that
according to \cite{9656623}, some of the non-causal features may provide useful contextual information for an image, which benefit the generalization of a machine learning model. They suggest retaining these features in backdoor adjustment. 

\subsubsection{Frontdoor Adjustment}
\label{sec:frontdoor_adjustment}
In the above text, we discussed adjustment variables and how we require a backdoor criterion to ensure that we are able to adjust the right variables, to estimate the true effect of the treatment on the outcome. However, there might be some cases where the backdoor criterion does not get satisfied, such as when the confounding variable is unobserved. In these cases, front-door adjustment can be used to de-confound the model.

Front-door adjustment method works using the two consecutive applications of backdoor adjustment to estimate the causal effect of $X$ on $Y$ (i.e., $P(Y=y|do(X=x))$). 
It introduces a variable $Z$ that is a mediator between $X$ and $Y$, with no backdoor path from $X$ to $Z$. 
This means that the correlation between $Z$ and $X$ is equal to the causal effect from $X$ to $Z$, i.e., $P(Z|do(X=x)) = P(Z|X)$. The front-door adjustment chain together two partial effects, i.e. $X$ on $Z$ and $Z$ on $Y$ to estimate the overall causal effect of $X$ on $Y$, as given in Equation~\ref{eqn:frontdoor_causal}. 
\begin{equation} \label{eqn:frontdoor_causal}
    P(Y = y|do(X = x)) = \sum_z P(Y|do(Z = z)) P(Z|do(X = x))
\end{equation}

We can write the expression $P(Z|do(X = x)) = P(Z|X)$ as described above. Meanwhile, for the other partial effect expression $P(Y|do(Z = z))$, we can observe a backdoor path between $Z$ and $Y$, i.e. $(Y, C, \tilde{C}, X, Z)$. 
Therefore, we can properly quantify the effect of $Z$ on $Y$ only if the backdoor path is blocked between $Z$ and $Y$. 
This can be achieved by adjusting for the variable $X$, which is observable in our backdoor path, arising the expression $P(Y|do(Z)) = P(Y|Z = z, X = x')P(X = x')$ as represented in the Equation~\ref{eqn:frontdoor_mod}.

\begin{equation} \label{eqn:frontdoor_mod}
    P(Y = y|do(X = x)) = \sum_z P(Z = z|X = x) \sum_{x'} P(Y|Z = z, X = x')P(X = x')
\end{equation}

Based on the graphical model in Figure~\ref{fig:adjustment}(b), the intermediate feature of the model $f_k (X; \theta)$ is considered as the mediator $Z$~\citep{Yang2021CausalAF, li2021confounder}.
Compared to the \ref{sec:backdoor_adjustment}, methodologies that use the front-door adjustment do not require the observation of confounders. 
This enables some recent works ~\citep{Yang2021CausalAF, li2021confounder} to use the observed data examples to eliminate spurious patterns, where the intermediate feature is taken as the mediator. 
\cite{Yang2021CausalAF} proposed a causal attention framework. They used Normalized Weighted Geometric Mean (NWGM) approximation~\cite{xu2015show} to absorb the outer summation on $Z$ and $X$ into the feature level, and used in-sample and cross-sample attention mechanisms to calculate embeddings for $Z$ and $X$ respectively. 
\cite{li2021confounder} used the gradient information of each example $X$ to model its confounding effect on $Z \rightarrow Y$, and used a clustering-based method to efficiently estimate Equation~\ref{eqn:frontdoor_mod} on the whole dataset.

\subsubsection{Inverse Probability Weighting} \label{sec:ipw}
In \ref{sec:backdoor_adjustment} and \ref{sec:frontdoor_adjustment} we have introduced two adjustment approaches to estimate the causal effect of $X$ on $Y$ in the presence of a confounder $Z$. However, both methods require considering each value or combination of values $z$ and estimating $P(y|x,z)$ separately, which may bring practical challenges. First, if the set of possible values is large, it is computationally expensive to estimate all the associated conditional probabilities. Second, some combinations of $(x,z)$ may be missing or scarce in the dataset, in which case it is difficult to give a reliable estimate of $P(y|x,z)$. Inverse probability weighting is an alternative approach that creates a pseudo-population where the confounder is independent of the treatment variable. Assume, for example, that $Z$ satisfies the backdoor criterion relative to $\{X, Y\}$ in a graphical model.

\begin{align} \label{eq:weighting}
\begin{split}
P(y | do(x)) &= \sum_{z} P(y | x, z) P(z) = \sum_{z} \frac{P(y | x, z) P(x | z) P(z)}{P(x | z)} \\
&= \sum_{z} \frac{P(y, x, z)}{P(x | z)}
\end{split}
\end{align}

As comparison, 
\begin{equation}
P(y | x) = \sum_{z} \frac{P(y, x, z)}{P(x)}
\end{equation}

The interventional distribution $P(y | do(x))$ differs from the original distribution $P(y | x)$ in that it replaces the constant $P(x)$ on the denominator with $P(x | z)$.
$P(x | z)$ is often referred to as the ``propensity score",
which captures how likely the treatment variable is given the confounder. 
If we can reliably estimate the propensity score (often using a parameterized model), 
we can weight each datapoint to remove the confounding effect of $z$. This approach provides lower variance estimation than adjustment methods when the distribution of $\tilde{C}$ is complicated and the dataset is relatively small.

A line of works discussed in Section~\ref{sec:robustness} weight data examples to counter spurious features for robustness \cite{hu2018does, sagawa2019distributionally, namkoong2016stochastic}, or improve learning on under-represented subpopulation for fairness \cite{goel2018non, jiang2020identifying, krasanakis2018adaptive, hu2018does, nam2020learning, Sagawa2020Distributionally}, which fall into this category of methods from a causal perspective. Based on our assumptions in the graphical model in Figure~\ref{fig:comparison}, $\tilde{C}$ satisfies the backdoor criterion relative to $\{C, X\}$. And according to our previous discussion on Equation~\ref{eq:xyc:0}, \ref{eq:xyc:1}, estimating $P(x|do(c))$ removes the confounding between $X$ and $Y$. Following Equation~\ref{eq:weighting},
\begin{align}
P(x | do(c)) = \sum_{\tilde{c}} \frac{P(x, c, \tilde{c})}{P(c | \tilde{c})}
\end{align}

\begin{wrapfigure}{r}{0.4\textwidth}
\footnotesize
\begin{tcolorbox}[colback=white!90!gray, colframe=white!70!black,left=1pt, right=1pt, top=0.5pt, bottom=0.5pt]
\textbf{Intuition}: A larger propensity score indicates that the causal feature of the datapoint is more easily explained by the confounder, or the datapoint is ``bias-aligned''. Otherwise, the datapoint is said to be ``bias-conflicting''. By giving more weights to the bias-conflicting samples, we balance the distribution and remove the bias in the dataset.
\end{tcolorbox}
\end{wrapfigure}

Consider the relationship $P(y|\tilde{c}) = \sum_{c} P(c|\tilde{c})P(y|c)$. 
For a fixed $y$, a higher $P(y|\tilde{c})$ implies higher propensity scores between $\tilde{c}$ and the causal feature value $c$'s corresponding to the label $y$. 
This tells us that $P(y | \tilde{c})$ is a good surrogate for $P(c | \tilde{c})$ to weight the data samples, which is very useful because $C$ is usually not observed. We can train a side model $h(\cdot;\phi)$ to estimate $P(y | \tilde{c})$, and train the main model $f(\cdot; \theta)$ on the weighted training data, where smaller weights are given to datapoints whose labels are more easily predicted from the bias, and vice versa. This gives us the following equation

\begin{align}
    \argmin_{\theta} 
    \dfrac{1}{n}\sum_{(x,y)\in (\X,\Y)} \frac{1}{h_y(x;\phi)} l(f(x;\theta), y),
    \label{eq:weighting_master}
\end{align}
where $h_y(\cdot; \phi)$ is the element of $\operatorname{Softmax}(h(\cdot; \phi))$ corresponding to label $y$. Similar ideas may be implemented in different ways, as in \citep{dagaev2023too, nam2020learning}. One advantage of weighting-based approach is that it does not require explicit modeling of the distribution of $\tilde{C}$. Instead, the assumptions about the distribution of $\tilde{C}$ is encoded in the architecture of $h(\cdot;\phi)$ or its learning algorithm.

\subsection{Counterfactuals: the third level} \label{sec:counterfactuals}
\mch{\begin{enumerate} 
    \item We cannot bind the works of counterfactuals into any equation, especially the ones that are above. However, we can write that they belong to the equation: $\mathbb{E}[Y_{\{X = 0\}} = 0|Y = 1, X = 1]$. However, they do not consider identifying the exogenous variables $U$ as they might consider that $U$ remains constant for all the samples. 
\end{enumerate}}

In this section, we will explore counterfactuals, which are considered the third level of causation $\mathcal{L}_3$.  
Typically, counterfactuals involve a hypothetical scenario or antecedent, 
where the question is posed with ``if'', and the condition after ``if'' may be untrue and contradicts the observed event. 

\begin{definition}{\footnotesize Counterfactuals}{cts}\label{box:counterfactual}
\footnotesize    
    Counterfactual analysis deals with the assessment of events that would have happened under an alternative condition $X=x'$, given that the event has already occurred under the actual condition $X=x$ with the outcome $Y=y$. This can be defined mathematically as Equation~\ref{eqn:counterfactual}
    \begin{equation}     \label{eqn:counterfactual}
        \mathbb{E}(Y_{X = x'}|Y = y, X = x)
    \end{equation}
    There are three major steps in estimating a counterfactual scenario:
    % \hwc{just use itemize for the formatting}

        \begin{description}
            \item[Abduction-] ``given the fact that $X = x$ and $Y = y$'', i.e. the observed values of endogenous variables in $V$ are used to infer the posterior distribution of exogenous variables $U$.
            \item[Action-] ``had $X$ been $x'$'', i.e. the causal model $\mathcal{M}$ is modified by replacing the structural equations for $X$ with adequate functions making $X = x'$, resulting in a modified model $\mathcal{M}_{x'}$
            \item[Prediction-] ``what Y would have been'', i.e. the modified model $\mathcal{M}_{x'}$ and the inferred distribution of exogenous variables $U$ are used to compute the counterfactual outcome $Y_{x'}$.
        \end{description}

\end{definition}

\subsubsection{Data augmentation} \label{sec:counterfactuals:da}
Counterfactual analysis is often used in data augmentation methods. Based on our assumptions in the graphical model in Figure~\ref{fig:comparison}, when we generate additional samples by perturbing the non-causal features $\tilde{C}$, we are essentially answering the causal question of ``what the input $X$ would have been had $\tilde{C}$ been set to a different value, with everything else been the same''. If we follow the three steps of counterfactual analysis, but assign to $\tilde{C}$ a distribution $P(\tilde{C})$ instead of a constant at the ``action'' step, we get the following loss function on the augmented data:
\begin{align}
    \mathcal{L}_{\textrm{aug}} = 
    \mathop{\mathbb{E}}_{(x, y) \sim P(X, Y)}\left[\mathop{\mathbb{E}}_{u \sim P(U|x,y)} \left[\mathop{\mathbb{E}}_{\tilde{c} \sim P(\tilde{C})} \left[l(f(X_{\mathcal{M}_{\tilde{c}}}(u);\theta), y)\right]\right]\right]
    \label{eq:ctft:1}
\end{align}
where $u$ is a realization of the exogenous variables, $u=(u_X, u_Y, u_C, u_{\tilde{C}})$. $X_{\mathcal{M}_{\tilde{c}}}(u)$ is the value of $X$ in the intervened causal model $\mathcal{M}_{\tilde{c}}$ at $U=u$, $X_{\mathcal{M}_{\tilde{c}}}(u) = \mathcal{F}_X(\{c, \tilde{c}\}, u_X) = \mathcal{F}_X(\{\mathcal{F}_C(\emptyset, u_C), \tilde{c}\}, u_X)$, where $\mathcal{F}_X$ and $\mathcal{F}_C$ denote the structural equations for $X$ and $C$ respectively. 

Recent generative methods for data augmentation, such as those based on Variational Autoencoders (VAEs) \cite{kingma2022autoencoding}, often use an encoder to estimate $P(U|x)$ and a decoder to generate $X_{\mathcal{M}_{\tilde{c}}}(u)$. 
Further, if we assume that $U$ can be uniquely determined by $X$, we can denote their mapping as $U = \mathcal{U}(X)$. Then every pair of $(x, \tilde{c})$ uniquely determines a counterfactual sample, denoted as $x_{\tilde{c}}$, by the equation $x_{\tilde{c}} = X_{\mathcal{M}_{\tilde{c}}}(\mathcal{U}(x))$. Equation~\ref{eq:ctft:1} can be simplified as below:
\begin{align} \label{eq:ctft:2}
\begin{split}
    \mathcal{L}_{\textrm{aug}} &=  
    \mathop{\mathbb{E}}_{(x, y) \sim P(X, Y)} \left[\mathop{\mathbb{E}}_{\tilde{c} \sim P(\tilde{C})} \left[l(f(X_{\mathcal{M}_{\tilde{c}}}(\mathcal{U}(x));\theta), y)\right]\right] \\
    &= \mathop{\mathbb{E}}_{(x, y) \sim P(X, Y)} \left[\mathop{\mathbb{E}}_{\tilde{c} \sim P(\tilde{C})} \left[l(f(x_{\tilde{c}};\theta), y)\right]\right]
    \end{split}
\end{align}
which corresponds to the training objective in Equation~\ref{eq:cda:rct}.

\begin{wrapfigure}{r}{0.4\textwidth}
\footnotesize
\begin{tcolorbox}[colback=white!90!gray, colframe=white!70!black,left=1pt, right=1pt, top=0.5pt, bottom=0.5pt]
\textbf{Intuition}: In data augmentation, we are implicitly following the three steps of counterfactual analysis to find out what the input object would have been under a different condition. For example, what if the turtle is placed into a bird nest before taking photos. 
\end{tcolorbox}
\end{wrapfigure}

\begin{figure}[h]
    \centering
    \subfloat[\centering Similarity between Generative Adversarial Network (GAN) and a Structural Causal Model (SCM)]{{\includegraphics[width=0.35\textwidth]{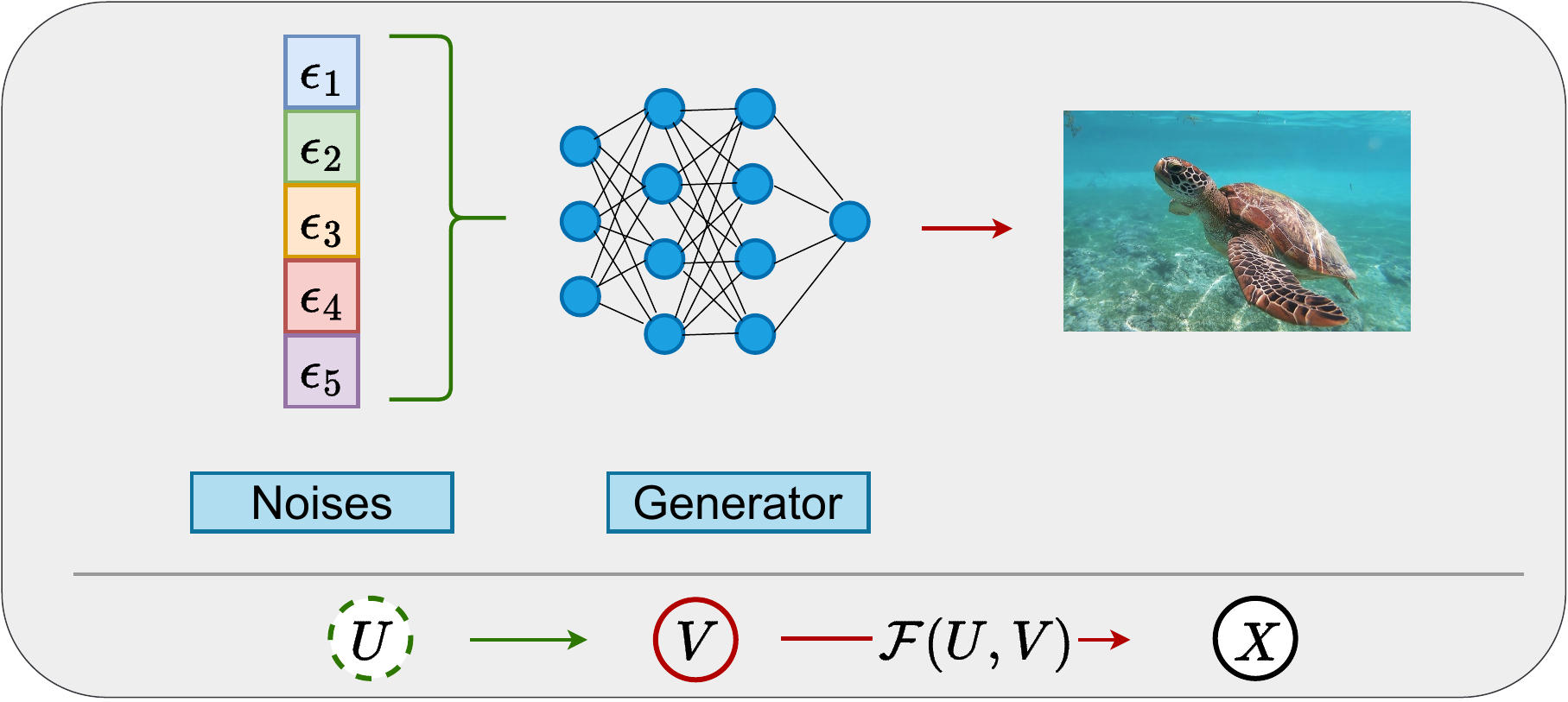}}}%
    \qquad
    \subfloat[\centering Similarity between Variational Auto-encoder (VAE) and a Structural Causal Model (SCM)]{{
    \includegraphics[width=0.55\textwidth]{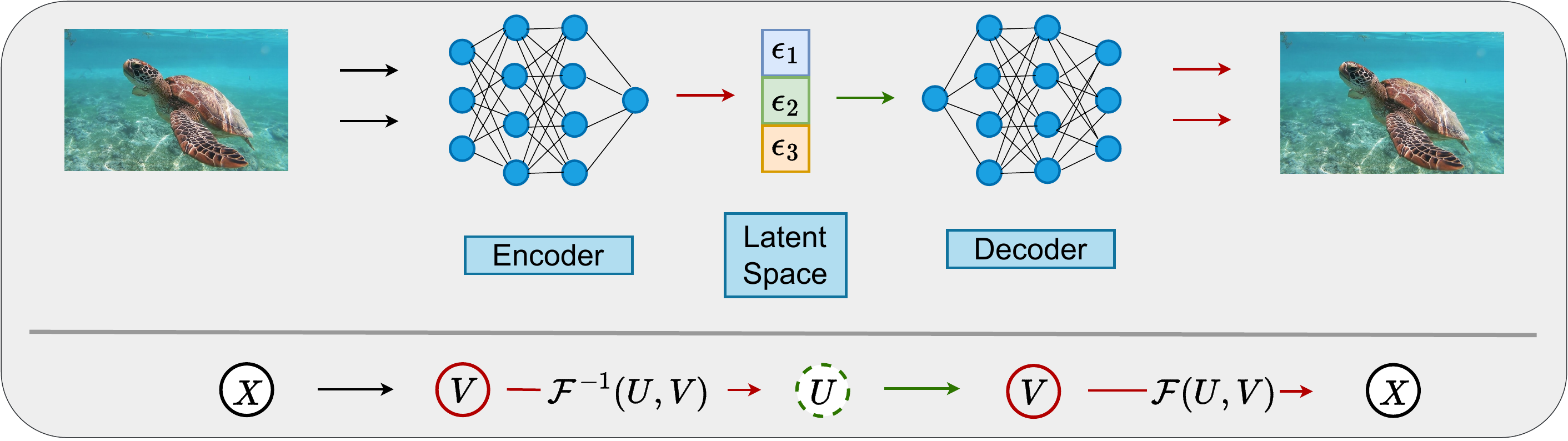}}}
    \caption{Similarities between SCM and latent variable models, with GAN and VAE as examples.
    (a) demonstrates the similarity between GAN and SCM, where exogenous variables ($U$) correspond to noise used by the GAN to generate the output. Meanwhile, endogenous variables ($V$) are neurons of the network and the image is the generated output ($X$). (b) Similarly, for VAE, exogenous variables ($U$) correspond to the latent vector \cite{suter2019robustly} extracted by the encoder. The encoder acts as an inverse function ($\mathcal{F}^{-1}(\cdot)$) of the structural equations, and the decoder uses the latent vector to reconstruct the input variable $X$. 
    }
    \label{fig:generative_exogenous}%
\end{figure}

Many deep learning techniques use generative \cite{sauer2021counterfactual} and latent variable-based methods \cite{suter2019robustly, yang2021causalvae, pawlowski2020deep} to extract the exogenous variables. 
These generative models, such as GAN and VAEs, treat the exogenous variables as noises in their respective models, as shown in the Figure~\ref{fig:generative_exogenous}. 
The works based on them intervene on these noises to generate a counterfactual sample in the majority of the scenarios.
\cite{suter2019robustly, yang2021causalvae} used a variational autoencoder to extract independent causal mechanisms or exogenous variables in a latent space, converting them to endogenous variables.
\cite{suter2019robustly, sauer2021counterfactual} were based on the assumption of ``Independent Causal Mechanism'' (ICM)
\cite{scholkopf2012causal, lemeire2006causal}, 
which states that ``the causal generative process of variables in a system is composed of autonomous modules that do not inform or influence each other''\cite{peters2017elements}. 
In ICM, ``independent'' does not mean that different ICMs are statistically independent of each other, but that they do not causally influence each other, i.e. intervening on one mechanism will not have any effect on other mechanisms. 
Additionally, \cite{pawlowski2020deep} 
performed the abduction step using normalizing flows and variational inference, and then intervened on the obtained exogenous values as shown in Figure~\ref{fig:generative_exogenous}(a).

Some other works \cite{Boukhers2022COINCI, eckstein2021discriminative, goyal2019counterfactual, Mertes2022GANterfactualCounterfactualEF, oh2021born, vermeire2020explainable, wang2020scout, zhao2020fast, parafita2019explaining} have used generative or latent-space based models to generate counterfactual images belonging to a different target class.
They achieved this by minimally perturbing the causal features of the original image, and some also used text associated with the original image \cite{ijcai2020-742}. 
This procedure aims to identify the features treated as causal to categorize the sample into the original class.  
The works \cite{vermeire2020explainable, white2021contrastive} segregated the image into segments to improve the perturbation quality and let the image remain in the prior distribution. 
\cite{white2021contrastive} introduced the concept of causal over-determination, 
which probes two or more causal factors. 
Generative adversarial network (GAN) \citep{goodfellow2014generative} is the most widely used technique to generate counterfactual samples in this direction \cite{wang2023ad, Boukhers2022COINCI, eckstein2021discriminative, Mertes2022GANterfactualCounterfactualEF, oh2021born, ijcai2020-742, zhao2020fast, parafita2019explaining, smith2020counterfactual}.

In addition to latent-space based methods, some works conducted minimal perturbation on the pixel level to generate counterfactual samples \cite{goyal2019counterfactual, jung2021counterfactual, holtgen2021deduce}. 
Aside from explaining the network behavior to predict a sample in a class, 
\cite{holtgen2021deduce, smith2020counterfactual} focused on generating a counterfactual sample to correct the model's prediction.
Rather than relying on the original latent space of the generative models, \cite{yang2021causalvae} introduced an SCM as a prior by adding a causal layer, which maps the exogenous variables to endogenous variables. 
Elements in this layer can be intervened with $do-$calculus to generate the counterfactual samples. 

Furthermore, a plethora of works \cite{parafita2019explaining, article, rodriguez2021trivial, yang2021generative, hvilshoj2021ecinn} generated counterfactual samples for explainability.
\cite{parafita2019explaining} generated counterfactuals to increase the explainability of the network to identify the network's behavior under some specific conditions not present priorly in the dataset. 
\cite{rodriguez2021trivial} aimed to find spurious features that produce non-trivial explanations. 
\cite{yang2021generative} generated counterfactuals for attribute informed latent space. 
\cite{hvilshoj2021ecinn} used an invertible network to identify how inputs can be altered to change the prediction of a classifier, and generated the changed input image to provide a visual explanation for prediction change. 
\cite{oh2021born} proposed a method to generate a multi-way counterfactual explanation, i.e., change the attributes of the sample with respect to any class resulting in a change in the prediction of the sample. 

As alternative methods, some recent works treat causal features as endogenous features, and non-causal features as exogenous features.
\cite{zevcevic2022pearl} proposed to generate images by considering the non-causal properties in the image as the exogenous variable $U$. 
\cite{yue2021counterfactual, https://doi.org/10.48550/arxiv.2108.02093} generated new counterfactual images by extracting the exogenous values using the three-step process defined in \ref{box:counterfactual}. 
They substituted the value of an object with another object.
\cite{9156448} extracted the exogenous variables from samples to generate counterfactual scenarios.
Several other works \cite{zhang2020counterfactual, liang-etal-2020-learning, chang2021robust, 9706896, chen2020counterfactual, https://doi.org/10.48550/arxiv.2106.08914} used the term factual to describe the process of augmenting a sample by removing non-causal features, and counter-factual to describe the process of retaining the non-causal features after removing the causal features. 
The goal of this augmentation is to let the machine learning model associate its output only with causal features and disregard any dependence on non-causal features. 
In addition, \cite{fu2020counterfactual, plumb2021finding, yi2020clevrer, sauer2021counterfactual, rosenberg2021vqa} also aimed to generate counterfactual samples to expand the observed dataset for robust learning.
They achieved it by leveraging various techniques such as 
adversarial learning to interpolate complex samples \cite{fu2020counterfactual}, diffusion-based data-generation \cite{sauer2021counterfactual}, and inserting scenarios in the dataset containing counterfactual situations \cite{yi2020clevrer}. 
\cite{plumb2021finding,rosenberg2021vqa} did binary classification on the existence of an object in the image.
\cite{rosenberg2021vqa} employed the process of counterfactual generation in two steps: defining the sample into different causal categories; and replacing the causal variable in a sample with that of another sample.

\subsubsection{Treatment Effects} \label{sec:counterfactuals:te}
The term \emph{treatment effect} refers to the difference in potential outcomes resulting from the application of a specific treatment or intervention. 
Since it is not possible for the same unit to be both treated and untreated,  
counterfactual analysis is often required in observational studies to estimate the potential outcomes. 

\begin{figure}[h]
    \centering
    \subfloat[\centering]{{\includegraphics[width = 0.26\textwidth]{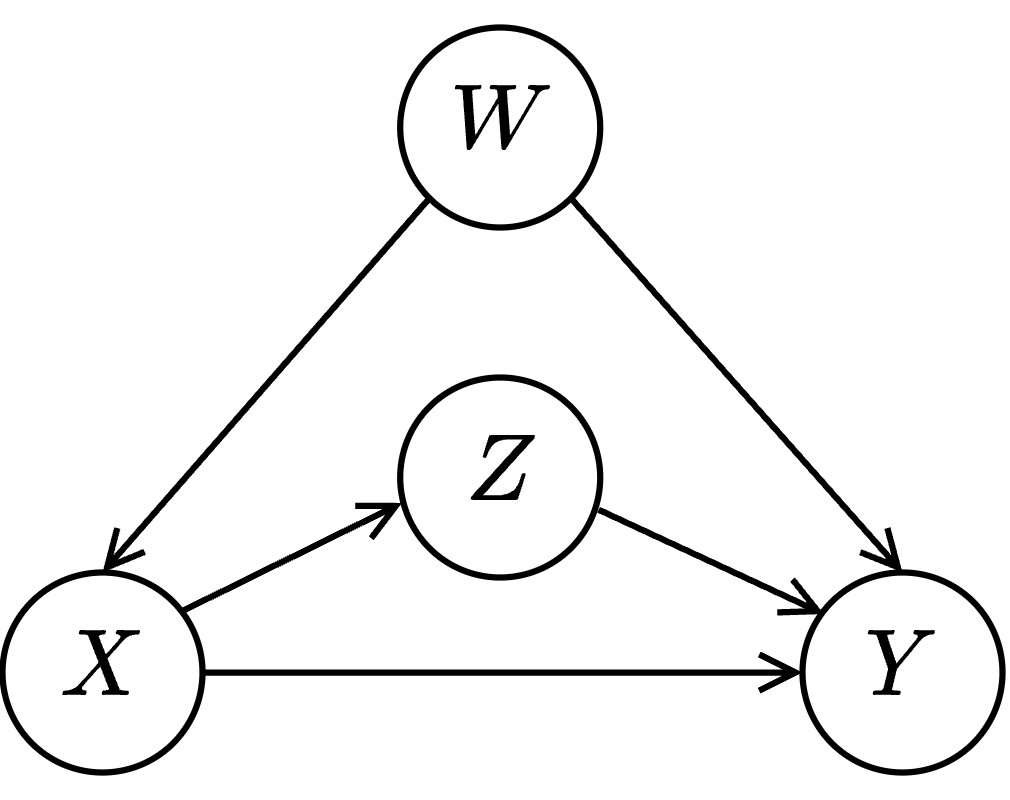} }}
    \hspace{0.1\textwidth}
    \subfloat[\centering \centering]{{\includegraphics[width=0.32\textwidth]{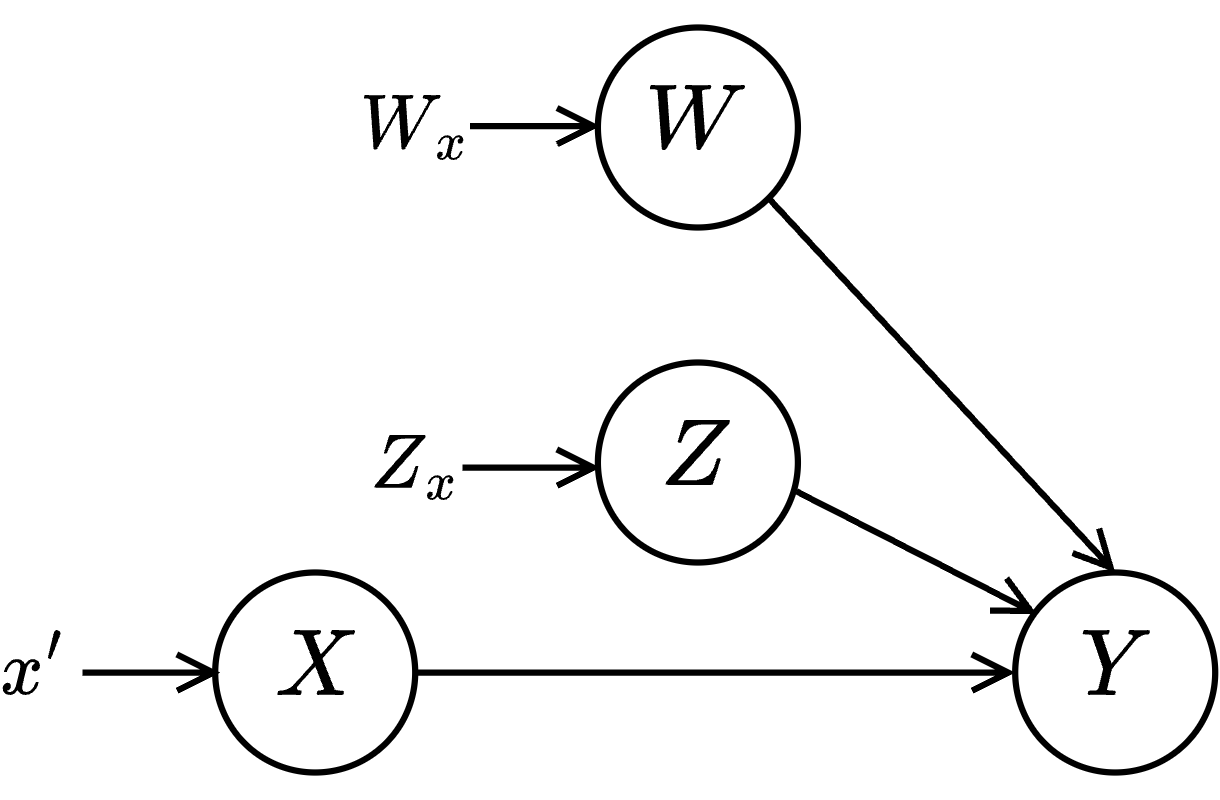} }}%
    \caption{Graphical models as an example of mediation, and how the natural direct effect is measured. (a): $X$ influences $Y$ through the paths $(X,Y)$ and $(X,Z,Y)$. $W$ is an arbitrary confounder. (b): When we change $X$ from $x$ to $x'$, we hold $\{Z, W\}$ to their pre-treatment distribution, and only transmit the effect of the change through the path $(X, Y)$.}
    \label{fig:lmt_treatment}%
\end{figure}

There are different types of treatment effects such as Natural Direct Effect (NDE), Natural Indirect Effect (NIE), and Total Effect (TE). Consider the graphical model in Figure~\ref{fig:lmt_treatment}(a), in which $X$ has influence on $Y$ both directly and through a mediator $Z$. Regarding the change in $Y$ induced by changing $X$ from $x$ to $x'$, different treatment effects can be defined using the counterfactual notation as below \cite{glymour2016causal}:

\begin{definition}{\footnotesize Treatment Effects}{te}
\footnotesize

\begin{equation} \label{eqn:treatment_effect_nde}
    \textsf{NDE}_{x,x'}(Y) = \mathbb{E}[Y_{x', M_x} - Y_{x,M_x}]
\end{equation}
\begin{equation} \label{eqn:treatment_effect_nie}
    \textsf{NIE}_{x,x'}(Y) = \mathbb{E}[Y_{x, M_{x'}} - Y_{x,M_x}]
\end{equation}
\begin{equation} \label{eqn:treatment_effect_te}
    \textsf{TE}_{x,x'}(Y) = \mathbb{E}[Y|do(X=x')] - \mathbb{E}[Y|do(X=x)] = \mathbb{E}[Y_{x',M_{x'}} - Y_{x,M_x}]
    %= \textsf{NDE}_{x,x'}(Y) - \textsf{NIE}_{x',x}(Y)
\end{equation}
\end{definition}
where $M$ is the set of parent variables of $Y$ except $X$, i.e., $M = \textrm{PA}_Y \setminus \{X\} = \{Z, W\}$, and $M_{x'}$ is the value $M$ would have attained under the condition $X=x'$.
The direct influence of the treatment variable $X$ on $Y$, or the ``direct effect'', is mingled with the mediating effect from $Z$. 
In order to uncover the direct influence of the treatment, we hold $M$ constant and change $X$ from $x$ to $x'$, as shown in Figure~\ref{fig:lmt_treatment}(b). 
The resulting change in outcome variable is considered as the ``Natural Direct Effect'' ($\textsf{NDE}$), as in Equation~\ref{eqn:treatment_effect_nde}.
The indirect effect refers to the anticipated change in the outcome variable $Y$ under the circumstance where the treatment variable $X$ remains unchanged, but $M$ is altered by the treatment's influence to become $M_{x'}$. 
To quantify the ``total effect'' of $X$ on $Y$, we can directly use the intervention $do(x)$, or use the relation between different treatment effects. From Equations~\ref{eqn:treatment_effect_nde}-\ref{eqn:treatment_effect_te}, it can be seen that
\begin{align}
   \textsf{TE}_{x,x'}(Y) = \textsf{NDE}_{x,x'}(Y) - \textsf{NIE}_{x',x}(Y) 
\end{align}
which means that the total effect of a transition is equal to the difference between the natural direct effect of the transition and the natural indirect effect of the inverse transition \cite{pearl09causality}.

Recent deep learning research has incorporated treatment effect method to either account for the impact of certain variables, or remove the influence of certain variables \cite{Hu2021DistillingCE, wang2021weaklysupervised, Singla2021UsingCA}. $Z$ often represents secondary cues used to predict $Y$, which are influenced by the primary cues. 
For instance, \cite{rao2021counterfactual} considered $Z$ as the refined features resulting from the use of soft attention \cite{xu2015show}, and used a network to learn basic feature maps and attention separately to create the structural causal model (SCM) displayed in Figure~\ref{fig:lmt_treatment}(a). 
Several approaches have been employed to address $Z$, considering it a bias resulting from a specific type of occurrence between particular objects or features \cite{tang2020unbiased, niu2021counterfactual}, or events observed in sequential data \cite{10.1145/3474085.3475472, chen2021human, wang2021weaklysupervised, ijcai2021-182}.
For example, some studies \cite{reddy2021causally, tang2020unbiased, 2019, ijcai2021-182} consider the variable $Z$ as the variable carrying bias in the output and used the natural direct effect. 
\cite{chen2021human, Hu2021DistillingCE, li2021driver, wang2021weaklysupervised} quantified the effect of a variable on the output by subtracting the nullified effect (the effect measured by fixing $Z$ to be $0$)
from the original effect. 
On the other hand, some works \cite{niu2021counterfactual, rao2021counterfactual, unknown, 10.1145/3474085.3475472} involve problem settings where $Z$ should be considered the main cue for predicting the output $Y$, and $X$ is the spurious variable. 
In such cases, the natural indirect effect was utilized to assign greater weightage to $Z$.
\cite{Singla2021UsingCA} utilized both the direct and indirect effect to determine the significance of different features in the output.

Many of the works in this section have focused on using different training objectives aligned with principles of causal inference, to prevent the model from learning biases in the training data. However, some other works \cite{chattopadhyay2019neural, geiger2021causal, geiger2022inducing} have taken a more model-driven approach by incorporating causal prior into the model. 
For instance, \cite{chattopadhyay2019neural} view a neural network as an SCM, and quantified the impact of features on the output via Average Causal Effect. \cite{geiger2022inducing} aligned representations in a neural network with variables in an SCM, and trained the network to mimic the counterfactual behavior of the causal model on the same base input.
Similarly, \cite{geiger2021causal} proposed a method to align the different sets of neurons in the neural network as nodes of an SCM. 
As such, it mimics intervention on the SCM by assigning values to the neurons artificially, supporting reasoning at the neuron level of the network.

\subsection{Summary}

In this section, we introduced Pearl's Ladder of Causation, various techniques of causal inference, and discussed their connections to trustworthy machine learning techniques. It can be seen from our discussion that the techniques we identified in Section~\ref{sec:robustness}, despite been based on various statistical theories or intuitions, can be grounded in the causal inference framework.  
When a new regularization or augmentation technique is proposed, it is important to be aligned with the causal assumptions of the corresponding task.
In addition, the direct application of causal inference to deep learning methods is challenged by the problems of unobserved variables and the entanglement of causal and non-causal features. For the communities who are interested in grounding trustworthy ML with causality, our results suggest future works can aim to alleviate these problems so that deep learning methods can more strictly follow the procedures of causal inference.

\section{Trustworthy Machine Learning in the Context of Pretrained Models} 
\label{sec:pretrained}
In previous sections, we discussed the technical advancements in trustworthy machine learning, synthesized these developments into a central theme, and connected it to the concept of causality. 
Now, we transition our focus to large, pretrained models. These models are usually trained on extensive web-scale data and thus will potentially introduce unique challenges related to trustworthiness.

Certain literature posits that the immense variety and volume of training data might mitigate some trustworthiness issues. 
However, an expanding body of research suggests that these trustworthiness challenges persist in pretrained models. 
Further, the complexity of these models and the opaque nature of their training processes may make these issues harder to address.

The proposition of retraining large models using techniques previously discussed could potentially alleviate some challenges. Nevertheless, this approach is often impractical due to the substantial parameter and data sizes. 
Fortunately, there are techniques such as fine-tuning, parameter-efficient tuning, and prompting. While these methods were initially designed to improve model performance and efficiency, they also provide opportunities to enhance the models' trustworthiness.

In the remaining part of this section, we will recap the advancements in large models and present different opinions on their trustworthiness. Then, we will review commonly utilized techniques for large models, aiming to establish a connection between these techniques to the ERM training strategies that previously discussed methods seek to improve. 
We believe the discussion will continue to connect the techniques of large pretrained models into the converged theme.
Finally, we will survey techniques explicitly devised to make large models more trustworthy, 
which will demonstrate that our central theme, while being established within the context of standalone model training, remains applicable in the context of large, pretrained models.

\subsection{Large Pretrained Models and Its Trustworthiness Challenges}

In recent years, we have witnessed an exceptional surge in the field of machine learning research, particularly with regards to pretrained models. As of the date this survey was drafted, this area of study can arguably be considered the most prominent in AI research and applications. Models like ChatGPT, for instance, have transcended the boundaries of machine learning communities, becoming the nucleus of numerous discussions at various events \cite{dwivedi2023opinion, kasneci2023chatgpt, sallam2023chatgpt}.

Concurrent with its widespread practical influence, the core concept behind large pretrained models is elegantly simple and widely comprehensible: as we increase the model parameter size together with the volume of data, many machine-learning challenges could eventually be addressed. 
Even in instances where specific challenges persist, fine-tuning the model with distinct datasets typically proves more effective than training an isolated model from scratch.

While the utilization of pretrained models has seen a dramatic upswing in popularity in recent times, it is probably not a recent invention. We hypothesize that earlier works likely sought to employ parameters trained on one dataset to augment performance in other areas, as seen in transfer learning \cite{pan2010survey, daume2007frustratingly, blitzer2007biographies}. However, such ideas gained traction in the context of deep learning with the advent of AlexNet \cite{krizhevsky2017imagenet} in computer vision and Word2Vec \cite{mikolov2013efficient} in NLP.

In computer vision, following the triumphant runs of AlexNet and other CNN models in the ImageNet competition, a segment of the community began enhancing their performance in other applications like object detection \cite{girshick2014rich, girshick2015fast, ren2015faster} and segmentation \cite{girshick2014rich, ronneberger2015u}. This methodology became so prevalent that the primary ImageNet-trained model was often referred to as the ``backbone'' in their method. In addition, the ``backbone'' can conveniently adapt to the latest architectures \cite{carion2020endtoend, gao2021fast, zhu2021deformable} along with ImageNet research progression.

In NLP, pretrained models gained popularity with the introduction of word vectors via Word2Vec \cite{mikolov2013efficient} or GloVe \cite{pennington2014glove}. These context-encoding word representations swiftly supplanted the discrete word input in other NLP models. More recently, the BERT \cite{devlin2018bert} pretraining regime, capable of excelling in various downstream NLP tasks with minimal fine-tuning, became popular. Subsequent to BERT's success, numerous language models were introduced \cite{yang2019xlnet, sun2020ernie2, liu2019roberta, lan2020albert, raffel2020exploring, clark2020electra, he2021deberta, brown2020language, beltagy2020longformer, ouyang2022training}, with models like ChatGPT significantly impacting communities beyond AI research.

It is crucial to note that the GPT family of models, including ChatGPT, are not the sole large models making substantial impacts beyond the AI community. For instance, AlphaFold \cite{jumper2021highly} has demonstrated considerable practical value in predicting protein's 3D structures; CLIP \cite{radford2021learning}, the core technique underpinning DALL-E, has also served as pretrained models in numerous vision-language learning tasks; and more recently, vision models like SAM \cite{kirillov2023segment} are also emerging as dominant players in various vision tasks. Their progress in their respective fields has been so significant that a portion of the research community has proposed naming them ``foundation models'' \citep{bommasani2022opportunities}. 

In this survey, our primary focus will remain on vision and NLP applications. Consequently, our survey will encompass techniques and discussions on the trustworthiness of NLP models such as the GPT family, BERT \cite{devlin2018bert}, PaLM \cite{chowdhery2022palm, anil2023palm}, and LLaMA \cite{touvron2023llama, touvron2023llama2}, vision models like SAM \cite{kirillov2023segment}, DINOv2 \cite{oquab2023dinov2}, SEEM \cite{zou2023segment}, and vision-language models like CLIP \cite{radford2021learning}, DALL·E \cite{ramesh2021zero, ramesh2022hierarchical}, Flamingo \cite{alayrac2022flamingo}, BLIP \cite{li2022blip}, and GPT-4 \cite{openai2023gpt4}.

It is worth mentioning that the core strength leading to these models' superior performance varies. For instance, while models like CLIP, SAM, and early-stage language models primarily benefit from data volume, Reinforcement Learning with Human Feedback (RLHF) \cite{christiano2017deep, stiennon2020learning} has significantly enhanced the practical performance of recent language models.

RLHF, or more broadly, machine learning with human feedback, is a process that enlists humans to guide and evaluate the outputs of AI systems. Initial guidance comes from demonstrations, where humans provide instances of correct behavior or desired outcomes (such as annotators' understanding of an image class). After the AI system generates its outputs, humans provide feedback by ranking output quality or providing direct corrections. This feedback is then used to refine the AI model, in a cycle that continues until satisfactory performance is achieved.

It is interesting to draw conceptual connections between machine learning with human feedback and our general definition of trustworthy machine learning. Specifically, the need for human feedback arises primarily due to discrepancies between the statistical loss that is typically trained and what humans find useful \cite{radford2019language, brown2020language, fedus2022switch, rae2021scaling, thoppilan2022lamda}.
Hence, machine learning with human feedback could potentially bring models closer to the trustworthy attributes we desire. This might be one reason why a part of the community has started to view these large models as possessing multiple trustworthy merits.

\subsubsection{Positive Opinions on Trustworthy Properties of Large Models} Despite the feast celebrated by the public and media about the strengths of these pretrained models, 
we do not find an abundance of rigorous academic publications 
supporting that these models are equipped with desired trustworthy merits. 
In the few examples we found, 
the discussions are also fairly objective stating that 
the pretrained models tend to have 
much higher performances on various benchmarks \cite{brown2020language,devlin2018bert,radford2019language}, 
even on out-of-distribution data (defined as datasets released after the date the model is trained) \cite{wang2023robustness}, 
or causal inference benchmarks \cite{kiciman2023causal},
and stating that these performances are not perfect 
and need further improvement. 

\subsubsection{Negative Opinions on Trustworthy Properties of Large Models} \label{sec:llm:negative}
On the other hand, 
as the community continues to explore the technical advancements of these large models, there is a growing recognition of their potential limitations, fueling concerns and driving further investigation.

\paragraph{Lack of Domain-specific Robustness}
Given the sheer volume of the data used to train the large pretrained models, 
it will be hard to test the models' robustness performances 
against traditionally defined out-of-domain data:
since the models might have already seen the domains during training. 

However, in certain application fields, where the data are of a different nature to the daily language and images, 
these large models are still suffering a fairly large performance drop, 
although still better than the previous standalone models.  

For example, the community has shown limited performances of SAM over medical images \cite{shi2023generalist, mazurowski2023segment}, 
of GPT over medical texts \cite{sallam2023chatgpt, dash2023evaluation},
of Stable Diffusion \cite{rombach2022high} over medical images and text \cite{chambon2022adapting}. 
Other domains, such as legal texts \cite{shaghaghian2020customizing} 
and financial texts \cite{wu2023bloomberggpt} are also facing challenges to directly applying the generic large models. 
The challenges of applying pre-trained models to specific domains have become significant, and there are surveys dedicated to the discussions \cite{ling2023beyond}.

With these specific challenges of the model's understanding over the knowledge in specific domains, it seems fine-tuning the pretrained models on these specific domains is a natural solution. 
However, it seems a direct fine-tuning of the pretrained models
on certain datasets only appear to solve the problems 
but will reveal more weakness when more sophisticated testing are performed \cite{wortsman2022robust,gudibande2023false}, 
which is potentially due to the 
catastrophic forgetting issue of large models \cite{chen2020recall},
or in general, machine learning models \cite{de2021continual}.

\paragraph{Lack of Fairness (equal representation)}
It is widely observed that machine learning models frequently generate outputs that perpetuate stereotypes and discrimination against marginalized groups.
These models not only reflect, but sometimes exacerbate the biases present in the dataset \cite{nozza-etal-2021-honest}. When it comes to these large pre-trained models, these issues are seemingly getting worse, as the construction of larger models typically requires a larger volume of data, which can result in reduced data quality. 

The issues exist in both discriminative models and generative models. 
A classifier may erroneously exploit demographic information for prediction \cite{kiritchenko2018examining, lin-etal-2022-gendered, baldini2021your}, 
and a text generation model could directly express prejudiced views against 
certain demographic groups \cite{nadeem2020stereoset,nozza-etal-2021-honest, schick2021selfdiagnosis}. For instance, bias pertaining to race, gender, and other demographic attributes has been discovered in models such as GPT-2 \cite{sheng-etal-2019-woman} and BERT \cite{tan2019assessing}. \cite{abid2021persistent} identified persistent bias towards Muslim people in GPT-3. \cite{zhao2019gender} discovered gender bias in earlier models like ELMO \cite{peters2018deep}, and \cite{lucy-bamman-2021-gender} identified gender stereotypes in more recent models like GPT-3. Although the RLHF technique improves the model's sentiment towards all social groups, it does not appear to reduce the disparities among them \cite{ouyang2022training}. Furthermore, biases have been assessed in pre-trained models across different languages \cite{chalkidis2022fairlex, wang2021assessing}, and systematic evaluations have identified bias pertaining to nine demographic attributes in masked language models \cite{nangia2020crows, kaneko2022unmasking}. These studies reveal the significant and widespread nature of bias in pre-trained language models \cite{stochastic_parrots}.

The papers discussing such issues are more than what our survey can cover. For a more comprehensive overview, one can refer to other surveys dedicated to discussions of the critical need for fairness in pre-trained language models \cite{blodgett-etal-2020-language,field-etal-2021-survey,kumar2022language}. 
Some studies \cite{sharma-etal-2022-sensitive, gupta2022mitigating, lauscher2021sustainable} focused on improving fairness in PLMs. 
For instance, \cite{sharma-etal-2022-sensitive} eliminated gender bias using instruction text. 
\cite{gupta2022mitigating} alleviated the transfer of bias from PLMs to student models during the distillation process. 
\cite{lauscher2021sustainable} employed the approach of PEFT to eliminate biases from the model.
Meanwhile, several studies such as \cite{mattern2022understanding} have focused on enhancing bias quantification metrics. 
It has been discovered that current metrics inherently incorporate bias \cite{sun-etal-2022-bertscore} and are occasionally unreliable \cite{delobelle2022measuring}.

\paragraph{Hallucination}
Another challenge for large pre-trained models is hallucination. 
In general, it refers to the 
state that the machine learning models, 
in particular, natural language generation models, 
can generate texts that are plausibly correct, 
but wrong \cite{ji2023survey}. 

While \cite{ling2023beyond} associated hallucination with the inability of the model to understand domain-specific knowledge, 
the broader scope of the community tend to consider it more generic. 
For example, 
it can happen in many generic text-generation task 
such as machine translation \cite{raunak2021curious}
and
image captioning \cite{rohrbach2018object,biten2022let}. 

More formally, hallucination is defined as ``the generated content that is
nonsensical or unfaithful to the provided source content'' \cite{ji2023survey,maynez-etal-2020-faithfulness}. 
There are also different definitions available \cite{thomson-reiter-2020-gold,dong-etal-2020-multi,raunak2021curious,lee2018hallucinations}. 
In summary, we believe the key differences of these definitions lie in the agreement on where the factuality should be aligned to, 
which again aligns with our general theme of trustworthy machine learning in this paper: the complete definition must be specified by the stakeholders.

\subsection{Techniques with Large Models} \label{sec:llm}
This section introduces several techniques commonly used with large models. In Sections \ref{sec:llm:finetune}-\ref{sec:llm:peft}, we introduce fine-tuning, prompting, and parameter-efficient fine-tuning, respectively. These techniques were initially designed to improve the performance or efficiency of large models. Section \ref{sec:llm:rlhf} introduces RLHF \cite{christiano2017deep}, a technique that aims to align the behaviors of language models with human values.

\subsubsection{Fine-tuning} \label{sec:llm:finetune}
Fine-tuning refers to the process of modifying the weights of a pretrained model for a downstream task. 
Recent work has shown that fine-tuning changes the underlying representation of data by increasing the distance between different labels, and thus, leads to better generalization \cite{zhou2021closer}. 
However, a significant challenge in fine-tuning is the lack of robustness in the adjusted weights compared to the pre-trained models \cite{merchant-etal-2020-happens, mosbach-etal-2020-interplay, hao-etal-2020-investigating}. 
To increase the robustness of fine-tuned models, researchers have suggested different approaches \cite{jiang-etal-2022-rose, jian2022embedding, liu2023twins, xiao2023masked, tian2023trainable, zhang-etal-2022-rochbert, zhang-etal-2022-fine-mixing, kumar2022fine}, 
which can be broadly categorized as architecture-driven and data-driven techniques.

Architecture-driven approaches \cite{jiang-etal-2022-rose, liu2023twins, tian2023trainable} deal with fine-tuning the model with a different approach compared to the conventional one. 
For example \cite{jiang-etal-2022-rose, tian2023trainable} discard the idea of fine-tuning all the parameters of the pre-trained model. 
ROSE \cite{jiang-etal-2022-rose} select parameters based on dropout using adversarial perturbation and tuning parameters that are less aggressively updated to avoid overfitting on spurious patterns.
Instead of selecting parameters for updates or layers, \cite{tian2023trainable} automatically imposes the constraint between the model weights and projection radii, i.e. the distance constraints between the layers of the fine-tuned and pre-trained model through weight projections to retain robust features in the fine-tuned model. 
Furthermore, \cite{liu2023twins} retains the pre-trained features by fine-tuning the pre-trained models to the downstream task by training dual networks (pre-trained and fine-tuned) with shared parameters, excluding the batch normalization layers. 
Without changing the architecture,
\cite{kumar2022fine} suggests employing linear probing followed by fine-tuning gives good results on Out-of-Distribution fine-tuning.
Similarly, \cite{chen-etal-2023-fine} proposes GNOME to penalize the representation related to semantic and non-semantic shifts move, in case it moves away from their respective mean.
It argues that fine-tuned pre-trained language models (PLMs) work on a trade-off between the semantic and non-semantic shifts. 
Furthermore, they find out that fine-tuning PLMs on redistribution data benefits detecting semantic shifts (OOD consists of unseen classes in the in-distribution task) but severely deteriorates detecting non-semantic shifts (OOD consists of a change in background information, but has same classes). 
Similarly, to infuse robustness in the fine-tuned model, the work \cite{he2023preserving} focuses to preserve the pre-trained features to help calibrate the fine-tuned PLMs by maintaining consistency between their trained features.

On the other hand, the data-driven approaches \cite{jian2022embedding, zhang-etal-2022-rochbert, xiao2023masked}
achieves robustness for fine-tuned models by data-oriented techniques, like \cite{jian2022embedding} insert additional hallucinated data with pseudo labels to provide the fine-tuned model with a more diverse dataset to avoid overfitting on the few-shot learning task. 
On a similar token, \cite{zhang-etal-2022-rochbert} generate adversarial text via curriculum learning method \cite{bengio2009curriculum} to also allow intermediate text as the sample becomes more uncertain. 
\cite{broscheit2022distributionally} discusses how covariate shift affects the performance of BERT poorly and how infusing covariate drift augmentation transforms to make the fine-tuned model distributionally robust, while also improving its performance.
Furthermore, it focuses to learn glymph and phonetic features via adversarial graphs as these features are usually perturbed to generate adversarial text. 
\cite{xiao2023masked} utilize the third level of causation~\ref{sec:counterfactuals} to generate counterfactual images by masking semantic or non-semantic features and filling them with randomly sampled images. 
Meanwhile focusing to maintain consistency between the prediction of the pre-trained and fine-tuned model to transfer robustness from the pre-trained model.
Based on these works, infusing robustness in a hybrid mode, i.e. using both the data-oriented and model-oriented way seems like a desirable solution.
One such work, \cite{zhang-etal-2022-fine-mixing} executes it for backdoor attacks by cleaning the potential backdoor embeddings, which is done by aligning the embedding of top words with that backdoor model of the pre-trained model. 
It also proposes a strategy to mix the backdoor model weights and pre-trained weights to train them on a small set of clean data. 

%%%%%%%%%%%%%%%%%%%%%%%%
%%%%%%%%%%%% forgetting
%%%%%%%%%%%%%%%%%%%%%%%%
Another branch of works \cite{yang-ma-2022-improving, lee2023surgical} focuses on solving the problem of catastrophic forgetting knowledge gained in the pre-training phase. 
From a broad perspective, the researchers have tried to solve this by selectively fine-tuning different parameters/layers in the model, as fine-tuning additional parameters on small datasets can cause forgetting\cite{lee2023surgical}.  
For example, \cite{lee2023surgical} selectively fine-tunes the layers of the model based on the distribution shift, also unlike conventional methods, which selectively fine-tunes the last or last few layers. 
It also gives liberty to only fine-tune the model using initial layers.
Similarly, to prevent unstable fine-tuning \cite{yang-ma-2022-improving} focuses on balancing the unbalanced gradients across different components thought the ``component-wise gradient norm clipping'' method to adjust the speed for different type of parameters. 
To make the fine-tuned model more trustworthy \cite{shi-etal-2022-just} aims to protect privacy leakage for the sensitive tokens defined in a policy function by fine-tuning large-transformer-based models. 
It achieves it by fine-tuning the model twice, i.e. firstly by masking the sensitive tokens to fine-tune the model on in-distribution data, giving a good initialization to the second phase of training on original in-domain data using a private optimizer \cite{Abadi_2016}. 

%%%%%%%%%%%%%%%%%%%%%%%%%%
%%%%%%%%%%%%% Others
%%%%%%%%%%%%%%%%%%%%%% 
Apart from these works, there exist other works \cite{ko2023meltr, yu2022actune, dessì2023crossdomain, zhang2022fine, yang2022gram, shi-etal-2022-just, bursztyn-etal-2022-learning, goyal2023finetune, maekawa-etal-2022-low} which focus on refining the fine-tuning technique to make it data-efficient and more effective.
They \cite{ko2023meltr} employ it by proposing new loss functions including additional loss functions by non-linearly combining losses corresponding to different auxiliary tasks.
Furthermore, \cite{zhang2022fine} proposes two new regularizers to accomplish the isotropic property of feature space for intent classification. 
\cite{goyal2023finetune} argues that one should not abandon the fashion of pre-training and fine-tuning ought to be done in the same manner. 
It accomplishes it by employing the contrastive loss between the class-descriptive prompt embedding and image embedding.
Similarly, to make the process more efficient \cite{bursztyn-etal-2022-learning} harness the idea of curriculum learning \cite{bengio2009curriculum}, i.e. by learning easier concepts and progressing to harder ones. It fine-tunes smaller language models on these sub-tasks to learn a target task.
\cite{dessì2023crossdomain} harnesses reinforcement learning for zero-shot fine-tuning an out-of-box image neural captioner by updating the language part. 
It uses the reward from the retriever, which selects the original image from which the caption was generated. 

To make the process of fine-tuning more data-efficient \cite{devlin2018bert} innovates upon active annotation.
ACTUNE fine-tunes BERT \cite{devlin2018bert} based on pseudo-labels obtained for low-certainty examples while actively annotating the high-uncertainty samples.
Furthermore, \cite{maekawa-etal-2022-low} proposes not to only have minimum labeling cost while randomly selecting the data but also focus on low acquisition latency. 
Additionally, it focuses on both diversity and uncertainty by selecting the cluster centers of acquired data points and selecting the ones having low-confidence prediction or high entropy to gain uncertain examples.

\subsubsection{Prompting} \label{sec:llm:prompting}

Prompting, or prompt learning, involves the process of using a template to alter the input, which can be done manually or automatically. 
The template is tailored to suit the specific task requirements. 
By making the form of the downstream task more similar to the pre-training task, we hope to better leverage the capabilities of the pre-trained model so that the downstream task can be performed with reduced or no need for additional training.

In prompt learning, we fill the input into a fixed location of the template, and let the model generate output at the reserved location, usually at the middle of the text for bi-directional language models, or at the end of the text for auto-regressive models.
However, sometimes the model output cannot be used directly as the answer. 
In this case, another step is often taken to map the output to the answer space, often known as verbalization.
Prompts can be obtained by manual engineering \cite{petroni-etal-2019-language, brown2020language, schick2021fewshot} or automated search \cite{shin-etal-2021-constrained, jiang-etal-2020-know}. 
In manual prompts, we manually define an appropriate prompt to guide the PLM towards completing a task. 
This process is more interpretable, but it relies more on human expertise and may not yield optimal results \cite{jiang-etal-2020-know} compared to automated search.
In automated engineering, various algorithms are employed to discover the most effective prompt for a given task. 
Prompts can also be categorized as ``discrete prompts'' \cite{jiang-etal-2020-know, haviv-etal-2021-bertese, shin2020autoprompt, wallace-etal-2019-universal, gao-etal-2021-making, bendavid2022pada, davison-etal-2019-commonsense} and ``continuous prompts''.
Discrete prompts are composed of words and are represented as strings, whereas continuous prompts pertain to prompts that exist within the embedding space of the model. 
Continuous prompts are considered as independent parameters and can be optimized from scratch or 
initialized using discrete prompts \cite{zhong2021factual, shin2020autoprompt, qin-eisner-2021-learning, hambardzumyan-etal-2021-warp}.
They can be incorporated into the sentence at the beginning \cite{li2021prefixtuning, lester2021power, tsimpoukelli2021multimodal} or at more flexible locations \cite{liu2021gpt, han2021ptr} as embedding vectors.

Instead of using single prompts to arrive at the desired answer, recent works \cite{schick2021exploiting, yuan2021bartscore, jiang-etal-2020-know, qin-eisner-2021-learning, schick2021exploiting, schick2021selfdiagnosis, lester2021power, hambardzumyan-etal-2021-warp, allenzhu2023understanding, schick2021exploiting, schick2021selfdiagnosis, gao-etal-2021-making, schick2021fewshot} used multiple prompts of the input.
These multiple answers were used by some works \cite{ schick2021exploiting, yuan2021bartscore, jiang-etal-2020-know, qin-eisner-2021-learning, schick2021exploiting, schick2021selfdiagnosis} to predict the output. 
Meanwhile other works 
\cite{allenzhu2023understanding, schick2021exploiting, schick2021selfdiagnosis, gao-etal-2021-making}
leveraged different small models as intermediaries corresponding to each of the multiple prompts to get trained and predict different outputs, which can then be used to predict the final output.

There has been a growing interest recently in the research of a special kind of prompt learning, known as in-context learning (ICL) \cite{brown2020language}. In ICL, a few input-answer pairs are incorporated into the prompt as demonstration data, helping the pre-trained model adapt to the new task without any training. ICL has shown to effectively improve the performance of PLMs in a variety of relatively complex tasks \cite{brown2020language, wei2022chain}. Several works have tried to understand the working mechanisms of ICL.
\cite{akyürek2023learning} found that transformer possesses the ability to implement learning algorithms for linear models implicitly in their hidden states, and update these models according to the demonstration samples. 
Other works \cite{dai2023gpt,li2023closeness, akyurek2023what,vonoswald2023transformers} discussed the theoretical connections between ICL and fine-tuning, as well as their similarities in empirical behaviors. 

Mathematically, the method of in-context learning can be defined using Equation~\ref{eqn:incontext}
\begin{equation} \label{eqn:incontext}
    \hat{P}(y|x) = f(y|x, \{(x_i,y_i)\}_{i = 0}^{i = n}, \mathcal{I};\theta)
\end{equation}

where $\mathcal{I}$ is a natural language instruction of the task, and $(x_i,y_i)$ is the $i$th pair of demonstration input and answer. The ICL ability can be improved using model warm-up, which acts as an intermediary fine-tuning stage between the pre-training and ICL inference stages.
Based on the finding that ICL performance benefits from the diversity of pre-training corpus, \cite{chen-etal-2022-improving, chen2023relation}
fine-tuned the parameters of the language models such that they outperformed language models with larger sizes. 
\cite{wei2022finetuned}
found that adding instructions while pre-training acts as a performance booster. 
On the other hand, many other works have explored the formatting or selection of demonstration examples. 
\cite{liu-etal-2022-makes, rubin-etal-2022-learning, sorensen-etal-2022-information, liu-etal-2022-makes} 
focused on selecting examples such that there is high information sharing
\cite{liu-etal-2022-makes, rubin-etal-2022-learning}
between the query input and the example based on different metrics such as distance or mutual information \cite{sorensen-etal-2022-information}. 
Similarly, example order plays a crucial role in ICL performance \cite{liu-etal-2022-makes},
where the rightmost example should be highly similar to the input sample.

\subsubsection{Parameter-efficient Tuning} \label{sec:llm:peft}
Fine-tuning involves modifying the weights of a pretrained model for a specific task. 
However, fine-tuning all parameters in a large language model is highly resource-intensive due to the exponential rise in the parameters from LMs like BERT \cite{devlin2018bert} with $350$ million parameters (released in $2018$) to GPT$-3$ \cite{brown2020language} containing $175$ billion parameters.
It is not feasible or cost-effective to train the entire language model from scratch for each task. 
As a result, a sub-field called ``Parameter-Efficient Fine-Tuning'' (PEFT) has emerged,
which aims to minimize
the computational, memory, and storage demands for fine-tuning.
The community has proposed various ways \cite{hu2021lora,edalati2022krona, vucetic2022efficient, zaken2021bitfit, guo2020parameter, sung2021training, donahue2014decaf, houlsby2019parameter, wang2022adamix, liu2022few, sung2022lst, hu2023llm, pfeiffer2020mad, pfeiffer2020adapterfusion, ruckle2020adapterdrop}
to accomplish PEFT, with the common ground of involving fewer parameters for updates during fine-tuning. 
The main challenge lies in identifying the optimal and minimal weights for fine-tuning the model.
Certain methods \cite{hu2021lora,edalati2022krona} 
attempt to address this by identifying the low-rank representation of current weights of LM model, and hence infusing them with the pre-trained weights as trainable parameters, while keeping the original weights frozen. 
Research \cite{aghajanyan2020intrinsic} supports this approach by considering pre-trained models as low-dimensional intrinsic models that can learn effectively even when projected onto a random subspace.

In addition to obtaining the low-rank representation for fine-tuning, certain studies \cite{vucetic2022efficient, zaken2021bitfit, guo2020parameter, sung2021training, donahue2014decaf} identified the most suitable parameters \cite{vucetic2022efficient, zaken2021bitfit, guo2020parameter, sung2021training}, and selectively fine-tuned layers \cite{donahue2014decaf} of the model. 
For instance, \cite{zaken2021bitfit} manually selected specific parameters for update, such as the ``bias'' parameter.
On the other hand, \cite{vucetic2022efficient, guo2020parameter} adopted a model-agnostic approach by masking parameters during training. 
In \cite{vucetic2022efficient}, parameters were categorized as trainable or non-trainable, while \cite{guo2020parameter} masked parameters to determine their relevance during training. 
To reduce the computation cost of finding the important parameters, \cite{sung2021training} proposed to pre-compute the crucial parameters to be fine-tuned during training using the Fisher Information.

Furthermore, several studies \cite{houlsby2019parameter, wang2022adamix, liu2022few, sung2022lst, hu2023llm, pfeiffer2020mad, pfeiffer2020adapterfusion, ruckle2020adapterdrop} have explored the incorporation of additional parameters into neural networks. 
These extra parameters can only be re-parameterized during the fine-tuning process while the existing weights remain unchanged. 
The concept of adapters \cite{houlsby2019parameter}  represents an early example in this category, where adaptive training modules are added within the transformer layer. 
Subsequent research \cite{wang2022adamix, hu2023llm, pfeiffer2020mad, pfeiffer2020adapterfusion, ruckle2020adapterdrop} has expanded on this concept to further improve the efficiency of fine-tuning. 
For multi-task learning, AdapterFusion \cite{pfeiffer2020adapterfusion} trained an adapter module for each task and utilized a fusion module combining these adapters to tackle a specific task. AdapterDrop \cite{ruckle2020adapterdrop} improved the memory efficiency of \cite{pfeiffer2020adapterfusion} by pruning adaptors in lower transformer layers. \cite{hu2023llm} integrated several methods such as Series adaptor \cite{houlsby2019parameter}, Parallel adaptor \cite{pfeiffer2020mad}, and LoRa \cite{hu2021lora} into large models for various tasks. AdaMix \cite{wang2022adamix} incorporated multiple adapter layers, randomly selected for each forward pass, to enhance performance. Mad-X \cite{pfeiffer2020mad} further introduced an invertible adapter architecture, which used the same set of parameters for adapting both input and output representations for cross-lingual transfer.

In addition to adapter methods, other techniques \cite{liu2022few, sung2022lst} also focus on introducing additional parameters for task-specific fine-tuning.
\cite{liu2022few} added parameters that scale the activations in the transformer layers based on the inferred task. 
LST \cite{sung2022lst} trained a small side network detached from the pre-trained model, and improved memory efficiency by  avoiding backpropagation through the backbone network.

\subsubsection{Fine-tuning with human feedback signals} \label{sec:llm:rlhf}
Large language models are often pretrained to predict the next token given the context in a corpus, which is different from the objective to ``follow the user’s instructions helpfully and safely'' \citep{radford2019language, brown2020language, fedus2022switch}. This misalignment in objectives may be the primary reason behind the widely observed phenomenon that Pretrained Language Models (PLM) sometimes produce uninformative, toxic, or misleading response \cite{askell2021general, ouyang2022training} despite the ever-increasing size of model parameters and pretraining corpus. To alleviate this problem, Reinforcement Learning from Human Feedback (RLHF) \cite{christiano2017deep, stiennon2020learning} has recently been used to steer a pretrained language model towards intended behaviors by incorporating human preference signals into the training pipeline. RLHF was originally proposed to train agents in relatively simple simulated environments \cite{christiano2017deep, ibarz2018reward, bahdanau2018learning}, and was later used in specific language tasks such as text summarization \cite{ziegler2020finetuning, stiennon2020learning, bohm2019better, wu2021recursively}, machine translation \cite{kreutzer2018neural, bahdanau2017an}, dialogue system \cite{jaques2019way, yi2019towards, hancock-etal-2019-learning}, semantic parsing \cite{lawrence2018improving}, and  story telling \cite{zhou2020learning}. As large language models show strong cross-task generalization abilities with in-context learning \cite{brown2020language, dong2023survey} and instruction tuning \cite{wei2022finetuned}, the community has recently focused on using RLHF to align a language model with human values to perform a wide range of tasks \cite{ouyang2022training}, as a general-purpose chatbot or language assistant \cite{bai2022training, ganguli2022red}. 

In a typical RLHF process \cite{stiennon2020learning, ouyang2022training, bai2022training}, a dataset is carefully curated by collecting user-generated prompts paired with human-written answers as demonstration. This dataset is used to fine-tune the PLM in an initial warm-up stage. Then, for each prompt, different answers are collected from the output of different versions of the PLM, the human demonstration, and various baselines. A group of annotators were asked to rank the quality of $K$ answers ($K \geq 2$) to the same prompt, according to some predefined human values, and all the $\binom{K}{2}$ pairs from each prompt are used to construct a comparison dataset. A reward model is trained to predict which answer in a pair is preferred by human by assigning a scalar reward to each answer according to the below formula
\begin{align}
\operatorname{loss}(\theta) = - \frac{1}{\binom{K}{2}}
\mathop{\mathbb{E}}_{(x, y_w, y_l) \sim D} [\log(\sigma(r_\theta(x, y_w) - r_\theta(x, y_l)))]
\end{align}
where $r_\theta(x, y)$ is the output of a reward model for prompt $x$ and answer $y$ with parameter $\theta$, $y_w$ is the preferred answer in the pair of $y_w$ and $y_l$, $\sigma(\cdot)$ is the Sigmoid function, and $D$ is the distribution of the comparison dataset. The output of the reward model is then normalized to have a mean of $0$ across the dataset \cite{stiennon2020learning, ziegler2020finetuning}.

With the learned reward model, we can sample answers with the PLM and optimize the expected reward using the PPO algorithm \cite{schulman2017proximal}. The auto-regressive generation of the PLM can be seen as a Markov Decision Process, where the state is the concatenated string of the prompt tokens and previously generated answer tokens, and the action is to predict the next token and append it to the current string. At the end of the generation, a reward is produced by the reward model and assigned to each token, with a KL loss to prevent the current policy from deviating too much from the original policy.
\begin{align}
R(x,y) = r_\theta(x, y) - \beta \frac{\pi(y|x)}{\pi_0(y|x)}
\end{align}
where $\pi$ are $\pi_0$ are the current and original policy models respectively, and $\beta$ is a coefficient. The collection of comparison data and PPO retraining of model can be done iteratively. \cite{bai2022training} update the reward models and RL policies with fresh human data on a weekly basis to continuously improve the quality of the dataset and models. Recent works have also tried to improve the RL algorithm. \cite{ramamurthy2023is} employ a policy mask by selecting the top $p$ tokens, which reduced the action space and stabilized training. They also developed an open-source library and a benchmark to optimize and evaluate language models with RL. \cite{snell2023offline} extend the Implicit Q-learning method \cite{kostrikov2022offline} to language generation tasks and show improved stability and performance.

Reinforcement learning is not the only way to incorporate human feedback signals in the training process. \cite{dong2023raft} learn a reward model from human preference and fine-tune the PLM on the highly ranked samples. \cite{yuan2023rrhf, zhao2023slichf} align the model-produced probabilities of answers with their human rankings. \cite{xu2023imagereward} fine-tune a diffusion model towards better scores from a reward model for text-to-image generation. \cite{zhou2023lima} used vanilla fine-tuning instead of RL on a small amount of instruction data, and got similar performance to RLHF method. This is differnt from the empirical finding in \cite{ramamurthy2023is} that RL-based methods are 5 times more data-efficient than vanilla fine-tuning for aligning a PLM. The comparison and relationship between RL and fine-tuning remain at the forefront of active discussions. 

Note that these methods of optimizing a PLM toward the preferences of a group of annotators still exhibit certain limitations at the current stage. According to the self-evaluation of recent works \cite{ouyang2022training, bai2022training}, although RLHF brings significant improvements in truthfulness, informativeness and toxicity, it does not help reduce bias towards social groups, and is not designed for robustness towards distribution shifts or adversarial attacks. \cite {bai2022training} self-identified lack of robustness of their reward model, which leads to overfitting of the policy model during training. Some later works \cite{chen2023robust, wang2023on} did systematic evaluation and found that current large models fine-tuned with RLHF show improved robustness but still suffer from significant performance degradation under adversarial attacks and distribution shifts.

\subsubsection{Connections to ERM}
We have discussed three prominent techniques tailored for large models, namely, fine-tuning, prompting, and parameter-efficient fine-tuning. All these techniques impose some restrictions on the hypothesis space to effectively preserve the ``knowledge'' in the pretrained model. Their training objectives also have a lot of commonalities.

Fine-tuning adopts the following objective
\begin{align}
    \argmin_{\theta} 
    \mathop{\mathbb{E}}_{(x,y) \sim P(X,Y)}l(f(x;\theta), y)
    \label{eq:llm:ft}
\end{align}
where $\theta_0 = \theta_{\textrm{pretrained}}$, meaning we initialize the model with the parameters of a pretrained model. Typically, we set a small learning rate and train the model for fewer epochs compared to a standalone model, so that the solution $\hat{\theta}^\ast$ is close to $\theta_0$.

Parameter-efficient fine-tuning adopts the following objective
\begin{align}
    \argmin_{\theta} 
    \mathop{\mathbb{E}}_{(x,y) \sim P(X,Y)}l(f(x;[\Theta; \theta]), y)
    \label{eq:llm:peft}
\end{align}
where $\Theta$ represents the parameters to be fixed and $\theta$ represents the parameters to be trained. In particular, for adaptor-based methods, $\Theta$ represents all parameters of the pretrained model, and $\theta$ represents the parameters of the adaptor. $\theta$ usually has far less parameters than $\Theta$, i.e., $|\theta| \ll |\Theta|$.

Prompting adopts the following objective
\begin{align}
    \argmin_{\theta} 
    \mathop{\mathbb{E}}_{(x,y) \sim P(X,Y)}l(f(h(x;\theta);\Theta), y)
    \label{eq:llm:prompt}
\end{align}
where $h(\cdot; \theta)$ is an encoder that modifies the input to make it more easily processed by the pretrained model. A special case is ``hard prompt'', where $\theta$ consists of embeddings of natural language tokens. Again, $|\theta| \ll |\Theta|$.

It can be seen that although these techniques differ in their parameterizations, they all adopt the objective to minimize the empirical risk on the training data. In Section~\ref{sec:robustness}, we have discussed various improvements to the ERM objective to equip a standalone model with trustworthy properties, and we will see that they are applicable to large pretrained models as well.

In other words, since the master equations Equations~\ref{eq:master:1}, \ref{eq:master:2}, and \ref{eq:master:3} can all be built upon the ERM (Equation~\ref{eq:erm}), these equations should also be built upon the above extensions of ERM (Equations~\ref{eq:llm:ft}, \ref{eq:llm:peft}, \ref{eq:llm:prompt}) to boost their trustworthiness properties. 

\subsection{Trustworthy Solutions for Large Models} \label{sec:llm:trustworthy_techniques}

In this subsection, we present techniques aimed at enhancing the trustworthiness of large models. 
In Sections~\ref{sec:llm:dan}, \ref{sec:llm:da} and \ref{sec:llm:wt}, we review recent methods that extend the trustworthy methods originally developed for standalone models, to large pretrained models with different parameterizations of the ERM objective. 
Lastly, in Section~\ref{sec:lls:others} we introduce other techniques that boost the trustworthiness of large models.

\subsubsection{Domain-invariant representation learning} \label{sec:llm:dan}
One line of work focuses on learning representations that exhibit certain invariance across environments. 
We consider this line of works building upon Equation~\ref{eq:master:1}, although these works might not be invented through this process explicitly. 

For instance, \cite{jia-zhang-2022-prompt, wu2022adversarial} used DANN on the representation of a prompt token to align the features across different domains for text classification tasks. 
\cite{jia-zhang-2022-prompt} additionally minimized the KL divergence of predicted probability distributions of tokens for each category across different domains.
\cite{feder-etal-2021-causalm} designed adversarial fine-tuning tasks that perform gradient reversal on token representations associated with a chosen concept, to remove bias in the model related to that concept. They also compared the predictions of the original and debiased models to evaluate the treatment effect of the concept.
\cite{zaharia2022domain} used DANN for the domain adaptation of complex word identification models in multilingual and multi-domain settings. 
\cite{yue-etal-2022-qa} used the MMD loss on the token representations for the domain adaptation of question answering models.
\cite{wu2022continual} used the MMD loss to align the feature distribution of the new and old data for continual learning.
\cite{yu-etal-2022-interventional} partitioned the samples to different domains using an unsupervised method, and then used IRM to improve the OOD performance of several language understanding tasks.
\cite{Ren_2023_CVPR} used a combination of inter-domain and intra-domain contrastive learning, to push closer the representations of samples within the same category but different domains, for the domain generalization of image captioning models.

\subsubsection{Data augmentation methods} \label{sec:llm:da}
A large body of work uses data augmentation to enhance the trustworthy properties of pretrained models, which corresponds to Randomized Controlled Trial in causal inference (Section~\ref{sec:rct}) and master equation Equation~\ref{eq:master:2}. 

For example, \cite{guo2022auto} used a bias-inducing prompt to compare the different predictions of the PLM only at the change of the demographic words, and aligned the predicted distributions of the masked token to mitigate bias. 
Similarly, \cite{he-etal-2022-mabel} generated counterfactual samples by gender word substitution, and used contrastive learning and regularization to align the counterfactual pairs.
\cite{wang2021identifying} used interpretability methods to find the tokens important to the model's prediction, and identified the bias tokens among them by cross-corpora analysis and knowledge-aware perturbation. They found that masking the tokens at training or inference time improved the robustness of models. 
\cite{xiao2023masked} masked patches of the images based on the class activation map and refilled them with randomly sampled images. 
\cite{calderon2022docogen} masked domain-related terms in the source domain text input, and use a T-5 model to reconstruct the counterfactual text corresponding to the target domain.
Similarly, for open-domain dialogue system, \cite{ou-etal-2022-counterfactual} identified and replaced the keyword corresponding to the response perspective, and used BART to reconstruct a response with different semantics for the same dialogue history. 
Data augmentation often comes with task-specific designs. For example, for the continual learning of relation extraction models, 
\cite{wang-etal-2022-learning-robust} used a mix-up style augmentation to interpolate between labels, and reversed the asymmetric relations to create new relation types. 
For entity typing, \cite{xu-etal-2022-model} prompted a PLM to identify six types of biases in the dataset, and devised counterfactual augmentation for each kind of bias.
For factual probing, \cite{li-etal-2022-spe} considered the symmetry of tasks (subject prediction and object prediction), and designed a symmetric prompt enhancement method for prompt learning. 

Sometimes it is easier and more flexible to augment the data in the representation space.
\cite{yue-etal-2022-qa} used data augmentation for the domain adaptation of question answering models. They sampled augmented embeddings of question tokens from the convex hull spanned by the 2-hop synonyms of the token.
\cite{zhao2022Certified} improved adversarial robustness of language models by data augmentaion in the embedding space.
\cite{de-raedt-etal-2022-robustifying} proposed a method to generate label-flipping counterfactual text representation. They used the mean difference between the text representation of both labels as the displacement vector to perturb the original text representation.
For domain adaptation, \cite{long2022domain} generated domain-adversarial perturbation on the sample representation that fools the domain classifier into predicting the target domain, and used contrastive learning to push the counterfactual pair closer, using some other samples in the original domain as negative samples. 
\cite{he-etal-2022-cpl} sampled counterfactual and factual image features and used contrastive learning to learn more robust prompts.

\subsubsection{Sample weighting methods} \label{sec:llm:wt}
Some other works have used techniques to weight training samples based on their difficulty or biased content, capturing the principles of group-DRO \cite{Sagawa2020Distributionally} to model the worst-case distribution shift, or inverse probability weighting (Section~\ref{sec:ipw}) to establish independence between the confounder and treatment variable. This line of work corresponds to the line of works we summarize in master equation Equation~\ref{eq:master:3}.

For instance, \cite{yu-etal-2022-coco} clustered the input queries using K-means, and used the implicit DRO loss to make dense retrieval models generalize better to unseen queries.
\cite{he2022on} weighted training data with the Focal Loss \cite {lin2017focal} for the imbalanced classification in bug detection.
As a special case of weighting, some works selected part of the samples for training and discarded the rest.
\cite{su2022comparison} studied active learning for domain adaptation tasks where the source domain data are not available. They selected target domain samples with a large entropy from the source domain model, and manually annotated them for fine-tuning.
\cite{broscheit2022distributionally} clustered the samples and then selected samples in the $k$ most difficult clusters, to train a model that is robust to covariate shift for spoken language understanding. 
For generation tasks, some works identified the biased samples and removed them for fine-tuning a large model. \cite{dong2023raft, zhao2023slichf} filtered the samples using a reward model trained with human feedback data.
\cite{wu2022generating} used $z-$filtering to remove bias-aligned samples and fine-tuned a GPT-2 model towards generating unbiased training data for the NLI task. They extract spurious features based on prior knowledge of the task, and used $z-$statistics to measure correlation between the bias and the label. 

In Section \ref{sec:llm:rlhf} we discussed RLHF, which can also be used to boost the trustworthiness of large models. However, it seems that RLHF and the above-mentioned methods are orthogonal in the aspects of trustworthiness they are good at. RLHF is good at the human-centered concepts such as politeness or adherence to user's instructions, which are hard to be mathematically defined. On the other hand, robustness and fairness can be more effectively approached with statistical and causal inference methods (Section~\ref{sec:robustness}, \ref{sec:review:main}). We hypothesize that a combination of both family of methods would endow large models with more trustworthy properties.

\subsubsection{Other trustworthy methods} \label{sec:lls:others}
We will also briefly introduce some other trustworthy methods which do not fall into the master equations of our survey.
As large language models have shown stronger generative abilities, recent works have focused on a new approach to explainability known as self-rationalization, in which the model provides rationales for its own predictions \cite{wiegreffe2021measuring}. There are primarily two kinds of rationales: extractive and abstractive. Extractive rationales \cite{lei2016rationalizing} involve extracting fragments from the input to serve as explanations, while abstractive rationales \cite{wiegreffe2021measuring} generate free text for explanations.

For instance, \cite{chen2022can} demonstrated that training a language model to produce extractive rationales can potentially enhance its robustness against adversarial attacks. \cite{marasovic2021few} used prompt learning to fine-tune a language model in a few-shot setting, enabling it to generate abstractive rationales. They observed that the quality of rationales and the performance on end-tasks both improved as the scale of the model increased. \cite{mao2023doubly} proposed a technique where an image is sent with a learned ``why prompt'' as input to help the visual model not only recognize the object in the image but also explain why that object is predicted.

Self-rationalizatin is closely related to another line of work that focuses on enhancing the reasoning abilities of language models. For instance, in \emph{chain-of-thought prompting} \cite{wei2022chain}, reasoning steps were incorporated into demonstration examples for in-context learning, prompting the model to generate reasoning steps before predicting the answers to target questions. This led to consistent performance gain on arithmetic, symbolic, and commonsense reasoning benchmarks. Another study \cite{kojima2022large} discovered that using a simple prompt such as ``Let's think step by step'' could elicit reasoning steps and enhance performance on reasoning tasks. Subsequent works such as \cite{zhou2023leasttomost, chen2023large, wang2022iteratively} have explored alternative ways of prompting or more reasoning tasks.

A growing body of evidence suggests that training or prompting a model to provide rationales or reasoning steps can improve its performance on various tasks that require reasoning \cite{rajani2019explain, wei2022chain}, and also enhances out-of-distribution robustness \cite{wei2022chain, zhou2023leasttomost, anil2022exploring}.
Self-rationalization also provides a way to potentially improve the reliability of outputs by verifying the truthness of rationales \cite{cobbe2021training} or ensembling different rationales \cite{wang2023selfconsistency}. For a more comprehensive overview, we direct interested readers to the survey papers on language model reasoning \cite{huang2023reasoning} and rationalization \cite{gurrapu2023rationalization}. 

There are other trustworthy techniques for large models that are not included in our survey. One line of work augments the large model with retrieval or web search functionality \citep{guu2020retrieval, izacard2022atlas, nakano2022webgpt, ram2023incontext, liu2023webglm} so that it can make predictions based on the grounding text. This helps the model to access truthful and up-to-date knowledge and reduce hallunication \cite{shuster-2021-retrieval}. Some other works modify the model's behaviors at inference time for more reliable predictions \citep{schick2021selfdiagnosis, meade2021empirical, udomcharoenchaikit-etal-2022-mitigating, ge2023improving, zou2021controllable}. While these methods have great application values, in this survey we focus on techniques that make large models \emph{inherently} more trustworthy.

\subsection{Summary}
In this section we discussed the trustworthy challenges, common techniques, and trustworthy solutions for large models. We discussed prompting, fine-tuning and parameter-efficient tuning, and unified them under the ERM objective, thereby showing that the master equations discussed in Section~\ref{sec:robustness} can be extended to large models. The RLHF technique seems to be complementary to our master equations in the aspects of trustworthiness it improves. As new techniques for large models are being continuously proposed, we believe their combination with the master equations has the potential to further enhance the trustworthiness of large models.

\section{Applications Where Trustworthy Machine Learning Play Important Roles} \label{sec:application}
In this section, we will focus on application scenarios where trustworthiness of models plays an important role. We introduce recent efforts that evaluate the performance of large models in these scenarios, or discuss their limitations or possible solutions. We will also give a brief overview of recent works that develop trustworthy methods focusing on the application perspective.
We categorize our discussion into vision applications, 
language applications, and vision-language applications. 

\subsection{Vision}

In vision applications, we often want a machine learning model to understand the semantic information of images, whereas the non-semantic factors such as the texture, background, viewpoints, lighting conditions and all kinds of noise can act as confounders that hurt the trustworthiness of vision models. Some application scenarios also suffer from the data scarcity and long-tail distribution problems. 
Different methods have been proposed to tackle the challenges in a variety of tasks
including image classification 
\cite{akula2021cxtom, eckstein2021discriminative, chang2021robust, goyal2019counterfactual, Wang2021CausalAF, tang2021adversarial, vermeire2020explainable, wang2021proactive, ijcai2020-742, zhao2020fast, plumb2021finding, 9656623, smith2020counterfactual},
object detection \cite{9643182},
object localization \cite{shao2021improving},
co-saliency detection \cite{https://doi.org/10.48550/arxiv.2108.02093},
facial recognition \cite{CIActionUnit2022}, 
emotion recognition \cite{chen2021unbiased},
semantic segmentation \cite{shen2021conterfactual, Zhang2020CausalIF}, 
meta-learning \cite{yue2020interventional, yue2021counterfactual, shen2021conterfactual}, 
sequential tasks
\cite{ijcai2021-182, 9607759, 10.1145/3474085.3475472, chen2021human, liu2021blessings},
dataset preparation \cite{ reddy2021causally, mcduff2021causalcity, https://doi.org/10.48550/arxiv.2108.02093, yi2020clevrer},
knowledge distillation
\cite{deng2021comprehensive, Hu2021DistillingCE},
and
medical imaging \cite{mohit2021, https://doi.org/10.48550/arxiv.2111.12525, Singla2021UsingCA, [https://doi.org/10.48550/arxiv.2203.01668](https://doi.org/10.48550/arxiv.2203.01668), article, [https://doi.org/10.48550/arxiv.2110.14927](https://doi.org/10.48550/arxiv.2110.14927), Mertes2022GANterfactualCounterfactualEF, oh2021born, singla2021explaining, white2021contrastive, 2020, Mertes2022GANterfactualCounterfactualEF, sani2021explaining}.

Many works have employed data augmentation methods to extract causal features or remove confounders \cite{chang2021robust, https://doi.org/10.48550/arxiv.2108.02093, sauer2021counterfactual, plumb2021finding, https://doi.org/10.48550/arxiv.2203.04694, tang2021adversarial}.

A subset of the literature approached the data scarcity problem by robust few-shot learning, intervening on features, normalizing them, or using synthetic images \cite{yue2020interventional, shen2021conterfactual, yue2021counterfactual}. 
Several studies aimed to mitigate bias in sequential tasks, often through treatment effects \cite{ijcai2021-182, 9607759, 10.1145/3474085.3475472, chen2021human}. 
For knowledge distillation,  
\cite{deng2021comprehensive}
used backdoor adjustment (Section~\ref{sec:backdoor_adjustment})
to eradicate the confounding effect of knowledge prior which determines the context or background features of the image.
Additionally, 
\cite{Hu2021DistillingCE} quantified the effect of an intervention, to eradicate the forgetfulness problem in student models 
by enforcing the same representation from the student model.

Furthermore, several works \cite{akula2021cxtom, eckstein2021discriminative, goyal2019counterfactual, vermeire2020explainable, wang2021proactive, ijcai2020-742, zhao2020fast, smith2020counterfactual} tried to enhance the interpretability of the models, mostly by generating an image that flips the label to emphasize the important features responsible for the original label. For example, \cite{smith2020counterfactual} aimed to measure robustness in robot control by quantifying the change required to generate a counterfactual image that has a ground-truth label but is predicted wrongly by the robot.

Additionally, some studies have created vision datasets for training trustworthy models \cite{reddy2021causally, mcduff2021causalcity, https://doi.org/10.48550/arxiv.2108.02093}. For example, \cite{mcduff2021causalcity} provided a tool for constructing data with specified characteristics for use with self-driving cars. The tool offers a high-fidelity simulation environment that can be used for causal and temporal reasoning, providing control over variables such as environment characteristics, vehicles, and traffic lights.

\paragraph{Medical Imaging} 

Machine learning applied to medical imaging holds great potential and societal value. By automatically recognizing the patterns in imaging data, machine learning models have the potential to greatly reduce the workload of radiologists and alleviate the strain in medical resources. They may also detect the subtle abnormalities reflecting the early stages of disease which may not be easily recognizable by humans \cite{esteva2017dermatologist}. However, the high-stake nature of medical imaging places high demands on the trustworthiness of machine learning models, because a wrong prediction could bring severe outcome to the patient's health. The model should be interpretable enough for medical experts to easily validate and adopt its predictions. Distribution shifts are common in imaging data due to different medical devices, imaging procedure and individual characteristics. High-quality labeled data are often scarce, due to privacy concerns and the expensive cost of annotation, which may lead to overfitting problems \cite{2020}.

Recent works have evaluated the performance of vision foundation models on medical imaging tasks.
\cite{Shi_2023} evaluated the SAM model on medical images in different applications including dermatology, ophthalmology, and radiology. The author discovers that SAM is not able to perform optimal for certain structured targets like continuous branch structures. However, fine-tuning the model on small amount of this data leads to several improvements.
Similarly, \cite{mazurowski2023segment} evaluated the SAM model on medical imaging tasks using prompt learning. They discovered that prompt performance varies on datasets and does perform poorer on ambiguous segmentation tasks. Furthermore, the author discover that when prompts are provided iteratively their performance does not improve as much as the previous models.
Besides numerical evaluation, \cite{trustworthyaiinml} took a societal view and highlighted the marginalization of people with rare diseases in the dataset, which may cause unintentional discrimination.

Different methods have been proposed to improve the trustworthiness of models for medical imaging. \cite{2020} discussed the importance of causal features to alleviate issues arising from domain-shift. Other studies such as \cite{mohit2021, https://doi.org/10.48550/arxiv.2111.12525} focused on delineating invariant features. \cite{mohit2021} distinguished between causal and contrast features for COVID-19 CT scans, whereas \cite{https://doi.org/10.48550/arxiv.2111.12525} employed augmentation and pseudo-correlations to address domain shift issues in cross-modality (CT-MRI) abdominal image segmentation, cross-sequence (bSSFP-LGE) cardiac MRI segmentation, and cross-center prostate MRI segmentation.
Moreover, \cite{Singla2021UsingCA} employed treatment effects (Section~\ref{sec:counterfactuals:te}) to perturb various clustered units representing human-understandable concepts, aiming to identify units that serve as causal features. For interpretability, the study also trained a decision tree to map output to an explanation, thereby creating a rule-based explainer.

Techniques of counterfactual image generation have been utilized by \cite{[https://doi.org/10.48550/arxiv.2203.01668](https://doi.org/10.48550/arxiv.2203.01668), article, [https://doi.org/10.48550/arxiv.2110.14927](https://doi.org/10.48550/arxiv.2110.14927), Mertes2022GANterfactualCounterfactualEF, oh2021born, mohit2021, singla2021explaining, white2021contrastive} to enhance the interpretability of medical models. Counterfactual images are generated to underscore causal features \cite{oh2021born}, and features contributing both negatively and positively are revealed \cite{[https://doi.org/10.48550/arxiv.2110.14927](https://doi.org/10.48550/arxiv.2110.14927)}. 
However, the task of image generation often results in poor quality. To mitigate this, \cite{singla2021explaining, Mertes2022GANterfactualCounterfactualEF,white2021contrastive} focused on improving the quality of generated images by emphasizing textural and structural information \cite{Mertes2022GANterfactualCounterfactualEF} for pneumonia classification from X-rays. Additionally, \cite{sani2021explaining} evaluated the reconstruction quality of different segments of chest X-ray images, along with object retention. 
Yet, focusing solely on intensified discriminatory pixels can compromise the interpretability of less-intensified pixels that nonetheless contribute significantly to prediction. To address this, \cite{white2021contrastive} proposed using a regression equation to quantify the contributions of different image segments rather than relying solely on visual appearance. This work also incorporated the factor of overdetermination to provide a broader view of the factors contributing to a label. 

Counterfactual explanations for medical images with high uncertainty occasionally produce blatantly incorrect explanations, but these can be improved \cite{article} through reliable elucidation of the intricate relationships between image signatures and the target attribute. Nevertheless, all the aforementioned techniques might not perform well with medical images exhibiting high variability in data-shift. As a result, \cite{[https://doi.org/10.48550/arxiv.2203.01668](https://doi.org/10.48550/arxiv.2203.01668)} sought to discover the statistical relationship between inter-species images, where the DL model fails due to its low generalizability properties. This work disentangled the information of medical images using a Directed Acyclic Graph to generate counterfactual images based on low-level and high-level features.

\subsection{Language}
Large language models have significant application potential in multiple areas such as medicine, education, software engineering, law, and finance. In the medical domain, large models can quickly digest and interpret the vast amount of medical literature, helping clinicians and researchers keep up with the latest research findings and treatment options. This is particularly helpful given the exponential growth of publications \cite{thapa2023chatgpt}. Based on the research literature and the latest medical guidelines, large models can also aid in clinical decision-making \cite{singhal2023large}. Medical chatbots can be developed to automate certain aspects of patient communication, such as answering frequently asked questions of patients \cite{gentner2020systematic}. Large models can also provide initial analysis on mental health issues \cite{yang2023interpretable}, improving the accessibility of mental health support. In the education domain, large models can aid in producing learning content and facilitate more accessible and personalized learning experience for individuals at all levels of education \cite{kasneci2023chatgpt}. For example, it can promote the curiosity-stimulating learning of children \cite{abdelghani2023gpt}, explain the difficult parts of learning materials, and provide information to college students on a particular research topic, etc. For teachers, it can aid in lesson planning by generating candidate lesson activities and practice problems, and semi-automate the assessment of students' work, helping teachers provide timely and personalized feedback to students \cite{kasneci2023chatgpt}. The potential applications of language models in other fields such as law \cite{cui2023chatlaw}, finance \cite{wu2023bloomberggpt}, and software engineering \cite{ross2023programmer} have also been actively explored by researchers and practitioners. Due to the breadth of these applications, they are beyond the scope of this survey. Readers interested in further exploration are encouraged to consult these referenced works.

In these scenarios, it is important for the model to provide truthful answers, and hallucination is a big issue. Because large models excel at simulating the form of human language, a wrong response from the model can be very misleading to humans. Due to the opaque nature of large models, it is not clear whether the model has been equipped with the adequate domain knowledge, or whether it can correctly conduct the reasoning steps over the knowledge, to generate a truthful response to the user's question. These problems, together with the privacy and fairness issues, affect the usefulness of language models in high-stake application scenarios.

Recent works have evaluated the performance of large language models for various applications.
\cite{dash2023evaluation} evaluated the GPT-3.5 and GPT-4 models in healthcare with the help of different physicians, based on the answers given by these models on several questions. In conclusion, these models were not defined as harmful in majority opinion, but sometimes generate inaccurate responses or hallucinate references. 
\cite{sallam2023chatgpt} highlighted the utility of ChatGPT in healthcare education, research, and practice. It gives an overview of how ChatGPT is affecting these domains of healthcare where it is being used in improving the scientific writing.
\cite{Shen2023-nn} discussed the merits and challenges of large language models in clinical settings. They highlighted the potential of these models to interact with patients, assist physicians, and simplify the communication with insurance providers. However, their limitations include generating unrealistic outputs (hallucinations), inability to access real-time data, and challenges with specific tasks like summarizing patient history or image manipulation. The use of LLMs in professional publications is also limited due to their authoritative style of writing.
In the education domain, \cite{kasneci2023chatgpt} discussed the trustworthy challenges of large language models, deeming that these challenges are not unique to education but inherent in the more generic technologies. They call for more intellectual efforts in understanding these technologies and their unexpected limitations, as well as pedagogical approach that emphasizes critical thinking abilities and strategies for fact checking.

Various methods have been proposed to address the trustworthy challenges of large language models, many of which are motivated by the notion of causality. 
These works have focused on the application of large models in various tasks such as 
sentiment analysis \cite{sridhar2019estimating, kaushik2020learning, gardner-etal-2020-evaluating, bi2021interventional, wang2022causal}, natural inference tasks \cite{kaushik2020learning}, temporal relation prediction \cite{gardner-etal-2020-evaluating}, question answering \cite{gardner-etal-2020-evaluating}, and multi-label intent classification \cite{calderon2022docogen}.
Based on the concepts outlined in section~\ref{sec:review:main},
we can discuss the variety of methods around the second and third levels of the Causal Ladder.
For example, the works \cite{robust2016textclassification, wang2022causal, bi2021interventional} harnessed the second level of causation to eradicate the confounding effects.
Specifically, \cite{robust2016textclassification} and \cite{bi2021interventional} used backdoor adjustment (Section~\ref{sec:backdoor_adjustment}) to achieve a causal-driven system.  
On the other hand, \cite{sridhar2019estimating, kaushik2020learning, gardner-etal-2020-evaluating, garg2019counterfactual, Feder_2021, zmigrod-etal-2019-counterfactual, wu2021polyjuice, calderon-etal-2022-docogen, vig2020investigating, meng2023locating, elazar2021amnesic} leveraged the third level of causation to solve various problems regarding fairness \cite{garg2019counterfactual, zmigrod-etal-2019-counterfactual, vig2020investigating}, interpretability \cite{elazar2021amnesic, meng2023locating} 
and 
robustness via data augmentation \cite{kaushik2020learning, gardner-etal-2020-evaluating, Feder_2021, wu2021polyjuice, calderon-etal-2022-docogen}.
For the task of online debates, \cite{sridhar2019estimating} used treatment effect (Section~\ref{sec:counterfactuals}) to enhance the quality by estimating the causal effect of reply tones on linguistic and sentiment changes. 
They mitigated the confounding effect of users' ideologies acting as confounders. 
Similarly, \cite{vig2020investigating} alleviated gender bias in pre-trained language models via treatment effect.
\cite{kaushik2020learning, gardner-etal-2020-evaluating} generated the counterfactual samples via manual engineering.
Meanwhile, \cite{zmigrod-etal-2019-counterfactual} focused on generating counterfactual data to mitigate gender biases via the replacement of keywords.
However, manually generating the data can be sup-optimal and time-consuming. 
Therefore, some research endeavors employed automatic methods for counterfactual data generation.
For instance, \cite{wu2021polyjuice} fine-tuned GPT-2 on multiple datasets of paired sentences to automatically generate counterfactual samples. 
\cite{calderon-etal-2022-docogen} generated counterfactuals by infusing different domains for domain adaptation of a sentiment classifier and a multi-label intent classifier.
Regarding interpretability, \cite{meng2023locating, elazar2021amnesic} used counterfactual intervention for identifying neuron activations that are important to a model’s predictions.

\subsection{Vision-Language}
Vision-Language models are more flexible and versatile in the way they process information and interact with humans, and therefore has immense potentials in medical applications. By integrating information from different modalities, models could offer more holistic medical decision support \cite{foundation_models_medical_ai}. They may also support automatic generation of radiology reports by describing findings from imaging data \cite{foundation_models_medical_ai}, or offer visual aid to patients for the chatbot to provide more easy-to-understand medical education. However, vision-language models inherit the trustworthy challenges from both vision and language models. Additionally, we need to make sure that the model does not exploit heuristics from a single modality, but integrates information from both modalities for prediction.

The performance of large vision-language models on medical tasks has been evaluated in recent work. \cite{chambon2022adapting} studied prompt-based medical image generation and found that the Stable Diffusion model \cite{rombach2022high} shows limited performance by both manual observation and the FID-score, before fine-tuning on medical data.  
\cite{Moon_2022} explored the performance of pre-trained models on the medical domain task, more specifically using BERT-based architecture to boost the multi-modal performance on radiology images and unstructured reports. They modified the attention mechanism of the model in the medical domain, where the attention map can attend to disease-discriminatory regions. Furthermore, their method is able to generate clinically appropriate reports using  abbreviated medical terms. 
\cite{qin2023medical} studied the transferability of vision-language models on the medical domain. They designed prompts by injecting image-specific information to bridge the gap across medical domains and improve generalizability. 
They developed prompts via automatic generation using three strategies: masked language model (MLM) driven auto-prompt generation, image specific auto-prompt generation, and a hybrid of both.
Apart from numerical evaluation, \cite{foundation_models_medical_ai} proposed to build generalist medical models by formally representing medical knowledge. In this way, it can perform the reasoning to solve unseen tasks, raise self-explanatory warnings, and give treatment recommendations.

Various methods have been proposed to alleviate the trustworthy issues of vision-language models on tasks such as visual question answering (VQA), dataset preparation \cite{ates2020craft}, sequential tasks \cite{chadha2021ireason, fu2020counterfactual, 10.1145/3474085.3475540}, and video-grounding \cite{https://doi.org/10.48550/arxiv.2106.08914, zhang2020counterfactual, nan2021interventional, wang2021weaklysupervised}. 
Video-grounding studies \cite{https://doi.org/10.48550/arxiv.2106.08914, zhang2020counterfactual} use counterfactual generation (Section~\ref{sec:counterfactuals}). \cite{https://doi.org/10.48550/arxiv.2106.08914} utilized an object-action factorized Structural Causal Model (SCM) to explicitly incorporate action and object information. \cite{zhang2020counterfactual} generated positive and negative counterfactuals for the model to learn causal features. 
In the same area, \cite{nan2021interventional, wang2021weaklysupervised} applied backdoor adjustment (Section~\ref{sec:backdoor_adjustment}). \cite{wang2021weaklysupervised} removed the spurious correlation between the background and causal features, while \cite{nan2021interventional} mitigated selection bias in the dataset, which causes the model to predict the moments based on the presence of certain objects rather than the activities.
For VQA tasks, several studies \cite{study1, study2, study3} noted over-reliance on linguistic correlations rather than multimodal reasoning, which \cite{niu2021counterfactual} addressed through treatment effects (Section~\ref{sec:counterfactuals:te}).
Further, \cite{chen2020counterfactual, 9706896, liang-etal-2020-learning} generated counterfactual and factual data using the third level of causation (Section~\ref{sec:counterfactuals}). \cite{chen2020counterfactual, 9706896} created counterfactual pairs of images and questions to enforce robust predictions, while \cite{liang-etal-2020-learning} used contrastive learning to adjust pair distance. 
\cite{9156448} increased model robustness by generating data, intervening on exogenous variables. For VQA robustness, \cite{wang2020visual} incorporated commonsense reasoning using backdoor adjustment (Section~\ref{sec:backdoor_adjustment}), while \cite{rosenberg2021vqa} introduced augmented questions for resilience testing. 
In terms of dataset preparation, the CRAFT dataset \cite{ates2020craft} introduced three types of questions: descriptive, causal, and counterfactual. The counterfactual questions involve scenarios where an existing object is removed, and the causal questions involve understanding the interactions between objects through notions like ``cause'', ``enable'', and ``prevent''.
Regarding sequential tasks, various studies \cite{chadha2021ireason, fu2020counterfactual, 10.1145/3474085.3475540} tried to tackle different challenges. The study \cite{chadha2021ireason} proposed iReason, which integrates a rationalizing causal module in vision and language captions, emphasizing the perception of causal relationships in time-ordered events within videos. Similarly, \cite{fu2020counterfactual} focused on 3D navigation in visual surroundings using natural language instructions. The study applied the concept of the third level of causation (Section~\ref{sec:counterfactuals}) to sample challenging paths based on adversarial learning, thereby making the model more robust for navigation tasks. In a similar vein, \cite{10.1145/3474085.3475540} worked to remove long-tail bias in unraveling the dynamic interaction across visual concepts. The research argues that non-uniformly distributed predicates can cause model biases, and therefore focused on incorporating predicates uniformly via causal intervention for every subject and object.

\section{Conclusion}
\label{sec:con}
In this survey, we review trustworthy and aligned machine learning. We focus on the four topics of trustworthy properties, namely, robustness, adversarial robustness, fairness, and interpretability. We start from the limitations of the i.i.d. assumption, and reviews recent methods to improve trustworthiness of standalone machine learning models for different topics. Then, we revisit these methods to understand them from Pearl's causal hierarchy, and review trustworthy techniques that are directly based on or inspired by causal inference methods. We move on to review the techniques commonly used with large pretrained models, and show that the trustworthy techniques can be combined with them to address the current limitations of large models. 

\paragraph{Principled Solutions} Our survey found consistent solution patterns across various aspects of trustworthiness. These techniques were likely developed independently in their fields, but their convergence suggests a universal applicability. We believe these shared techniques will apply to large pretrained models and may also be useful for future generations of machine learning.

\paragraph{Models with All Trustworthy Merits} Our survey covered topics like robustness, adversarial robustness, fairness, and interpretability. While there has been much research in each area, we found no efforts to create models with all these properties at once. Our understanding of these aspects provides an opportunity to build a model that combines all these merits.

\paragraph{Modern Trustworthy ML with Modern Causality Theory} We connected various aspects of trustworthy machine learning to established causality theories. As causality research advances, new insights might inspire more effective trustworthy machine learning methods.

\paragraph{Additional Tiers of Trustworthiness} We focused mainly on the models' statistical understanding of data patterns, along with techniques such as augmentation and regularization. We predict that in the future, this level of trustworthiness will be a basic requirement for AI, especially as AIs begin to work together. More complex scenarios will likely demand higher levels of trustworthiness.

\subsection*{Acknowledgement}
The collaboration of this survey was initiated by the reading group event from Trustworthy Machine Learning Initiative (\href{https://www.trustworthyml.org/}{trustworthyml.org}). One can find more information of the reading group from the initiative or the gate website of this survey (\href{http://trustai.one}{trustAI.one}).

% \label{sec:apd}
% \input{secs/appendix}

%     Bibliography
%--------------------------------------------------------------
\bibliographystyle{acm}
\bibliography{ref}
\end{document}